\RequirePackage{snapshot}
\documentclass[10pt,twocolumn,letterpaper]{article}

\usepackage{iccv}
\usepackage[accsupp]{axessibility}  %
\usepackage{times}
\usepackage{epsfig}
\usepackage{graphicx}
\usepackage{amsmath}
\usepackage{amssymb}

\usepackage{booktabs}
\usepackage{graphbox} %
\usepackage{makecell}
\usepackage{pifont}%

\newcommand{\cmark}{\textcolor{green}{\ding{51}}}%
\newcommand{\xmark}{\textcolor{red}{\ding{55}}}%
\usepackage{gensymb}
\usepackage{capt-of} %
\usepackage{mathtools} %
\usepackage{multirow}
\usepackage{placeins} %
\usepackage{adjustbox}

\usepackage[pagebackref=true,breaklinks=true,letterpaper=true,colorlinks,bookmarks=false]{hyperref}

 \iccvfinalcopy %

\newcommand{\blfootnote}[1]{%
  \begingroup
  \renewcommand\thefootnote{}\footnote{#1}%
  \addtocounter{footnote}{-1}%
  \endgroup
  }
  
\begin{document}
\title{Geometry-Free View Synthesis: Transformers and no 3D Priors}

\author{Robin Rombach\thanks{} \qquad Patrick Esser\footnotemark[1] \qquad Bj\"orn Ommer\\
Ludwig Maximilian University of Munich\quad\&\quad IWR, Heidelberg University, Germany\\
\small{*Both authors contributed equally to this work. Code is available at
\url{https://git.io/JRPPs}.}
}
\providecommand{\imwidth}{}
\providecommand{\impath}[1]{}
\providecommand{\impatha}[1]{}
\providecommand{\impathb}[1]{}
\providecommand{\impathc}[1]{}
\providecommand{\impathd}[1]{}
\providecommand{\impathe}[1]{}

\newcommand{\compressions}{
\begin{figure}[h!]
\centering
    \setlength{\tabcolsep}{1pt}
    \renewcommand{\arraystretch}{1}
\begin{footnotesize}
\begin{tabular}{ccc}
Inputs & \footnotesize{\makecell{VQGAN f8: \\ $208\times 368 \to 26\times 46$}} & \footnotesize{\makecell{VQGAN f16: \\ $208\times 368 \to 13\times 23$}}\\
\includegraphics[width=0.3\textwidth]{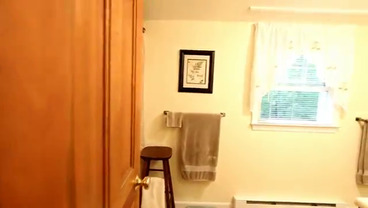} & 
\includegraphics[width=0.3\textwidth]{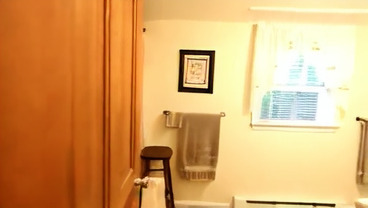} & 
\includegraphics[width=0.3\textwidth]{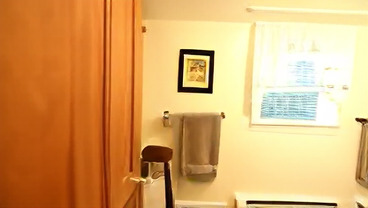} \\
\includegraphics[width=0.3\textwidth]{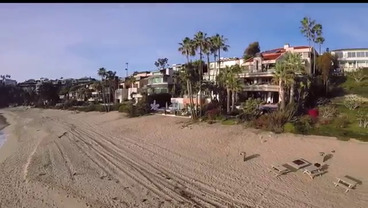} & 
\includegraphics[width=0.3\textwidth]{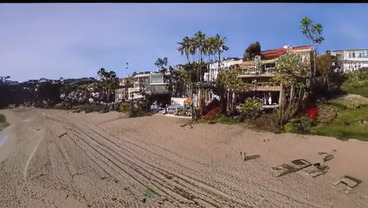} & 
\includegraphics[width=0.3\textwidth]{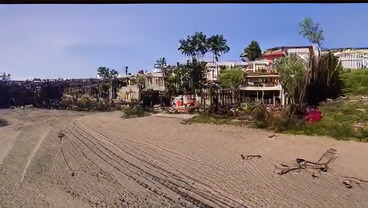} \\
\includegraphics[width=0.3\textwidth]{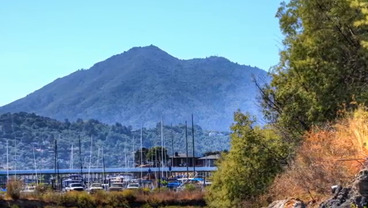} & 
\includegraphics[width=0.3\textwidth]{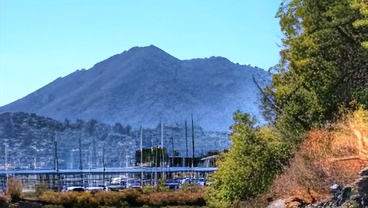} & 
\includegraphics[width=0.3\textwidth]{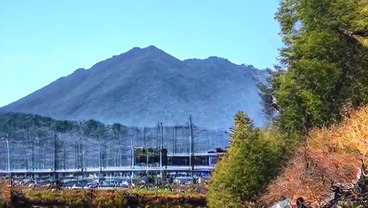} \\
\includegraphics[width=0.3\textwidth]{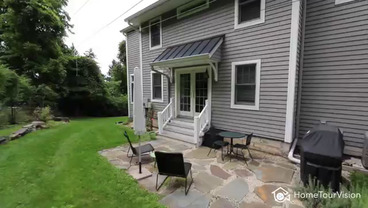} & 
\includegraphics[width=0.3\textwidth]{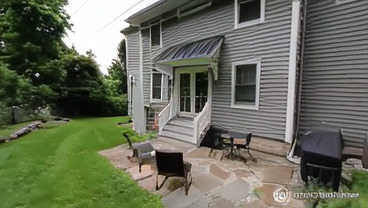} & 
\includegraphics[width=0.3\textwidth]{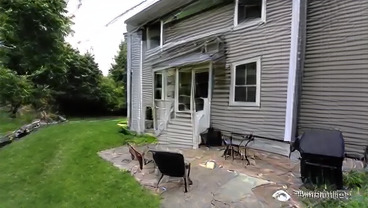} \\
\includegraphics[width=0.3\textwidth]{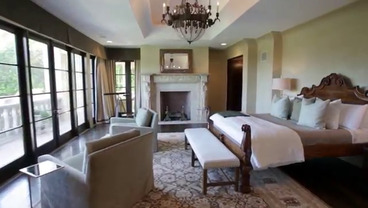} & 
\includegraphics[width=0.3\textwidth]{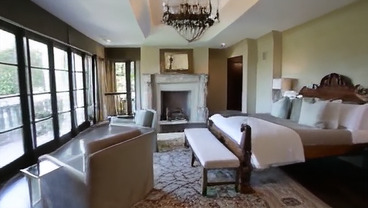} & 
\includegraphics[width=0.3\textwidth]{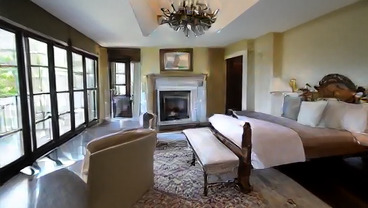} \\
\end{tabular}
\caption{\label{fig:compressions} Qualitative visualization of different compression rates.}
\end{footnotesize}
\end{figure}
}

\newcommand{\reconstructionmetrics}{
\begin{table}[thbp]
\centering
\begin{footnotesize}
\begin{tabular}{l cccc}
Model & PSIM $\downarrow$ & R-FID $\downarrow$ & PSNR $\uparrow$ & SSIM $\uparrow$ \\
\toprule
VQGAN f16 & $1.48 \pm 0.37$ & $2.46$ & $21.14 \pm 2.45$ & $0.67 \pm 0.14$ \\
VQGAN f8 & $0.80 \pm 0.28$ & $1.52$ & $25.23 \pm 2.96 $ & $0.82 \pm 0.11$ \\
\bottomrule
\end{tabular}
\end{footnotesize}
\caption{\label{tab:reconstructionmetrics} Reconstruction metrics for different
compressions.
  }
\end{table}
}
\newcommand{\thenumbers}{
\begin{table}[thbp]
\centering
\begin{footnotesize}
\begin{tabular}{l ccc}
Model & PSIM $\downarrow$ & PSNR $\uparrow$ & SSIM $\uparrow$ \\
\toprule
  SynSin (small cam-$\Delta$)                    & $1.13 \pm 0.54$ & $23.03 \pm 4.54$ & $0.77 \pm 0.12$ \\
  SynSin (at $208\times 368$)   & $1.32 \pm 0.53$ & $22.39 \pm 4.24$ & $0.76 \pm 0.12$ \\
  SynSin (w/o best of 2)   & $1.72 \pm 0.72$ & $19.75 \pm 4.69$ & $0.67 \pm 0.16$ \\
  SynSin (w/o $\text{src}\equiv\text{tgt}$)   & $1.84 \pm 0.66$ & $18.86 \pm 3.98$ & $0.64 \pm 0.15$ \\
  SynSin (w/ 50\% large)   & $2.48 \pm 0.91$ & $16.05 \pm 3.99$ & $0.57 \pm 0.13$ \\
\bottomrule
\end{tabular}
\end{footnotesize}
\caption{\label{tab:thenumbers} Effects of
  removing evaluation biases to small changes.}
\end{table}
}

\newcommand{\compare}{
\begin{figure}[bt]
\centering
\includegraphics[width=0.7\textwidth]{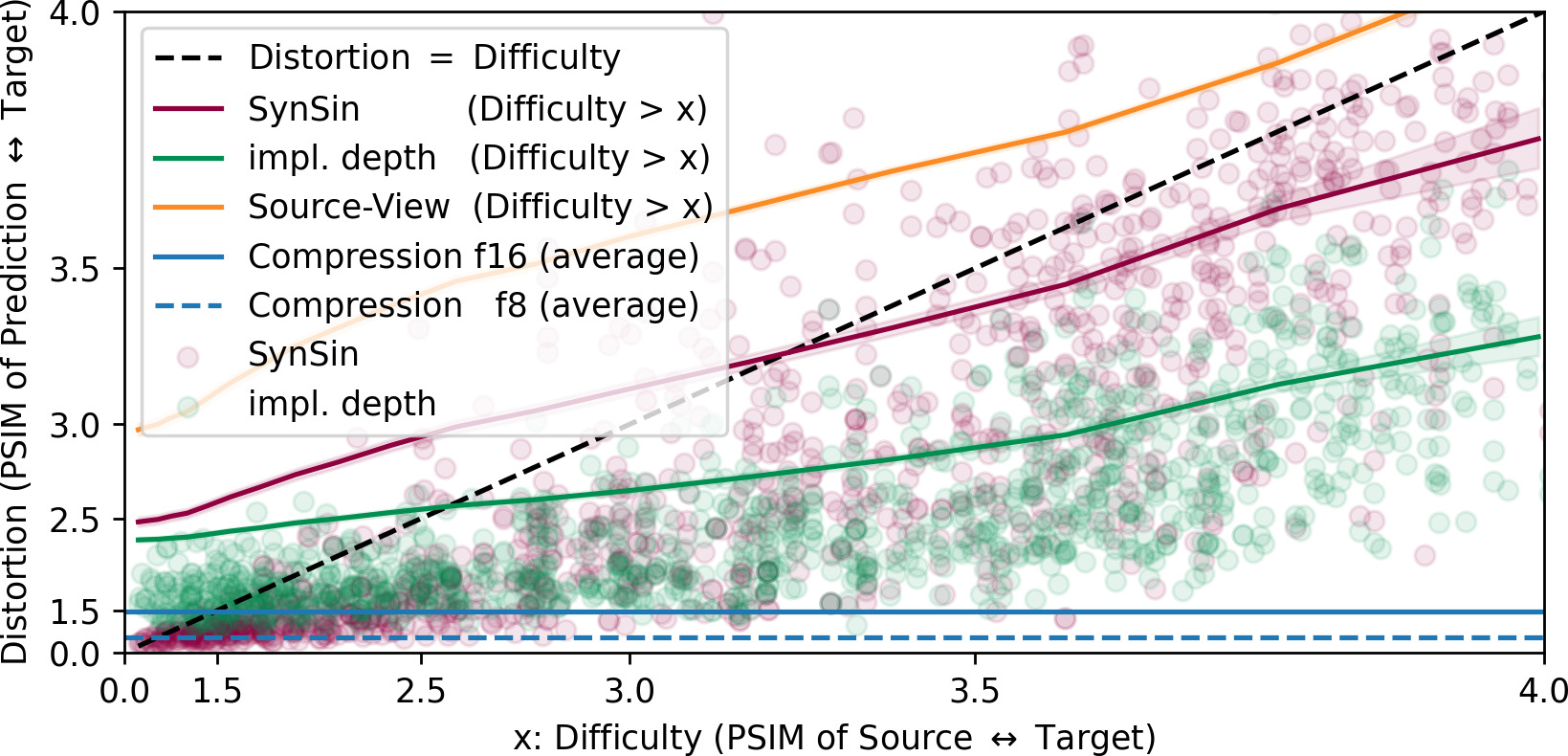}
\caption{\label{fig:compare} Comparison on small and large viewpoint changes.}
\end{figure}
}

\newcommand{\sequence}{
\begin{figure}[ht]
\centering
    \setlength{\tabcolsep}{0pt}
    \renewcommand{\arraystretch}{0}
\begin{footnotesize}
  \adjustbox{max width=\textwidth}{
\begin{tabular}{ccccc}
\includegraphics{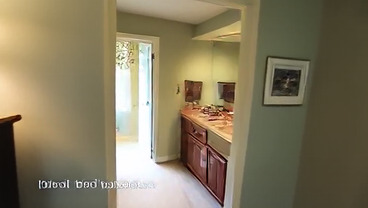} & 
\includegraphics{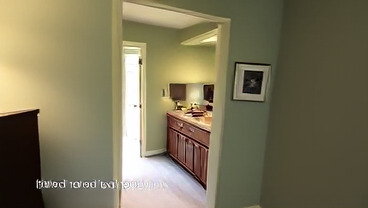} & 
\includegraphics{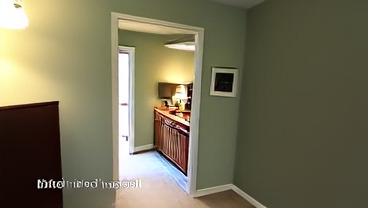} & 
\includegraphics{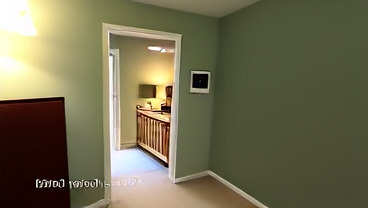} & 
\includegraphics{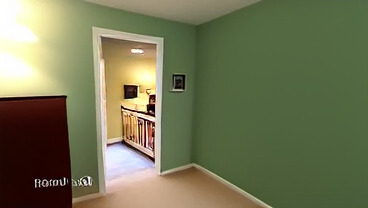} \\

\includegraphics{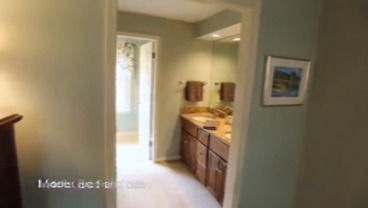} & 
\includegraphics{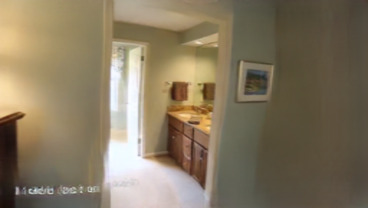} & 
\includegraphics{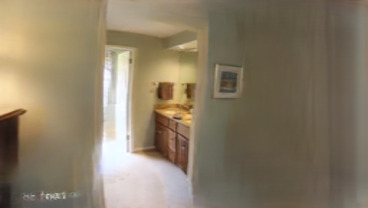} & 
\includegraphics{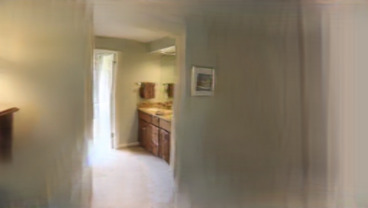} & 
\includegraphics{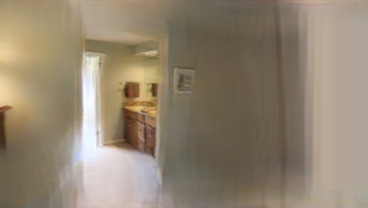} \\
\end{tabular}
  }
  \caption{\label{fig:sequence} Continuous trajectories: Ours {\footnotesize (top)}
  \& SynSin {\footnotesize (bottom)}.}
\end{footnotesize}
\end{figure}
}

\newcommand{\figdepthprobinglayers}        {
   \begin{figure}[t]
   \begin{center}
   \includegraphics[width=0.975\linewidth]{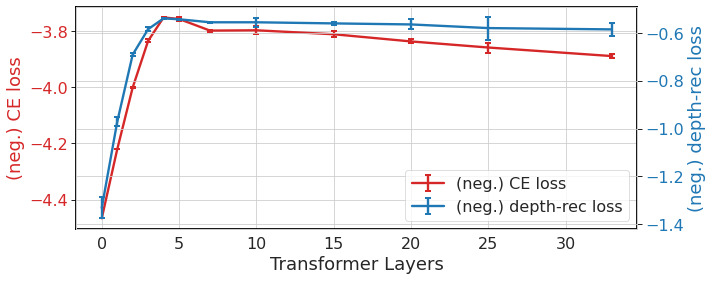}
   \end{center}
   \vspace{-1.5em}
      \caption{
Minimal validation loss and reconstruction quality of depth predictions obtained from linear probing
as a function of different transformer layers. The probed variant is \varvi{}. \vspace{-0.75em}
      }
   \label{fig:depthprobinglayers}
   \end{figure}
}

\newcommand{\figvideopreview}        {
   \begin{figure}[h!]
   \begin{center}
   \includegraphics[width=0.875\linewidth]{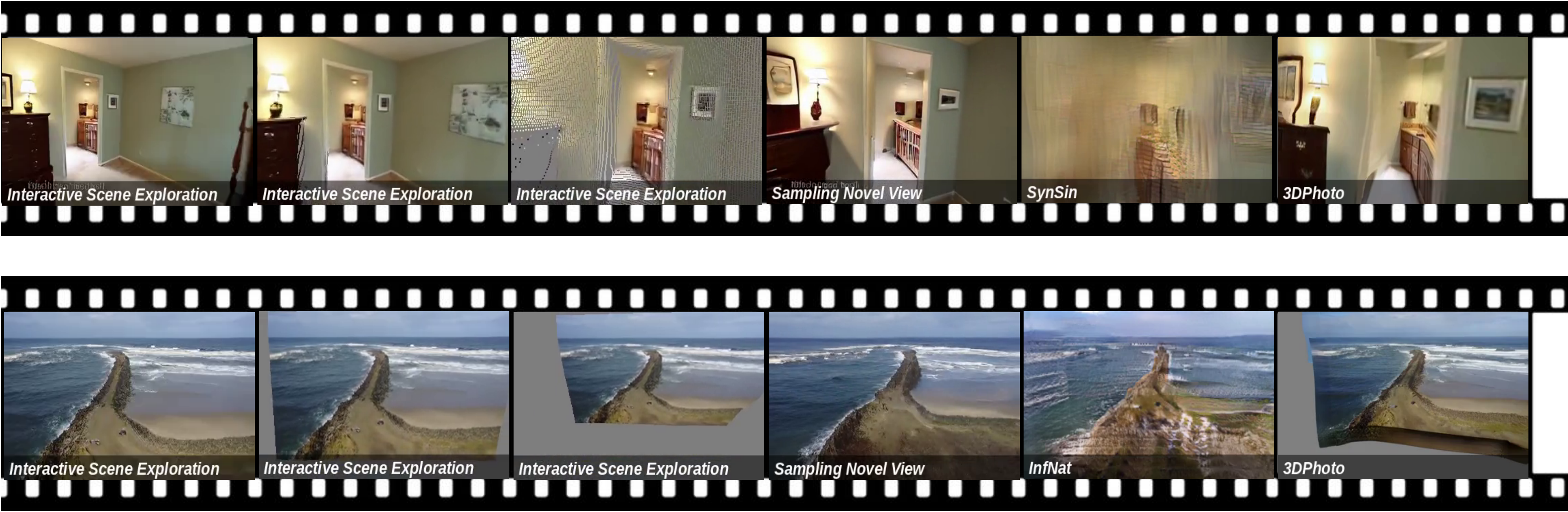}
   \end{center}
      \caption{Preview of the videos available at \url{https://git.io/JOnwn},
      which demonstrate an interface for interactive 3D exploration of images.
      Starting from a single image, it allows users to freely move around in
      3D. See also Sec.~\ref{sec:addresults}.}
   \label{fig:videopreview}
   \end{figure}
}

\newcommand{\figfirstpagefigure}{
  \vspace{-2.75em}
    \begin{center}
        \includegraphics[width=1.00\textwidth]{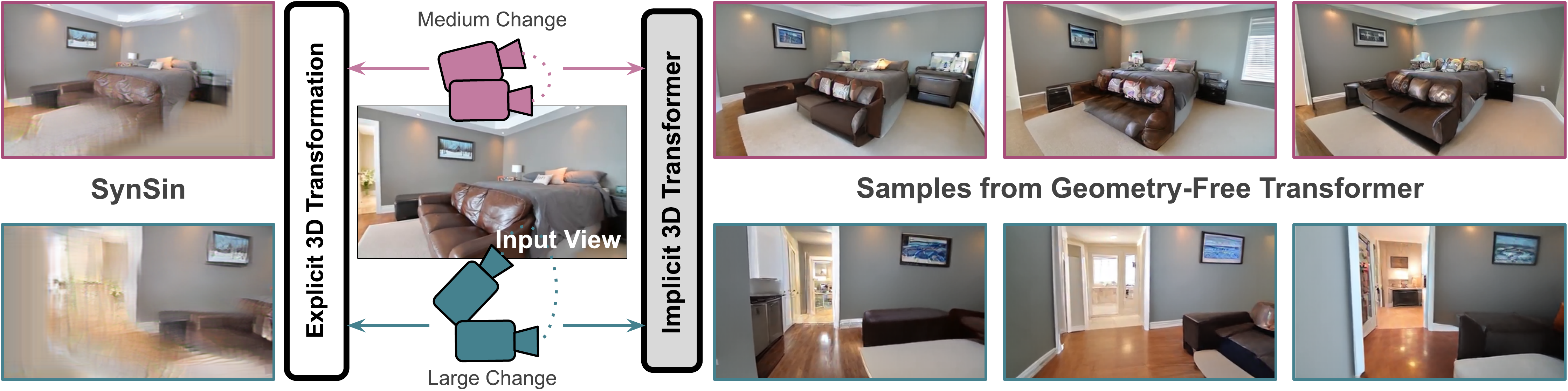}
      \captionof{figure}{We present a probabilistic approach to Novel View Synthesis based on transformers, which does not require explicit 3D priors. Given a single source frame and a camera transformation (center), we synthesize plausible novel views that exhibit high fidelity (right). For comparison, \synsin{} (left) yields uniform surfaces and unrealistic warps for large camera transformations.\vspace{-0.7em}
      }
      \label{fig:firstpagefigure}
    \end{center}
  \vspace{0.5em}
}

\newcommand{\fignllacid}        {
   \begin{figure}
   \begin{center}
   \includegraphics[width=0.975\linewidth]{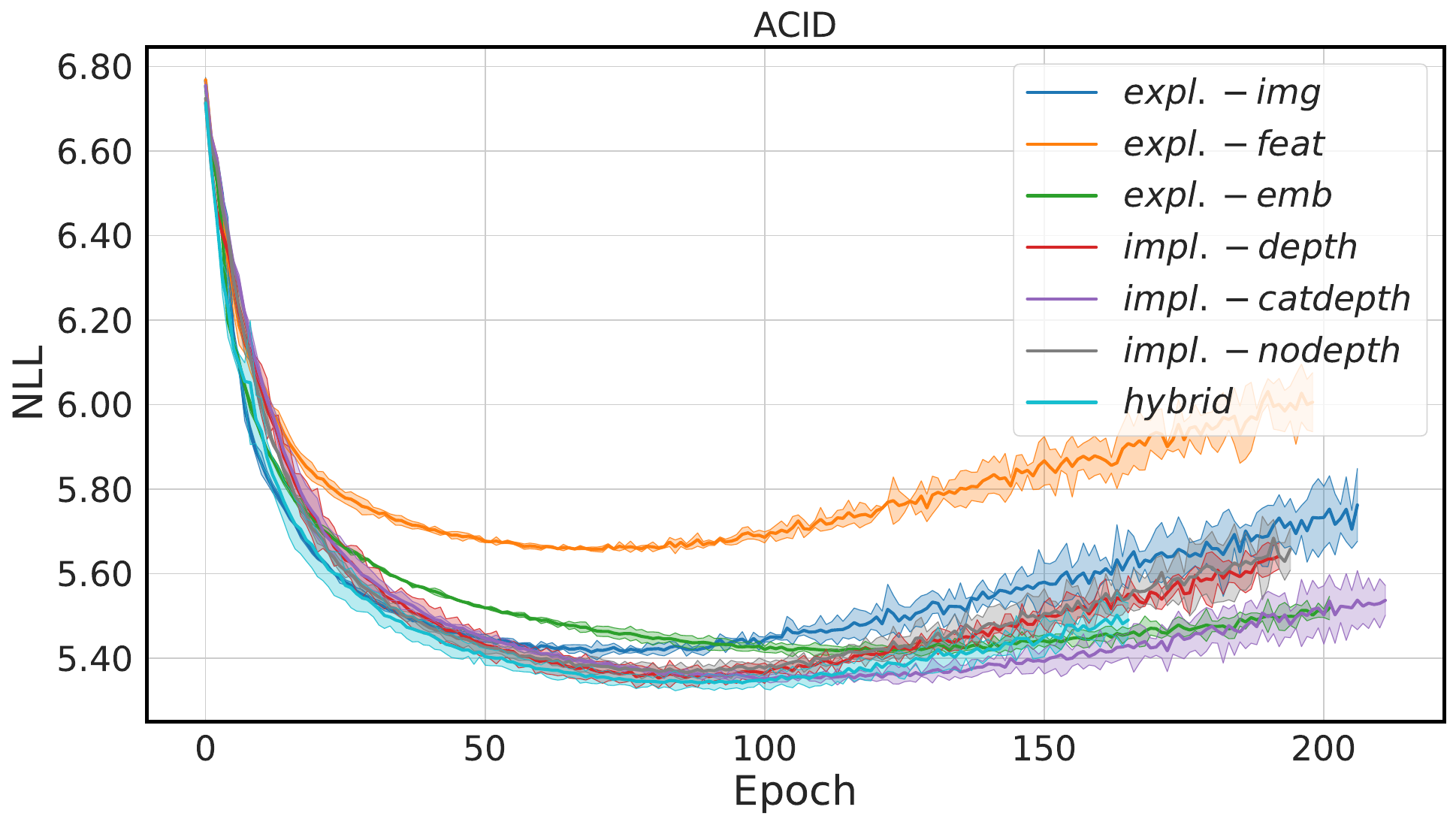}
   \end{center}
      \caption{
NLL vs epochs on ACID, evaluated on \todo{CustomTest}}
   \label{fig:nllacid}
   \end{figure}
}

\newcommand{\fignllreal}        {
   \begin{figure}
   \begin{center}
   \includegraphics[width=0.975\linewidth]{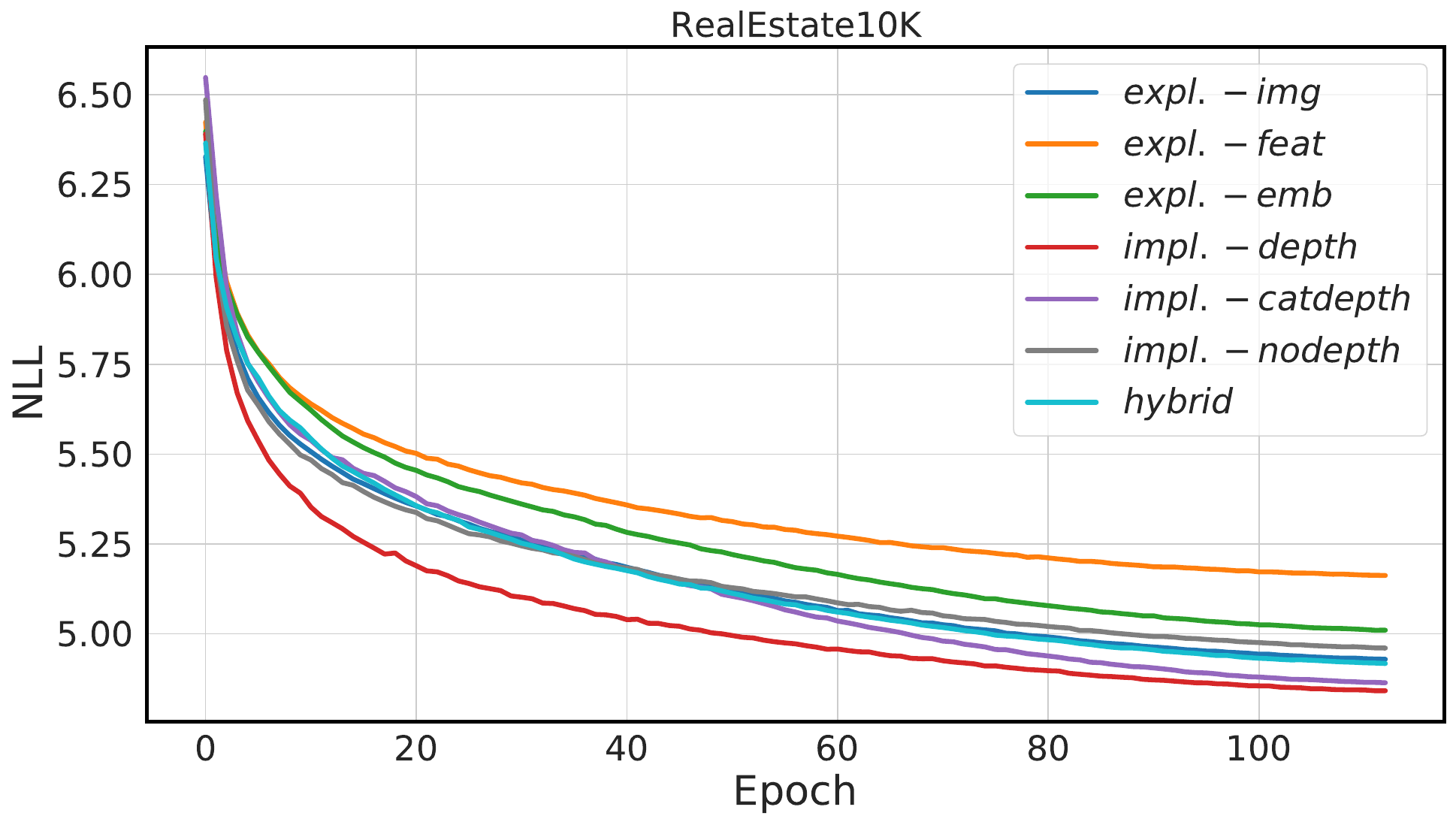}
   \end{center}
      \caption{
NLL vs epochs on REAL, evaluated on \todo{CustomTest}}
   \label{fig:nllreal}
   \end{figure}
}

\newcommand{\fignllrealandacid}        {
   \begin{figure}[thbp]
   \begin{center}
   \includegraphics[width=0.4975\linewidth]{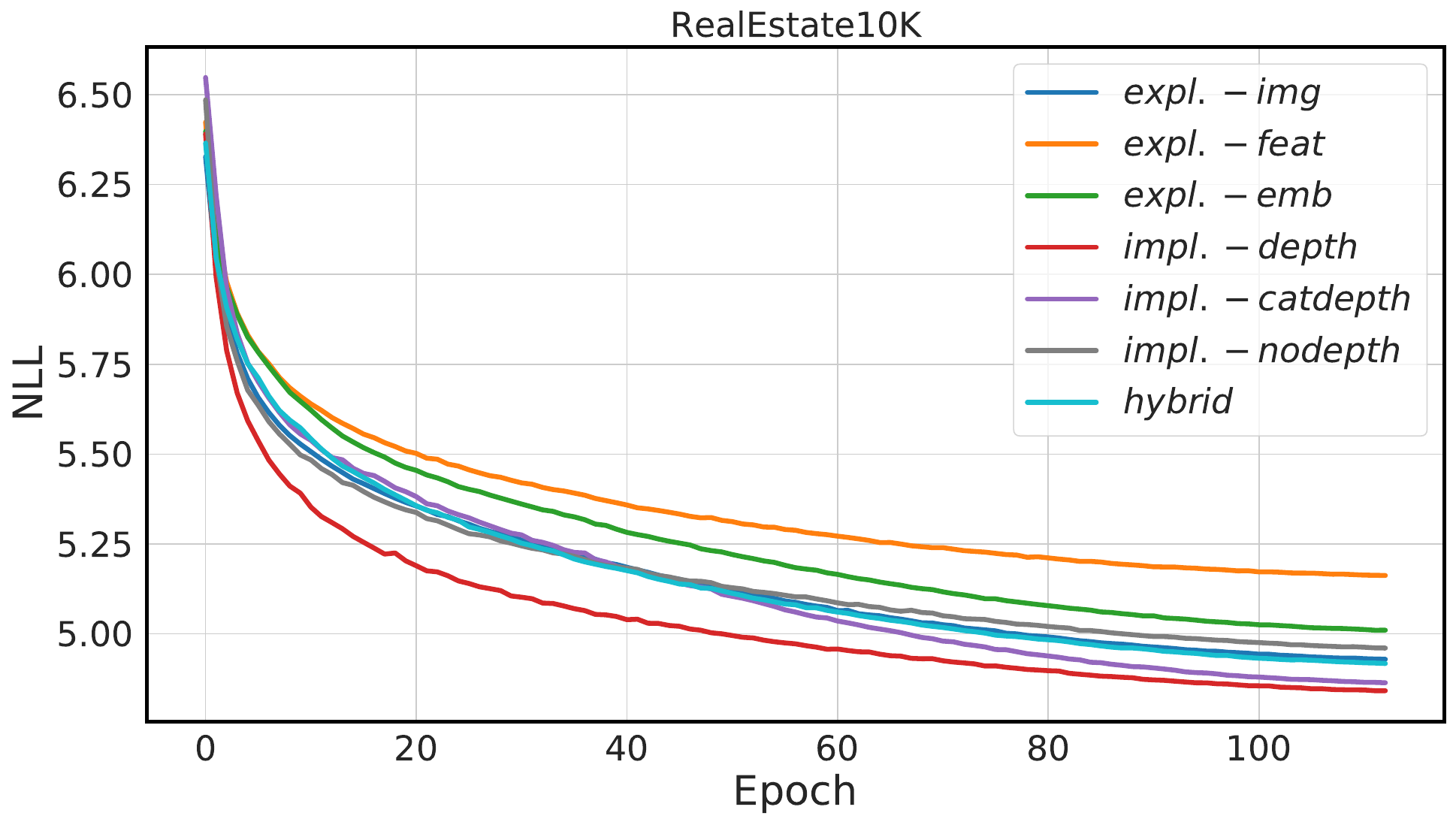}
   \includegraphics[width=0.4975\linewidth]{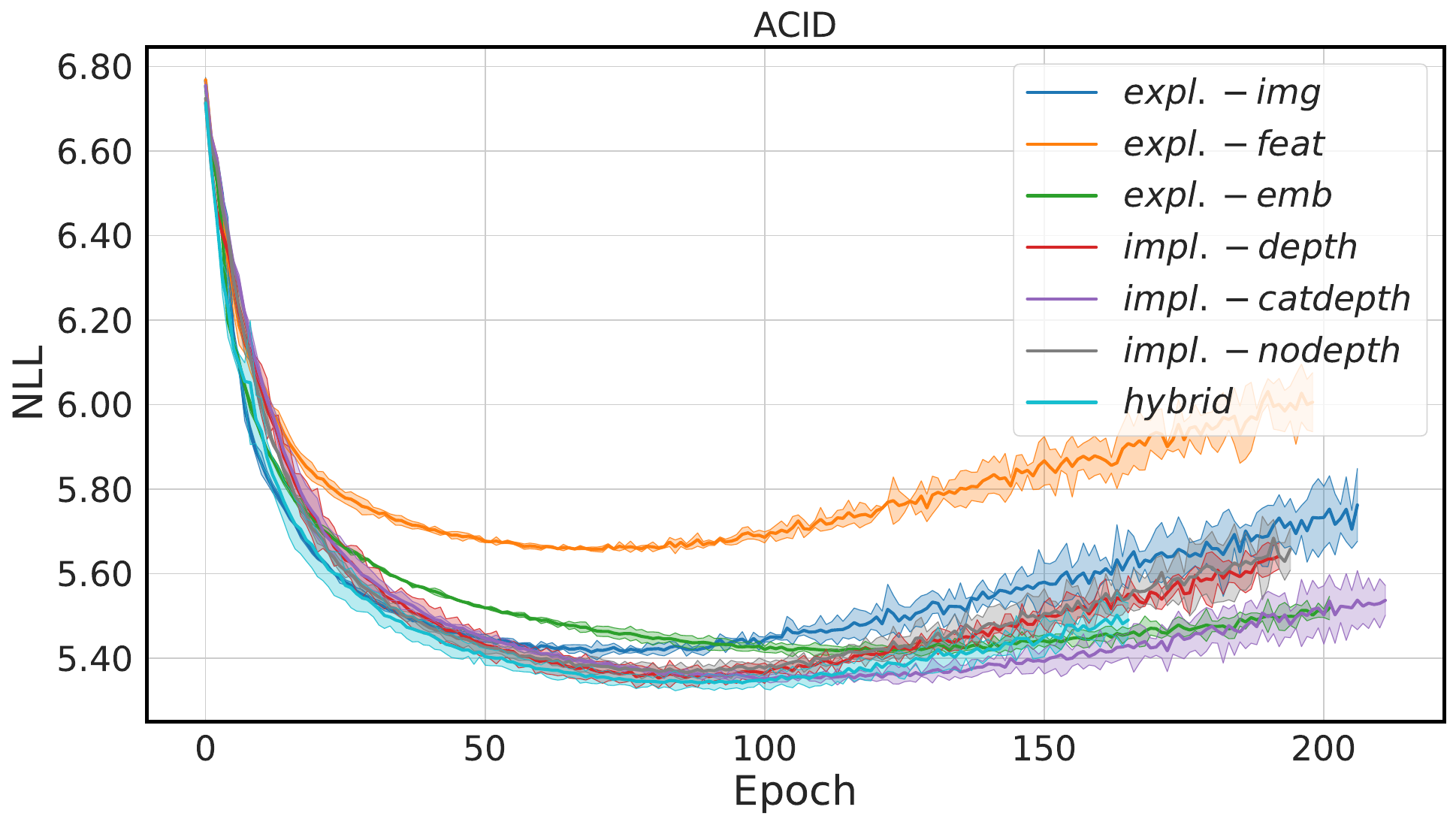}
   \end{center}
      \caption{
        Negative log-likelihood over the course of training on RealEstate10K (left) and ACID (right). Implicit
        variants achieve the best results, see Sec.~\ref{subsec:implicitvsexplicit} for a discussion.}
   \label{fig:nllrealandacid}
   \end{figure}
}

\newcommand{\figreconstructionnodepth}        {
   \begin{figure*}
   \begin{center}
   \includegraphics[width=0.32\textwidth]{img/ours_custom_nodepth_psim}
      \includegraphics[width=0.32\textwidth]{img/ours_custom_nodepth_ssim}
         \includegraphics[width=0.32\textwidth]{img/ours_custom_nodepth_psnr}
   \end{center}
      \caption{
Reconstruction out of k, no-depth}
   \label{fig:reconstructionnodepth}
   \end{figure*}
}

\newcommand{\figreconstructionbothdepth}        {
   \begin{figure*}
   \begin{center}
   \includegraphics[width=0.33\textwidth]{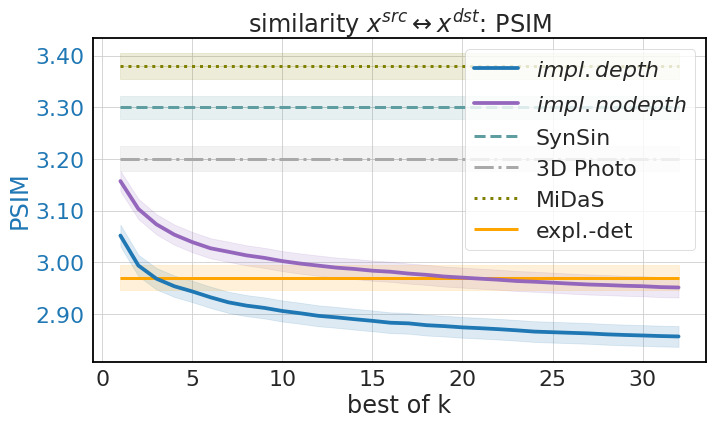}
      \includegraphics[width=0.33\textwidth]{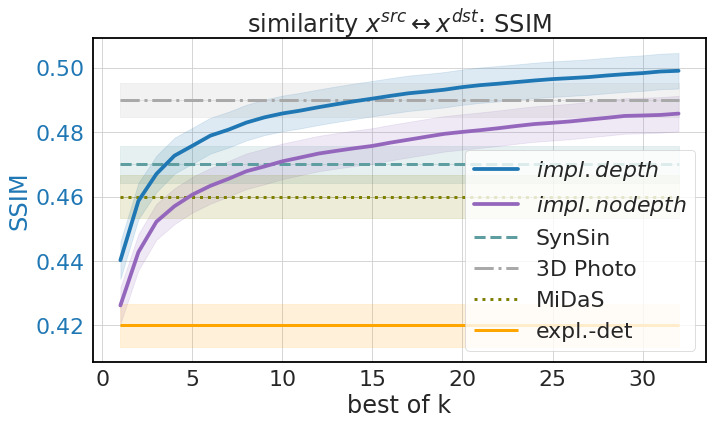}
         \includegraphics[width=0.33\textwidth]{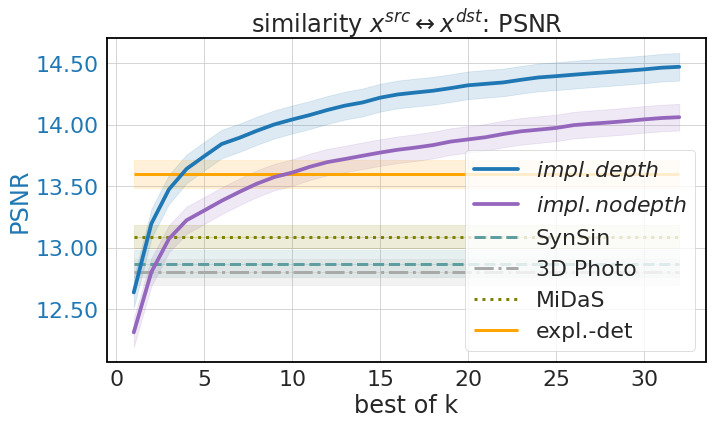}
   \end{center}
   \vspace{-1.5em}
      \caption{
        Average reconstruction error of the best sample as a function of the
        number of samples on RealEstate. With just four
        samples, \varv{} reaches state-of-the-art performance in two out of three metrics,
        and with 16 samples in all three of them.\vspace{-0.5em}
     }
   \label{fig:reconstructionbothdepth}
   \end{figure*}
}

\newcommand{\figrealestate}        {
   \begin{figure*}[bthp]
   \begin{center}
     \resizebox{\linewidth}{!}{%
       \setlength{\tabcolsep}{0.05em}%
\begin{tabular}{@{}cc@{\hskip 0.75em}ccc@{\hskip 0.75em}cc@{}}%
\toprule%
  Source&Target&\threedp{}&\synsin{}&\vqwarper{}&\varv{}&\varvi{}\\%
\midrule%
\includegraphics[width=0.14\linewidth,height=0.08\linewidth]{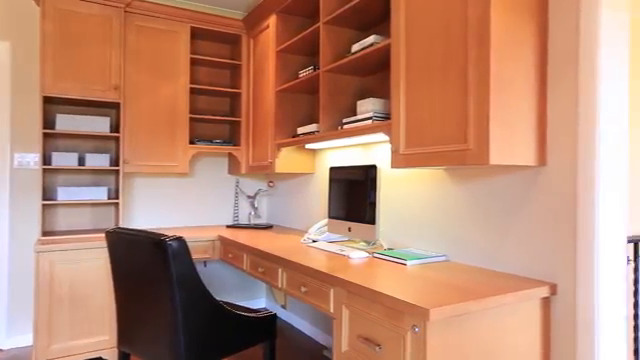}&\includegraphics[width=0.14\linewidth,height=0.08\linewidth]{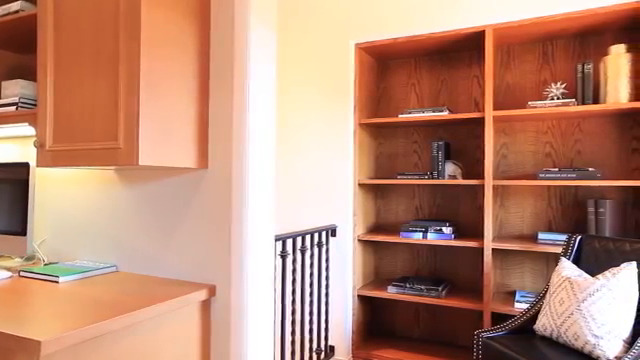}&\includegraphics[width=0.14\linewidth,height=0.08\linewidth]{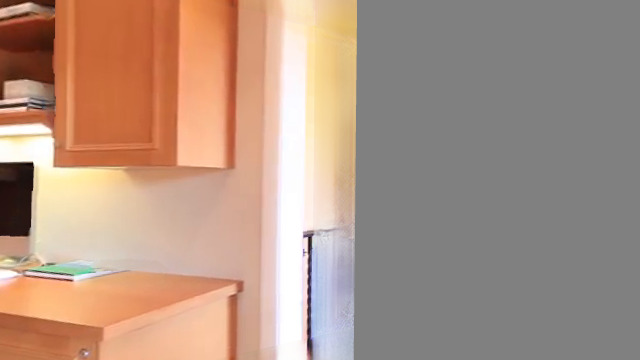}&\includegraphics[width=0.14\linewidth,height=0.08\linewidth]{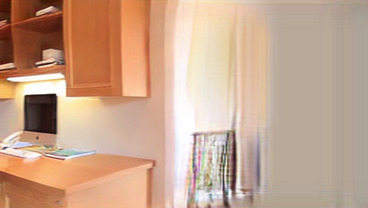}&\includegraphics[width=0.14\linewidth,height=0.08\linewidth]{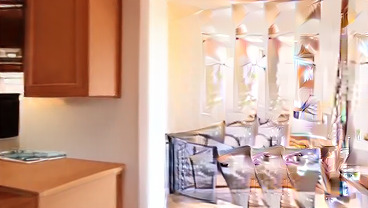}&\includegraphics[width=0.14\linewidth,height=0.08\linewidth]{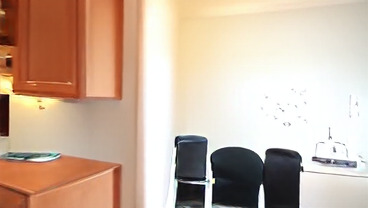}&\includegraphics[width=0.14\linewidth,height=0.08\linewidth]{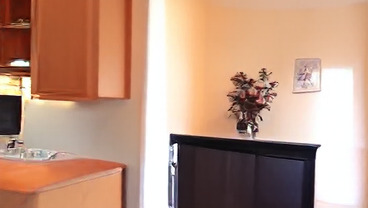}\\%
\includegraphics[width=0.14\linewidth,height=0.08\linewidth]{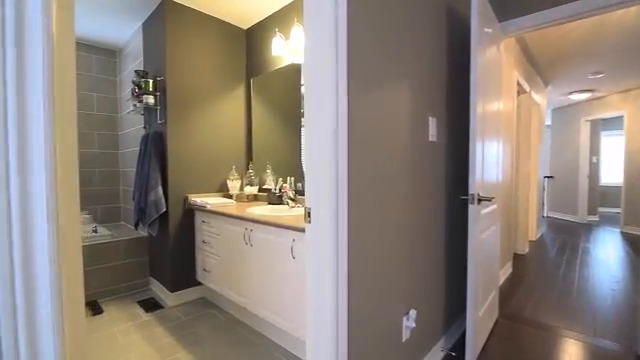}&\includegraphics[width=0.14\linewidth,height=0.08\linewidth]{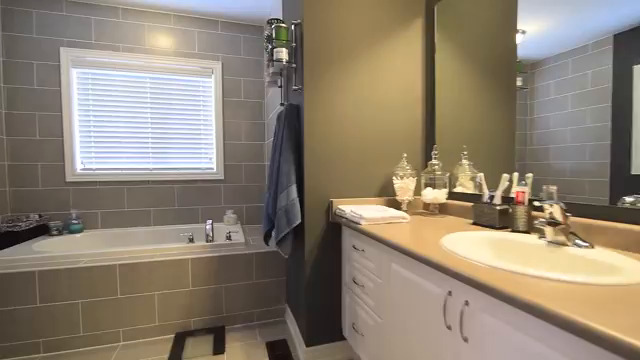}&\includegraphics[width=0.14\linewidth,height=0.08\linewidth]{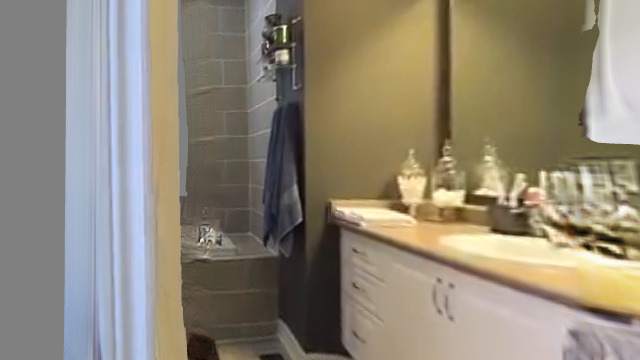}&\includegraphics[width=0.14\linewidth,height=0.08\linewidth]{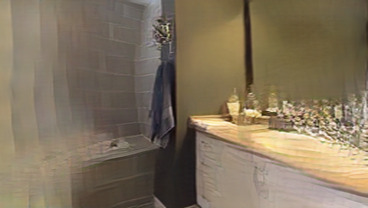}&\includegraphics[width=0.14\linewidth,height=0.08\linewidth]{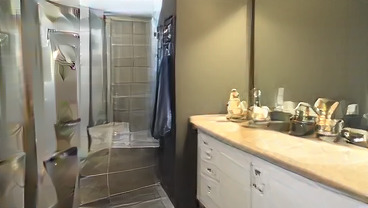}&\includegraphics[width=0.14\linewidth,height=0.08\linewidth]{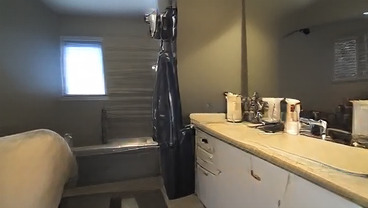}&\includegraphics[width=0.14\linewidth,height=0.08\linewidth]{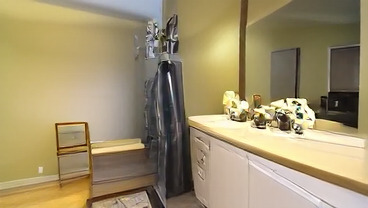}\\%
\includegraphics[width=0.14\linewidth,height=0.08\linewidth]{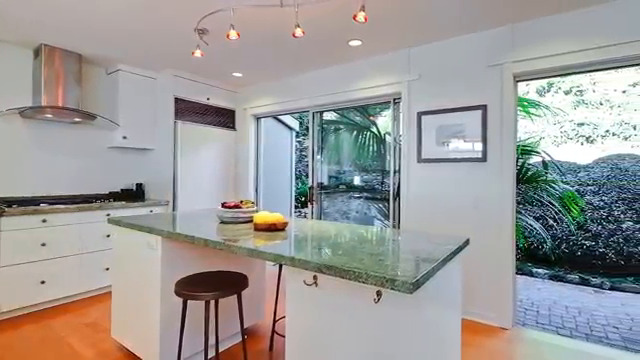}&\includegraphics[width=0.14\linewidth,height=0.08\linewidth]{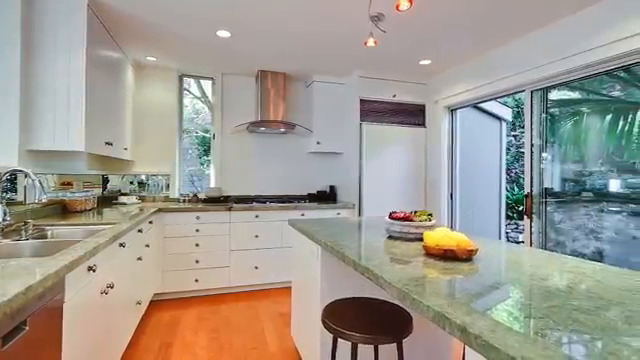}&\includegraphics[width=0.14\linewidth,height=0.08\linewidth]{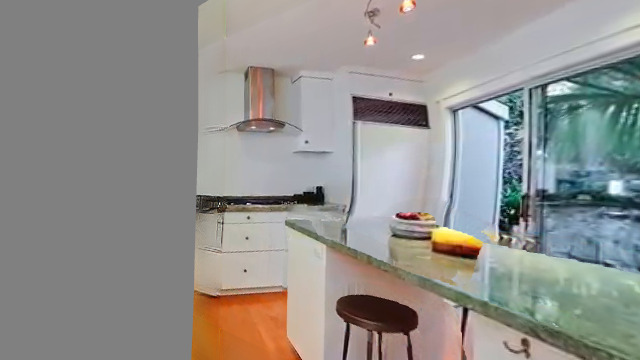}&\includegraphics[width=0.14\linewidth,height=0.08\linewidth]{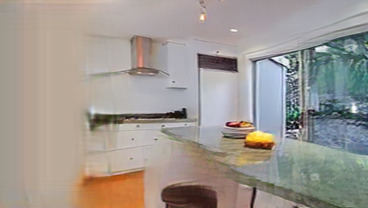}&\includegraphics[width=0.14\linewidth,height=0.08\linewidth]{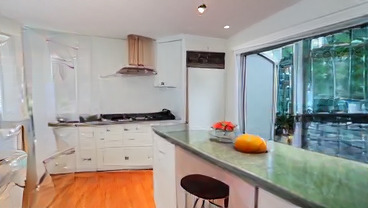}&\includegraphics[width=0.14\linewidth,height=0.08\linewidth]{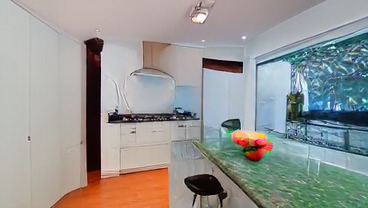}&\includegraphics[width=0.14\linewidth,height=0.08\linewidth]{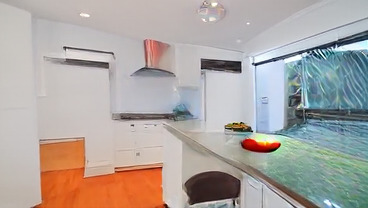}\\%
\end{tabular}
}
   \end{center}
     \vspace{-1.00em}
      \caption{\textbf{Qualitative Results on RealEstate10K:} 
We compare three deterministic convolutional baselines (\threedp{}, \synsin{},
     \vqwarper{}) to our implicit variants \varv{} and \varvi{}. Ours is able
     to synthesize plausible novel views, whereas others
     produce artifacts or blurred, uniform areas. The depicted
     target is only one of many possible realizations; we visualize samples
     in the supplement.       \vspace{-1.0em}
     }
   \label{fig:realestate}
   \end{figure*}
}

\newcommand{\figacid}        {
   \begin{figure*}
   \begin{center}
     \resizebox{\linewidth}{!}{%
       \setlength{\tabcolsep}{0.05em}%
\begin{tabular}{@{}cc@{\hskip 0.75em}ccc@{\hskip 0.75em}cc@{}}%
\toprule%
  Source&Target&\threedp{}&\infnat{}&\vqwarper{}&\varv{}&\varvi{}\\%
\midrule%
\includegraphics[width=0.14\linewidth,height=0.08\linewidth]{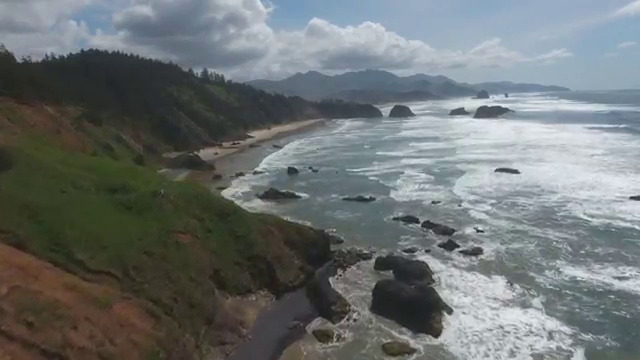}&\includegraphics[width=0.14\linewidth,height=0.08\linewidth]{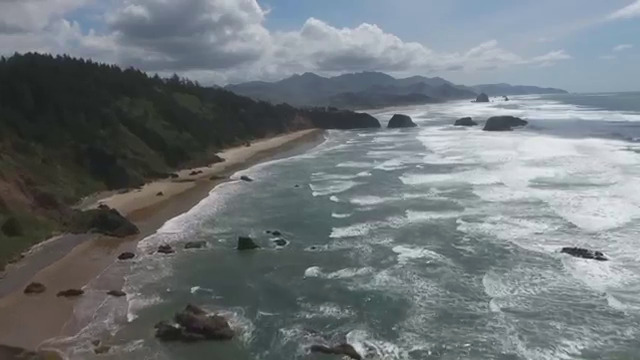}&\includegraphics[width=0.14\linewidth,height=0.08\linewidth]{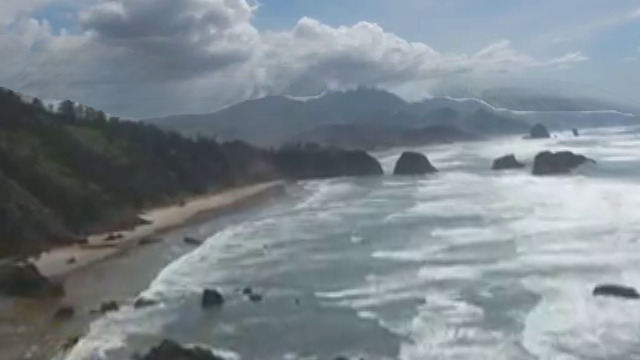}&\includegraphics[width=0.14\linewidth,height=0.08\linewidth]{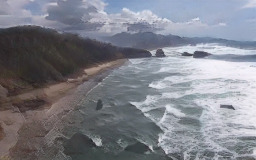}&\includegraphics[width=0.14\linewidth,height=0.08\linewidth]{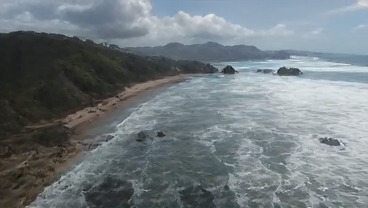}&\includegraphics[width=0.14\linewidth,height=0.08\linewidth]{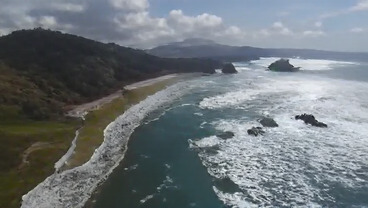}&\includegraphics[width=0.14\linewidth,height=0.08\linewidth]{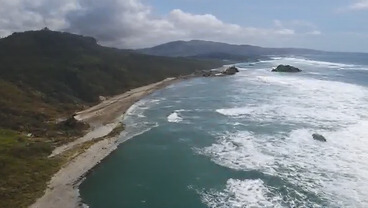}\\%
\includegraphics[width=0.14\linewidth,height=0.08\linewidth]{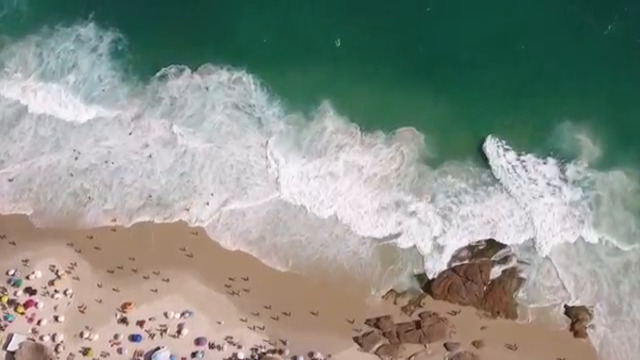}&\includegraphics[width=0.14\linewidth,height=0.08\linewidth]{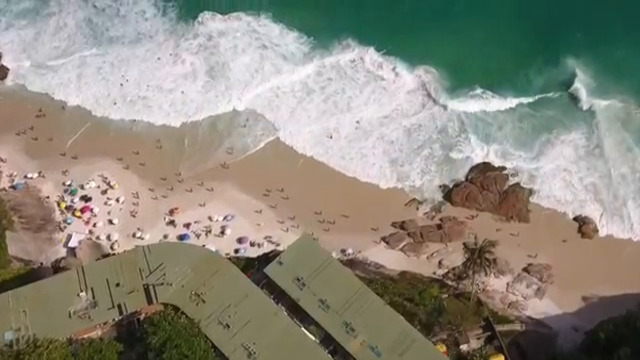}&\includegraphics[width=0.14\linewidth,height=0.08\linewidth]{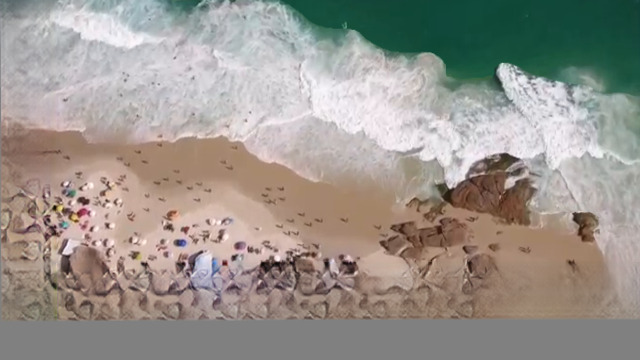}&\includegraphics[width=0.14\linewidth,height=0.08\linewidth]{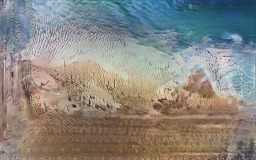}&\includegraphics[width=0.14\linewidth,height=0.08\linewidth]{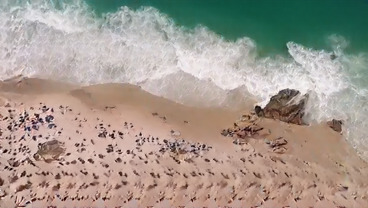}&\includegraphics[width=0.14\linewidth,height=0.08\linewidth]{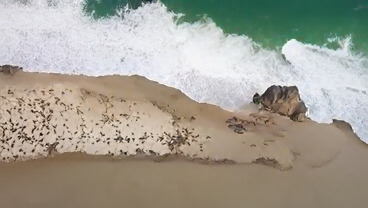}&\includegraphics[width=0.14\linewidth,height=0.08\linewidth]{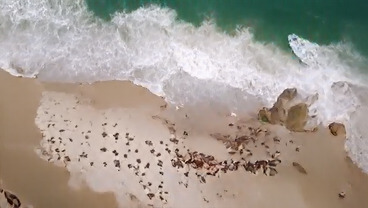}\\%
\includegraphics[width=0.14\linewidth,height=0.08\linewidth]{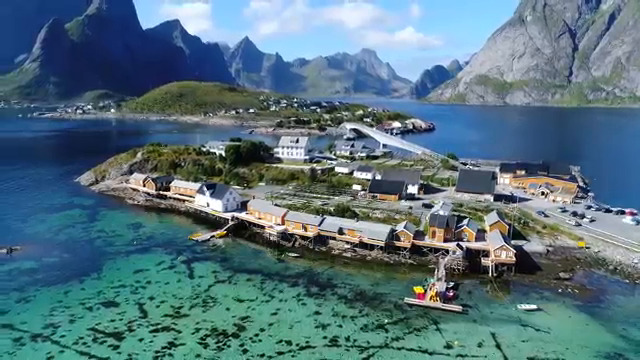}&\includegraphics[width=0.14\linewidth,height=0.08\linewidth]{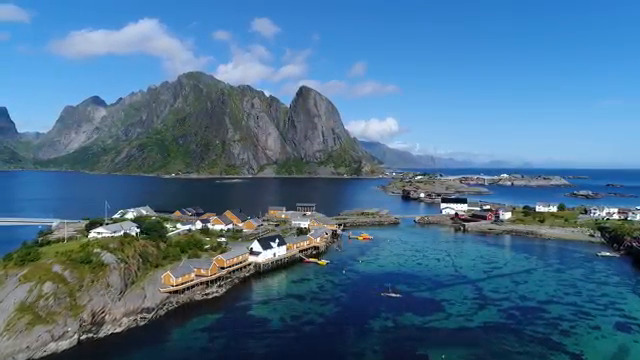}&\includegraphics[width=0.14\linewidth,height=0.08\linewidth]{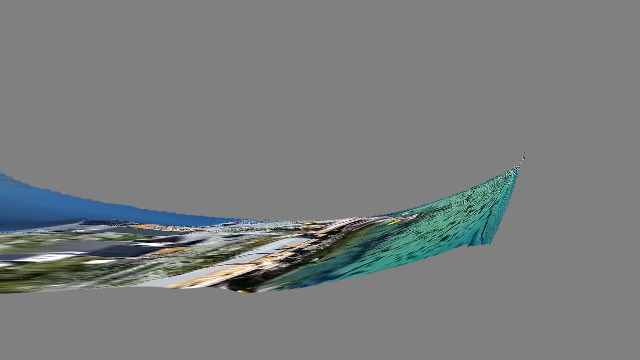}&\includegraphics[width=0.14\linewidth,height=0.08\linewidth]{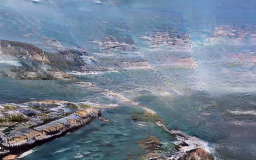}&\includegraphics[width=0.14\linewidth,height=0.08\linewidth]{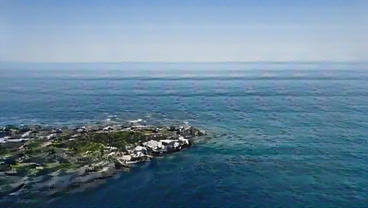}&\includegraphics[width=0.14\linewidth,height=0.08\linewidth]{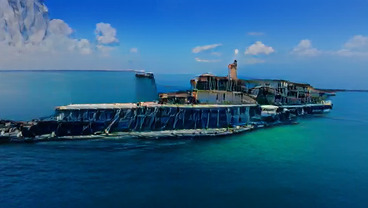}&\includegraphics[width=0.14\linewidth,height=0.08\linewidth]{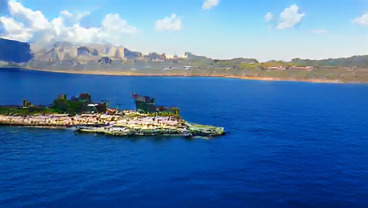}%
\end{tabular}
}
   \end{center}
     \vspace{-1.00em}
      \caption{
        \textbf{Qualitative Results on ACID:} 
The outdoor setting of the ACID dataset yields similar results as the indoor setting in Fig.~\ref{fig:realestate}.
Here, we evaluate against the baselines \threedp{}, \infnat{} and \vqwarper{}. For \infnat{}, we use 5 steps
to synthesize a novel view. 
}   
  
   \label{fig:acid}
     \vspace{-1em}
   \end{figure*}
}

\newcommand{\figdepthprobesqualitative}{
\begin{figure}[t]
  \renewcommand{\impath}[1]{img/depthprobes/img/layer0/src_image/##1}
  \renewcommand{\impatha}[1]{img/depthprobes/img/layer0/depth_prediction/##1}
  \renewcommand{\impathb}[1]{img/depthprobes/img/layer1/depth_prediction/##1}
  \renewcommand{\impathc}[1]{img/depthprobes/img/layer4/depth_prediction/##1}
  \renewcommand{\impathd}[1]{img/depthprobes/img/layer4/depth_reconstruction/##1}
  \renewcommand{\imwidth}{0.0975\textwidth}
  \setlength{\tabcolsep}{0.1pt}
\centering
\begin{tabular}{c ccc c}
\footnotesize{Input} & \footnotesize{Layer \#0} & \footnotesize{Layer \#1} & \footnotesize{\textbf{Layer \#4}} & \footnotesize{Depth Rec.} \\
\toprule

	\includegraphics[width=\imwidth, align=c]{\impath{125_src_image_000763}} &	         
	\includegraphics[width=\imwidth, align=c]{\impatha{125_depth_prediction_000763}} &	         
	\includegraphics[width=\imwidth, align=c]{\impathb{125_depth_prediction_000763}} &	         
	\includegraphics[width=\imwidth, align=c]{\impathc{125_depth_prediction_000763}} &	         
	\includegraphics[width=\imwidth, align=c]{\impathd{125_depth_reconstruction_000763}} \\	
	
	\includegraphics[width=\imwidth, align=c]{\impath{132_src_image_000478}} &	         
	\includegraphics[width=\imwidth, align=c]{\impatha{132_depth_prediction_000478}} &	         
	\includegraphics[width=\imwidth, align=c]{\impathb{132_depth_prediction_000478}} &	         
	\includegraphics[width=\imwidth, align=c]{\impathc{132_depth_prediction_000478}} &	         
	\includegraphics[width=\imwidth, align=c]{\impathd{132_depth_reconstruction_000478}} \\	

	\includegraphics[width=\imwidth, align=c]{\impath{155_src_image_000058}} &	         
	\includegraphics[width=\imwidth, align=c]{\impatha{155_depth_prediction_000058}} &	         
	\includegraphics[width=\imwidth, align=c]{\impathb{155_depth_prediction_000058}} &	         
	\includegraphics[width=\imwidth, align=c]{\impathc{155_depth_prediction_000058}} &	         
	\includegraphics[width=\imwidth, align=c]{\impathd{155_depth_reconstruction_000058}} \\

\end{tabular}
\vspace{0.2em}
\caption{
  Linearly probed depth maps for different transformer layers.
  The results mirror the curve in Fig.~\ref{fig:depthprobinglayers}:
After a strong initial increase, the quality for layer 4 is best.
The depth reconstructions in the right column
provide an
upper bound on achievable quality.
}
\label{fig:figdepthprobesqualitative}
 \vspace{-1.0em}
\end{figure}
}

\newcommand{\figentropy}{
\begin{figure}[b!]
  \vspace{-1em}
  \renewcommand{\impath}[1]{img/entropy/nodepth/##1}
  \renewcommand{\imwidth}{0.16\textwidth}
  \setlength{\tabcolsep}{0.1pt}
\centering
\begin{tabular}{c c c}
\footnotesize{$\xsrc$} & \footnotesize{$\xdst$} & \footnotesize{$\transformer$'s entropy} \\
\toprule
	\includegraphics[width=\imwidth, align=c]{\impath{15_conditioning_000046}} &	         
	\includegraphics[width=\imwidth, align=c]{\impath{15_inputs_000046}} &	         
	\includegraphics[width=\imwidth, align=c]{\impath{15_spatial_entropy_data_upsampled}} \\
	
	\includegraphics[width=\imwidth, align=c]{\impath{542_conditioning_000033}} &	         
	\includegraphics[width=\imwidth, align=c]{\impath{542_inputs_000033}} &	         
	\includegraphics[width=\imwidth, align=c]{\impath{542_spatial_entropy_data_upsampled}} \\

\end{tabular}
\caption{
  Visualization of the entropy of the predicted target code distribution for
  \varvi{}.
  Increased confidence (darker colors) in regions which are visible in the source image
  indicate its ability to relate source and target geometrically, without
  3D bias.
}
\label{fig:entropy}
\end{figure}
}

\newcommand{\figentropysupp}{
\begin{figure*}[thbp]
  \vspace{-1em}
  \renewcommand{\impath}[1]{img/entropy/nodepth/##1}
  \renewcommand{\imwidth}{0.155\textwidth}
  \setlength{\tabcolsep}{0.1pt}
\centering
\begin{tabular}{c c c  c c c}
\footnotesize{$\xsrc$} & \footnotesize{$\xdst$} & \footnotesize{$\transformer$'s entropy} &
\footnotesize{$\xsrc$} & \footnotesize{$\xdst$} & \footnotesize{$\transformer$'s entropy} \\
\toprule
	\includegraphics[width=\imwidth, align=c]{\impath{254_conditioning_000010}} &	         
	\includegraphics[width=\imwidth, align=c]{\impath{254_inputs_000010}} &	         
	\includegraphics[width=\imwidth, align=c]{\impath{254_spatial_entropy_data_upsampled}} &
	
	\includegraphics[width=\imwidth, align=c]{\impath{209_conditioning_000029}} &	         
	\includegraphics[width=\imwidth, align=c]{\impath{209_inputs_000029}} &	         
	\includegraphics[width=\imwidth, align=c]{\impath{209_spatial_entropy_data_upsampled}} \\
	
	\includegraphics[width=\imwidth, align=c]{\impath{357_conditioning_000011}} &	         
	\includegraphics[width=\imwidth, align=c]{\impath{357_inputs_000011}} &	         
	\includegraphics[width=\imwidth, align=c]{\impath{357_spatial_entropy_data_upsampled}} & 
	
	\includegraphics[width=\imwidth, align=c]{\impath{431_conditioning_000031}} &	         
	\includegraphics[width=\imwidth, align=c]{\impath{431_inputs_000031}} &	         
	\includegraphics[width=\imwidth, align=c]{\impath{431_spatial_entropy_data_upsampled}} \\
	
	\includegraphics[width=\imwidth, align=c]{\impath{456_conditioning_000032}} &	         
	\includegraphics[width=\imwidth, align=c]{\impath{456_inputs_000032}} &	         
	\includegraphics[width=\imwidth, align=c]{\impath{456_spatial_entropy_data_upsampled}} & 
	
	\includegraphics[width=\imwidth, align=c]{\impath{506_conditioning_000592}} &	         
	\includegraphics[width=\imwidth, align=c]{\impath{506_inputs_000592}} &	         
	\includegraphics[width=\imwidth, align=c]{\impath{506_spatial_entropy_data_upsampled}} \\
	
\end{tabular}
\caption{
  Additional visualizations of the entropy of the predicted target code distribution for
  \varvi{}.
  Increased confidence (darker colors) in regions which are visible in the source image
  indicate its ability to relate source and target geometrically, without
  3D bias. See also Sec.~\ref{subsec:implicitvsexplicit} and Sec.~\ref{sec:entropydetails}.
}
\label{fig:entropysupp}
\end{figure*}
}

\newcommand{\figapproach}        {
   \begin{figure}[t]
   \begin{center}
   \includegraphics[width=0.995\linewidth]{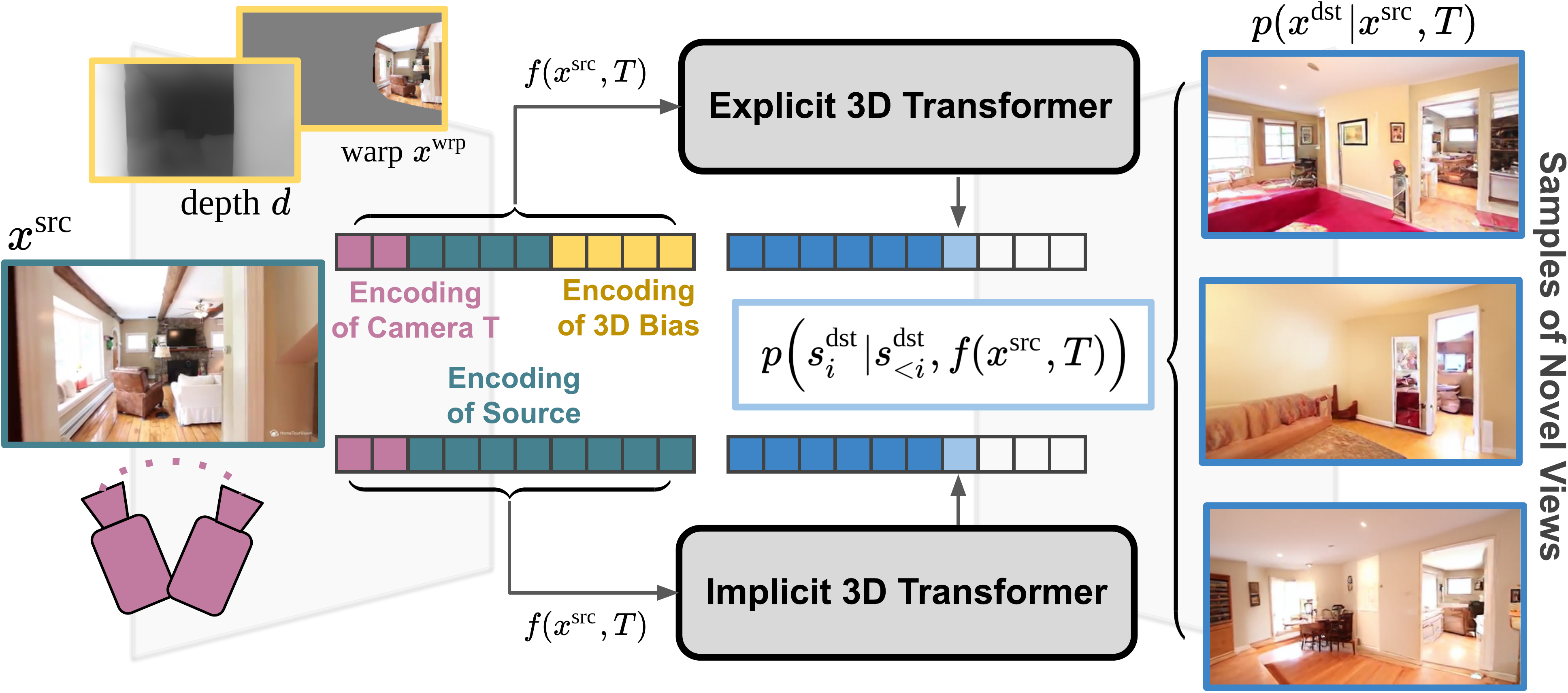}
   \end{center}
      \caption{
\todo{our method. add point cloud, make sure depth corresponds to scene. remove arrow. cam is $\cam$?}}
   \label{fig:approach}
   \end{figure}
}

\newcommand{\figapproachmerged}        {
   \begin{figure*}[t]
   \begin{center}
   \resizebox{.995\textwidth}{!}{%
     \raisebox{-0.5\height}{%
   \includegraphics[width=0.55\linewidth]{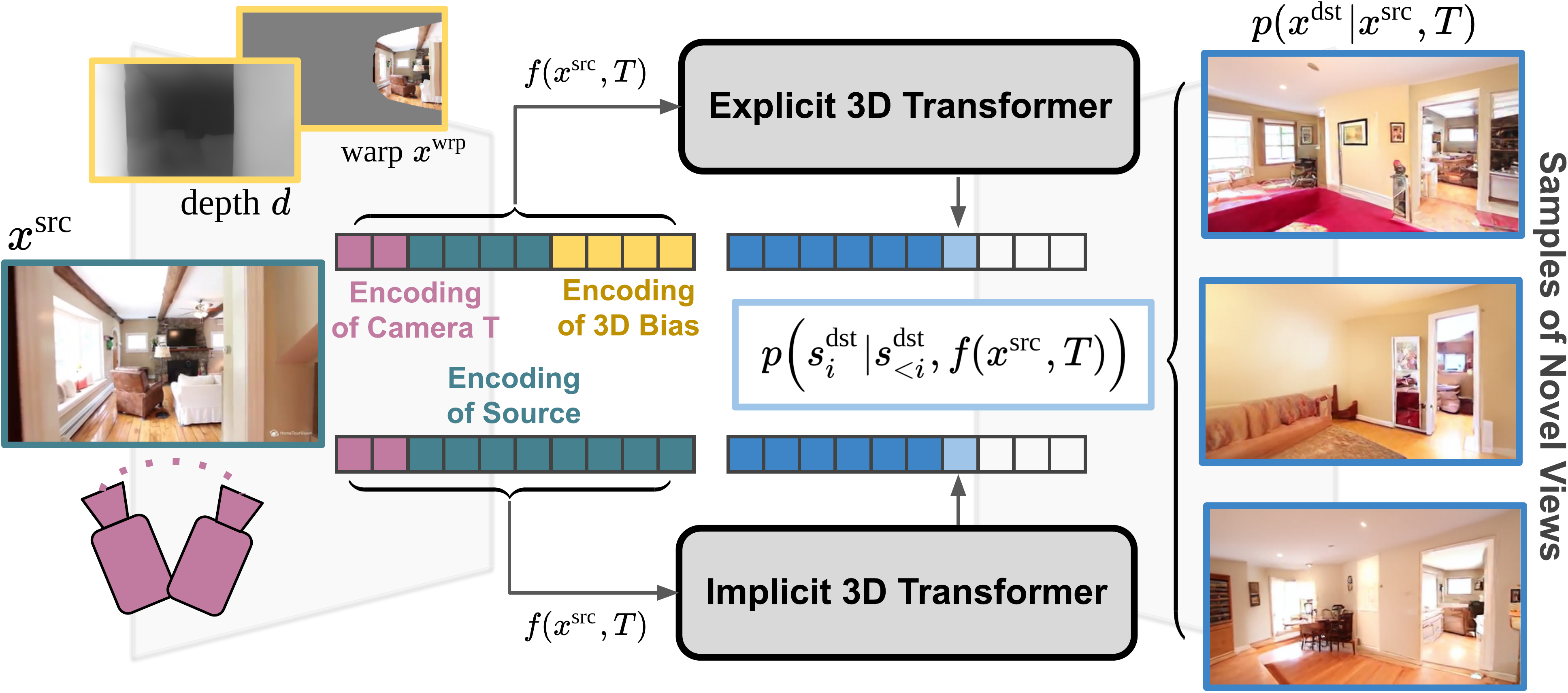}
     }
   \begin{footnotesize}
     \setlength{\tabcolsep}{0.30em}
\begin{tabular}{l c c c c}
\toprule
variant & \makecell{explicit \\ warp} & \makecell{requires \\ depth} & \makecell{warped \\ features} & $f(\xsrc, \cam)$ \\
\midrule
\vari{} & \cmark & \cmark & $\xsrc$ & Eq.\eqref{eq:fvari} \\
\varii{} & \cmark & \cmark & $\encoder(\xsrc)$ & Eq.\eqref{eq:fvarii} \\
\variii{} & \cmark & \cmark & $e(\encoder(\xsrc)),\,e^{\text{pos}}$ & Eq.\eqref{eq:fvariii} \\
\midrule
\variv{} & \xmark & \cmark & -- & Eq.\eqref{eq:fvariv} \\
\varv{} & \xmark & \cmark & -- & Eq.\eqref{eq:fvarv} \\
\varvi{} & \xmark & \xmark & -- & Eq.\eqref{eq:fvarvi} \\
\midrule
\varvii{} & \cmark & \cmark & $e(\encoder(\xsrc)),\,e^{\text{pos}}$ & Eq.\eqref{eq:fvariii}+Eq.\eqref{eq:fvarv} \\
\bottomrule
\end{tabular}%
\end{footnotesize}
}
   \end{center}
      \caption{
        We formulate novel view synthesis as sampling from the distribution
        $p(\xdst \vert \xsrc, \cam)$ of
        target images $\xdst$ for a given source image $\xsrc$ and
        camera change $\cam$.
        We use a VQGAN to model this distribution autoregressively with a
        transformer %
        and introduce a conditioning function $f(\xsrc, \cam)$ to encode
        inductive biases into our model. We analyze explicit variants, which
        estimate scene depth $d$ and warp source features into the novel
        view, as well as implicit variants without such a warping.
        The table on the right summarizes the variants for $f$.
        \vspace{-0.6em}
     }
   \label{fig:approachmerged}
   \end{figure*}
}

\newcommand{\figsupprealestate}        {
   \begin{figure*}[bthp]
   \begin{center}
     \resizebox{\linewidth}{!}{%
       \setlength{\tabcolsep}{0.1em}%
\begin{tabular}{@{}cc@{\hskip 0.75em}ccc@{\hskip 0.75em}cc@{}}%
\toprule%
Source&Target&\threedp{}&\synsin{}&\vqwarper{}&\varv{}&\varvi{}\\%
\midrule%
\includegraphics[width=0.14\linewidth,height=0.08\linewidth]{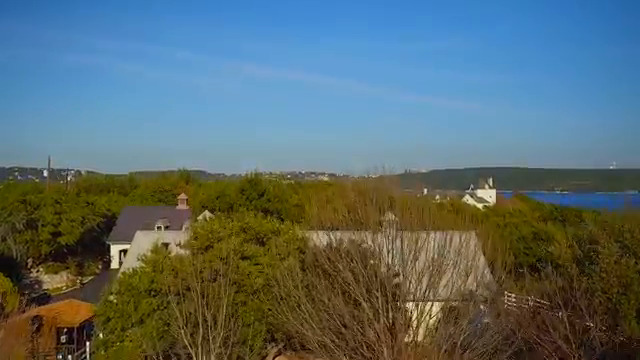}&\includegraphics[width=0.14\linewidth,height=0.08\linewidth]{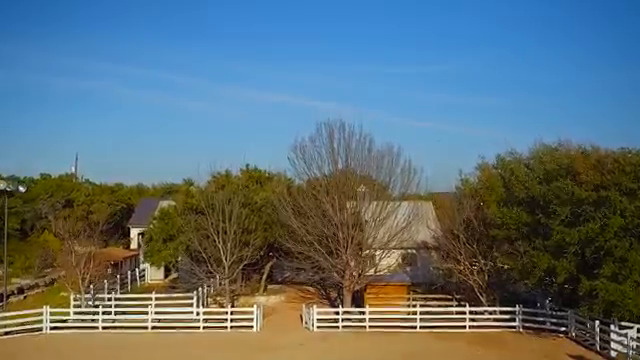}&\includegraphics[width=0.14\linewidth,height=0.08\linewidth]{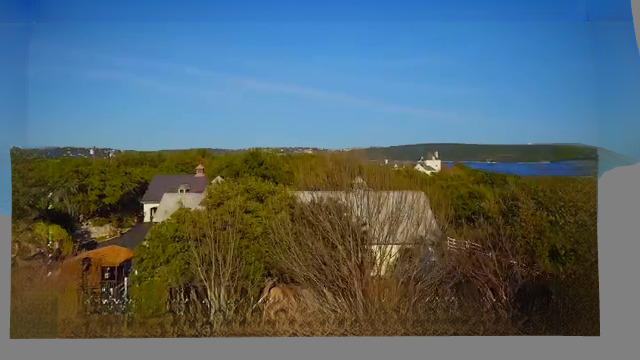}&\includegraphics[width=0.14\linewidth,height=0.08\linewidth]{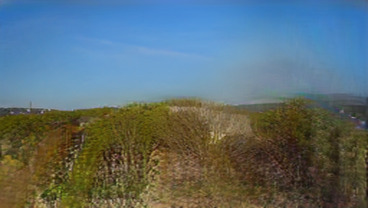}&\includegraphics[width=0.14\linewidth,height=0.08\linewidth]{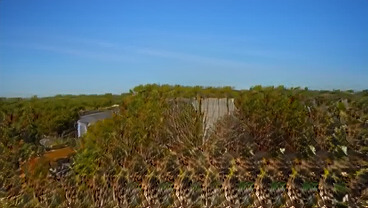}&\includegraphics[width=0.14\linewidth,height=0.08\linewidth]{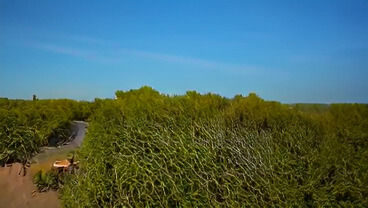}&\includegraphics[width=0.14\linewidth,height=0.08\linewidth]{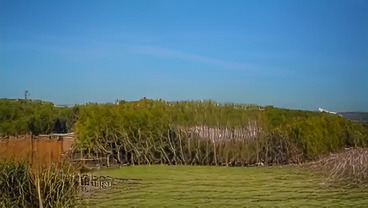}\\%
\includegraphics[width=0.14\linewidth,height=0.08\linewidth]{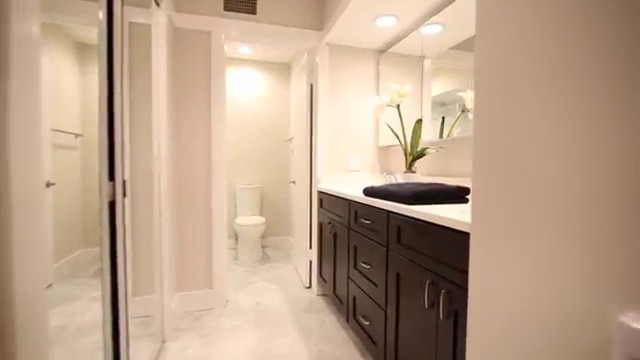}&\includegraphics[width=0.14\linewidth,height=0.08\linewidth]{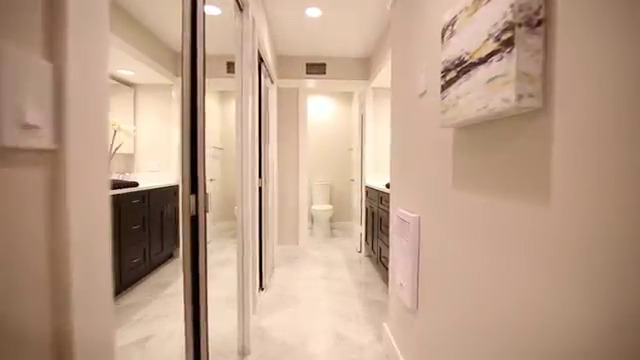}&\includegraphics[width=0.14\linewidth,height=0.08\linewidth]{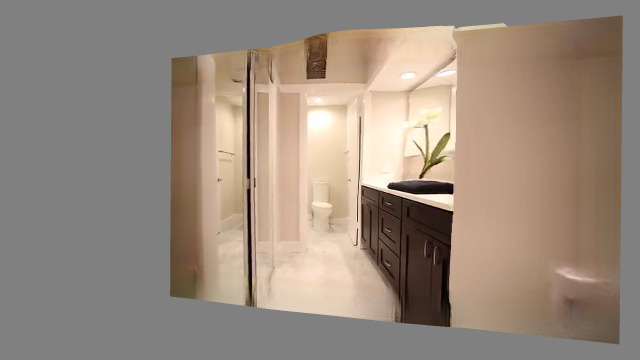}&\includegraphics[width=0.14\linewidth,height=0.08\linewidth]{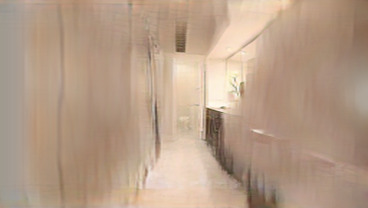}&\includegraphics[width=0.14\linewidth,height=0.08\linewidth]{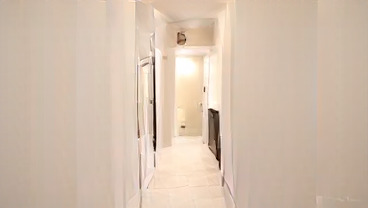}&\includegraphics[width=0.14\linewidth,height=0.08\linewidth]{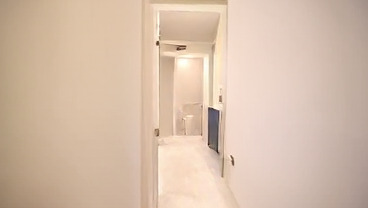}&\includegraphics[width=0.14\linewidth,height=0.08\linewidth]{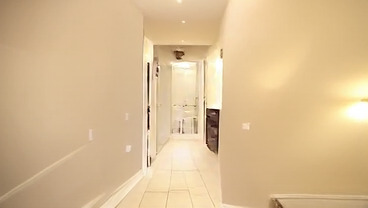}\\%
\includegraphics[width=0.14\linewidth,height=0.08\linewidth]{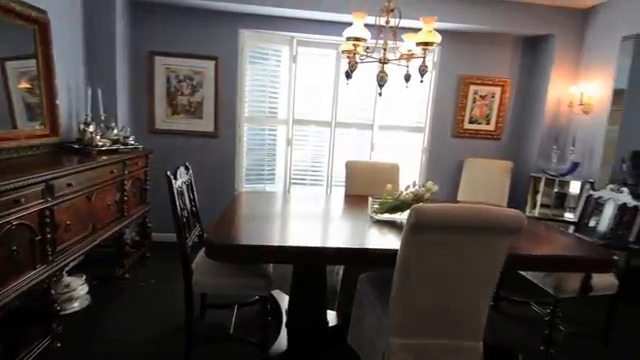}&\includegraphics[width=0.14\linewidth,height=0.08\linewidth]{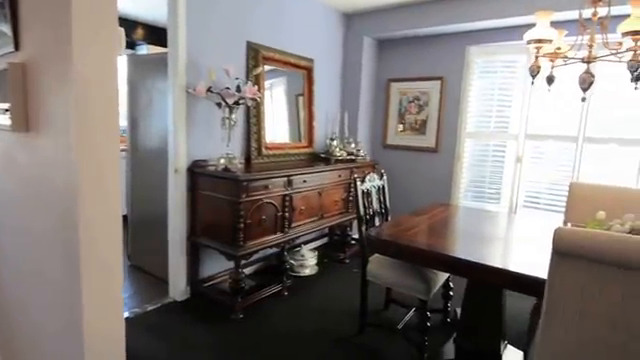}&\includegraphics[width=0.14\linewidth,height=0.08\linewidth]{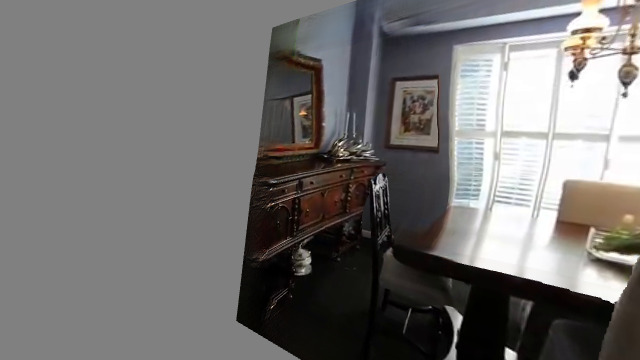}&\includegraphics[width=0.14\linewidth,height=0.08\linewidth]{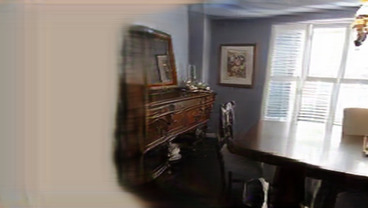}&\includegraphics[width=0.14\linewidth,height=0.08\linewidth]{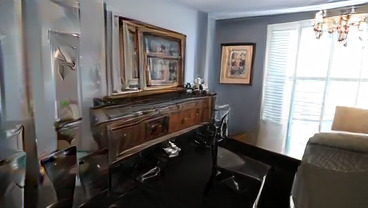}&\includegraphics[width=0.14\linewidth,height=0.08\linewidth]{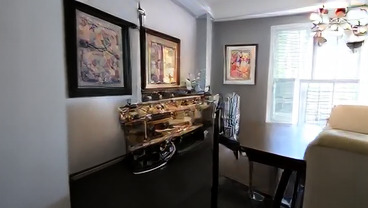}&\includegraphics[width=0.14\linewidth,height=0.08\linewidth]{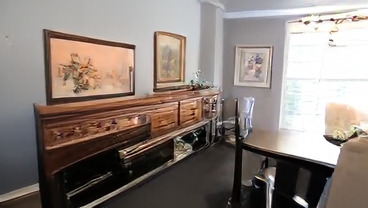}\\%
\includegraphics[width=0.14\linewidth,height=0.08\linewidth]{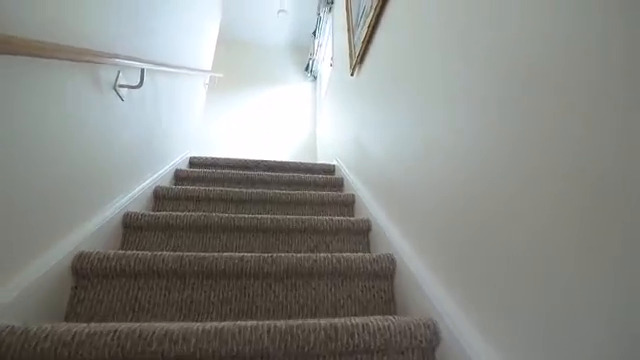}&\includegraphics[width=0.14\linewidth,height=0.08\linewidth]{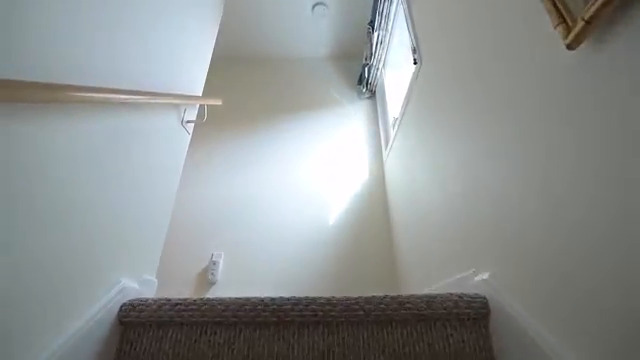}&\includegraphics[width=0.14\linewidth,height=0.08\linewidth]{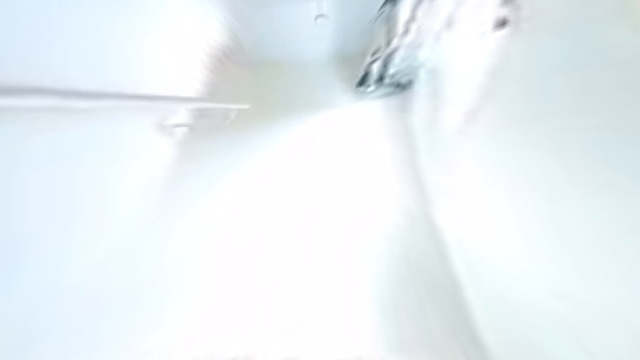}&\includegraphics[width=0.14\linewidth,height=0.08\linewidth]{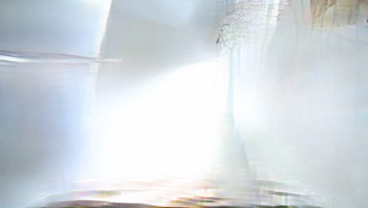}&\includegraphics[width=0.14\linewidth,height=0.08\linewidth]{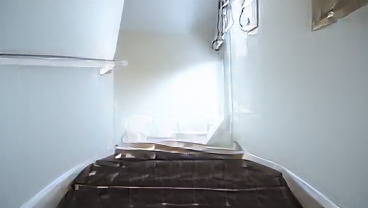}&\includegraphics[width=0.14\linewidth,height=0.08\linewidth]{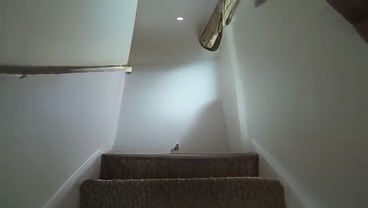}&\includegraphics[width=0.14\linewidth,height=0.08\linewidth]{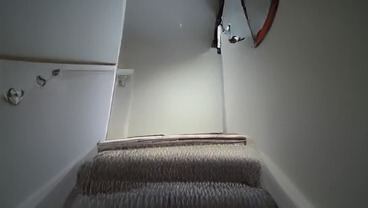}\\%
\includegraphics[width=0.14\linewidth,height=0.08\linewidth]{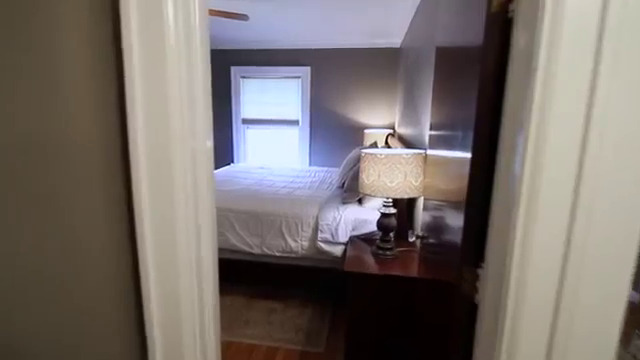}&\includegraphics[width=0.14\linewidth,height=0.08\linewidth]{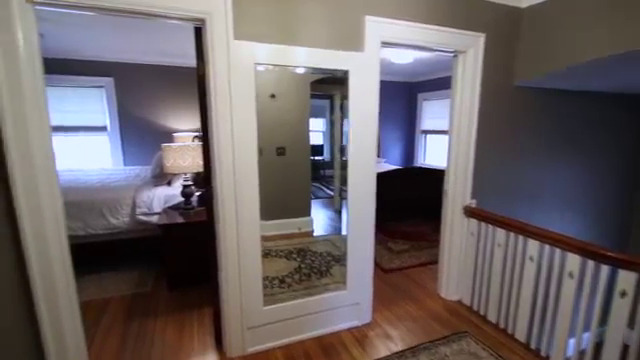}&\includegraphics[width=0.14\linewidth,height=0.08\linewidth]{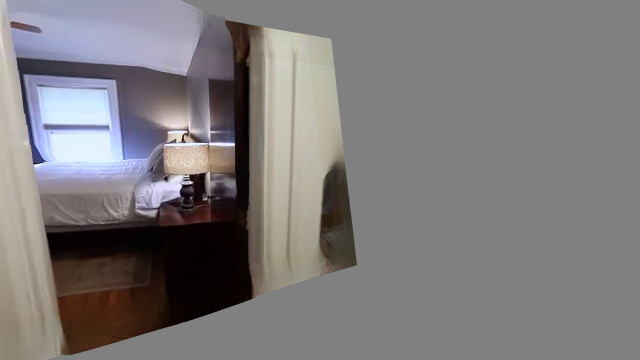}&\includegraphics[width=0.14\linewidth,height=0.08\linewidth]{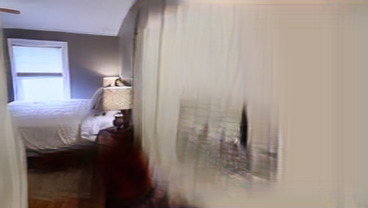}&\includegraphics[width=0.14\linewidth,height=0.08\linewidth]{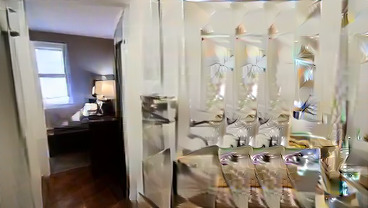}&\includegraphics[width=0.14\linewidth,height=0.08\linewidth]{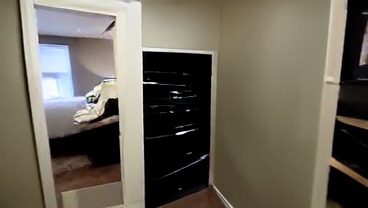}&\includegraphics[width=0.14\linewidth,height=0.08\linewidth]{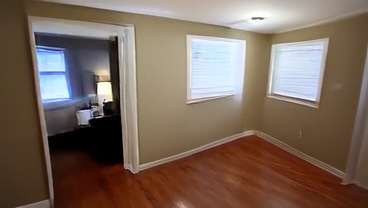}\\%
\includegraphics[width=0.14\linewidth,height=0.08\linewidth]{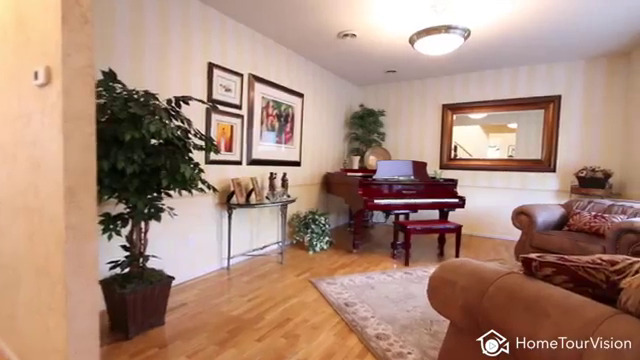}&\includegraphics[width=0.14\linewidth,height=0.08\linewidth]{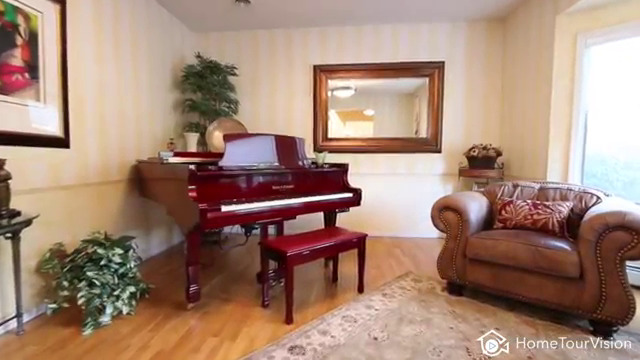}&\includegraphics[width=0.14\linewidth,height=0.08\linewidth]{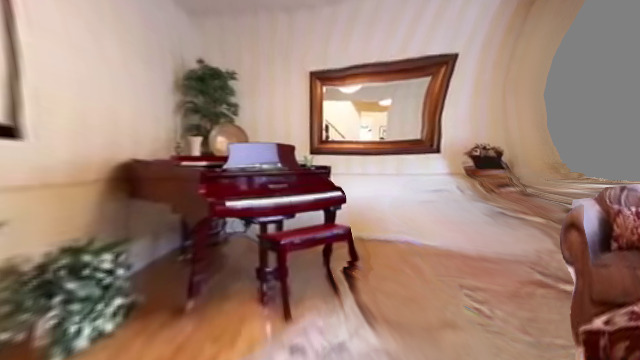}&\includegraphics[width=0.14\linewidth,height=0.08\linewidth]{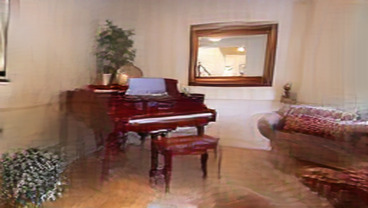}&\includegraphics[width=0.14\linewidth,height=0.08\linewidth]{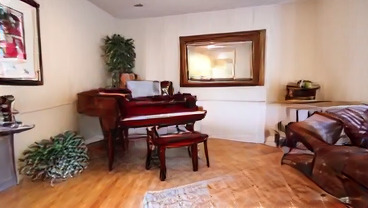}&\includegraphics[width=0.14\linewidth,height=0.08\linewidth]{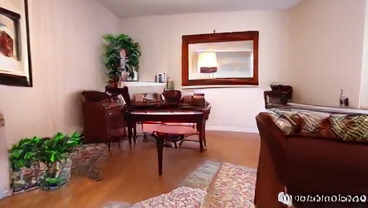}&\includegraphics[width=0.14\linewidth,height=0.08\linewidth]{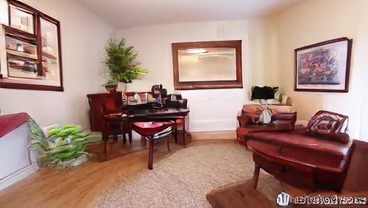}\\%
\includegraphics[width=0.14\linewidth,height=0.08\linewidth]{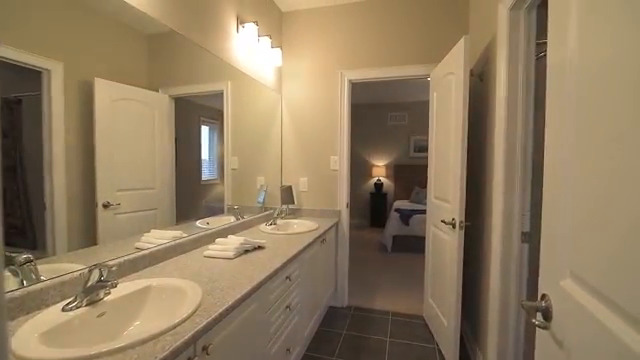}&\includegraphics[width=0.14\linewidth,height=0.08\linewidth]{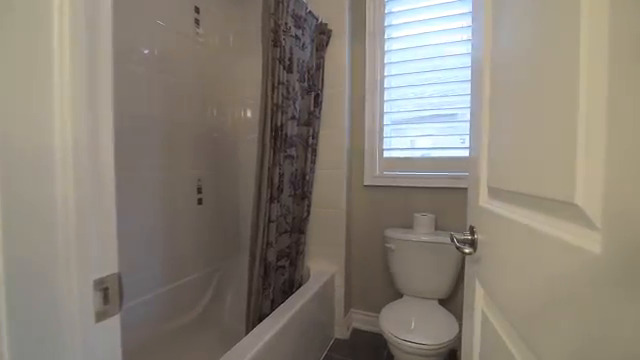}&\includegraphics[width=0.14\linewidth,height=0.08\linewidth]{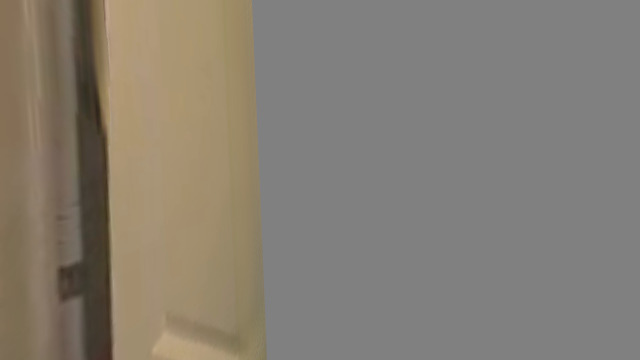}&\includegraphics[width=0.14\linewidth,height=0.08\linewidth]{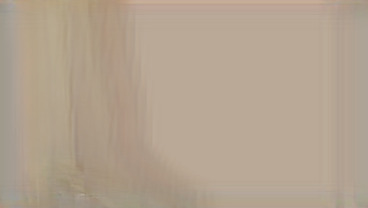}&\includegraphics[width=0.14\linewidth,height=0.08\linewidth]{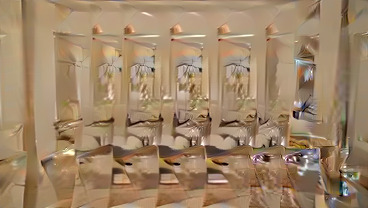}&\includegraphics[width=0.14\linewidth,height=0.08\linewidth]{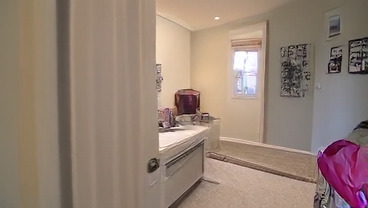}&\includegraphics[width=0.14\linewidth,height=0.08\linewidth]{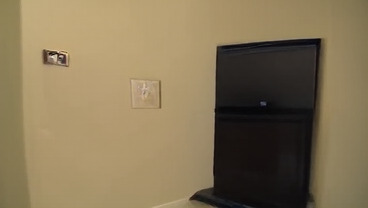}\\%
\includegraphics[width=0.14\linewidth,height=0.08\linewidth]{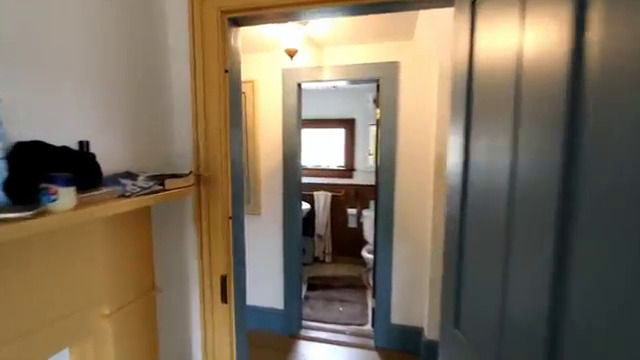}&\includegraphics[width=0.14\linewidth,height=0.08\linewidth]{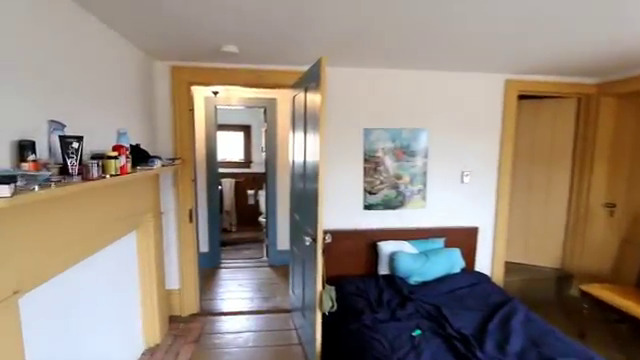}&\includegraphics[width=0.14\linewidth,height=0.08\linewidth]{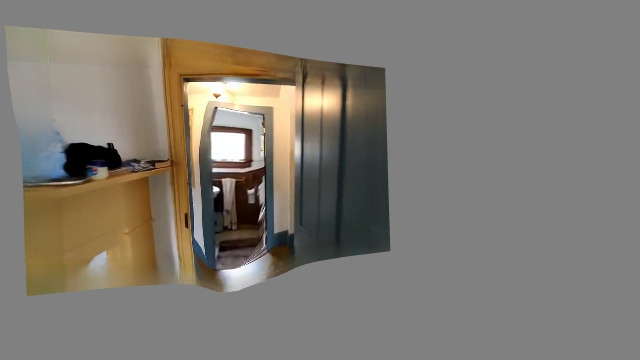}&\includegraphics[width=0.14\linewidth,height=0.08\linewidth]{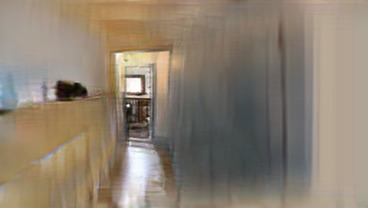}&\includegraphics[width=0.14\linewidth,height=0.08\linewidth]{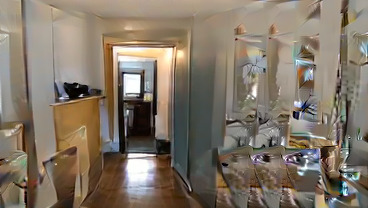}&\includegraphics[width=0.14\linewidth,height=0.08\linewidth]{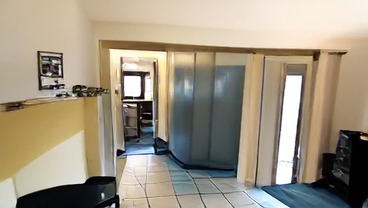}&\includegraphics[width=0.14\linewidth,height=0.08\linewidth]{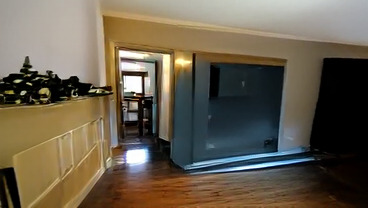}\\%
\includegraphics[width=0.14\linewidth,height=0.08\linewidth]{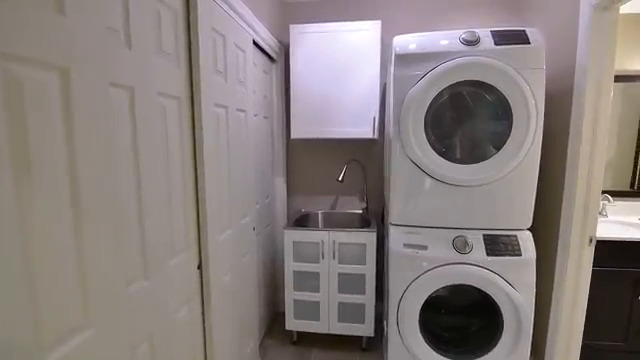}&\includegraphics[width=0.14\linewidth,height=0.08\linewidth]{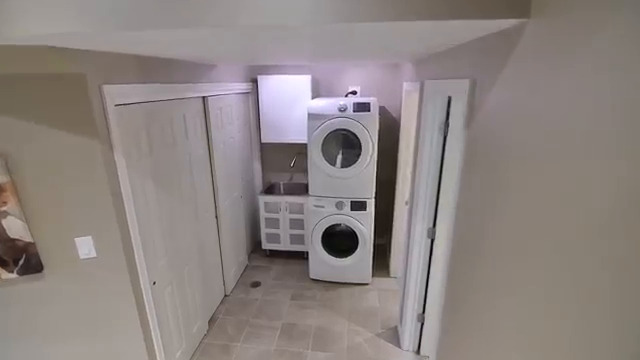}&\includegraphics[width=0.14\linewidth,height=0.08\linewidth]{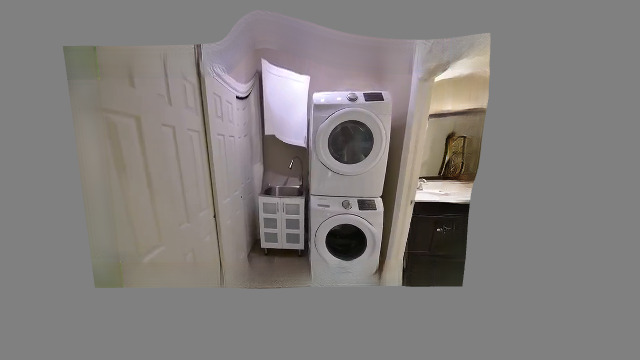}&\includegraphics[width=0.14\linewidth,height=0.08\linewidth]{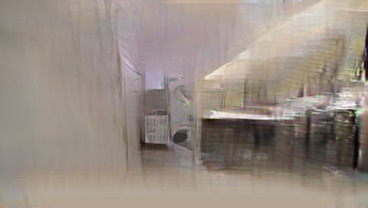}&\includegraphics[width=0.14\linewidth,height=0.08\linewidth]{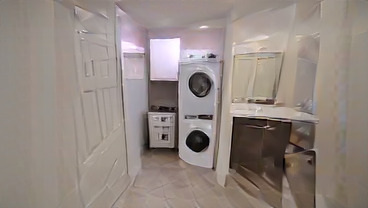}&\includegraphics[width=0.14\linewidth,height=0.08\linewidth]{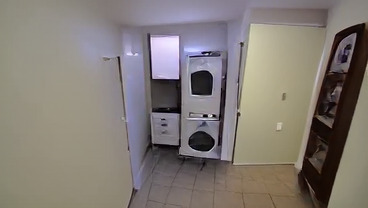}&\includegraphics[width=0.14\linewidth,height=0.08\linewidth]{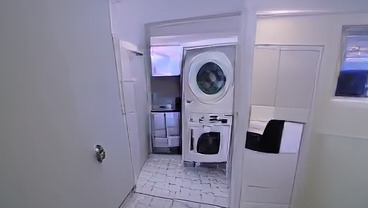}\\%
\includegraphics[width=0.14\linewidth,height=0.08\linewidth]{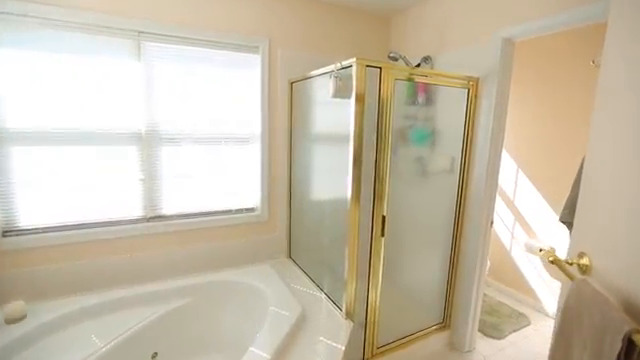}&\includegraphics[width=0.14\linewidth,height=0.08\linewidth]{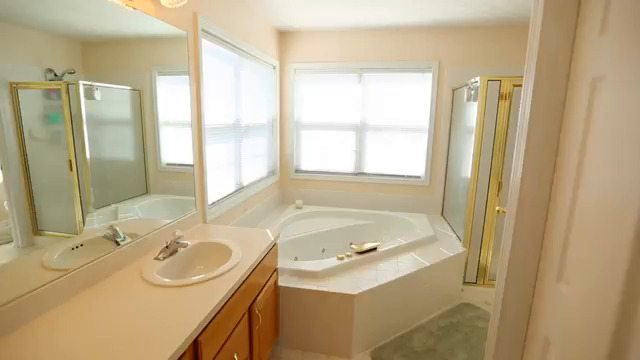}&\includegraphics[width=0.14\linewidth,height=0.08\linewidth]{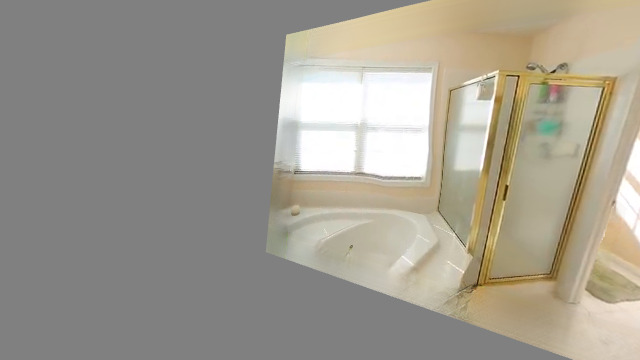}&\includegraphics[width=0.14\linewidth,height=0.08\linewidth]{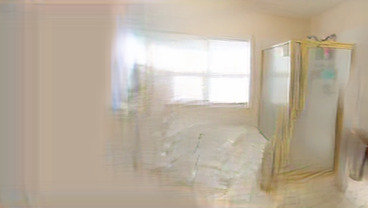}&\includegraphics[width=0.14\linewidth,height=0.08\linewidth]{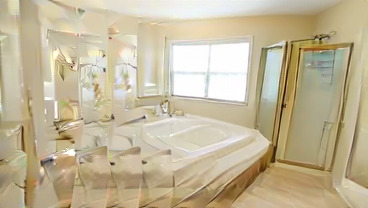}&\includegraphics[width=0.14\linewidth,height=0.08\linewidth]{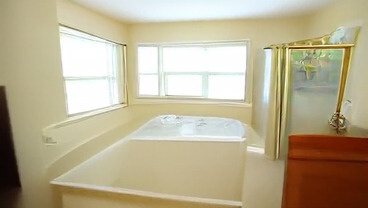}&\includegraphics[width=0.14\linewidth,height=0.08\linewidth]{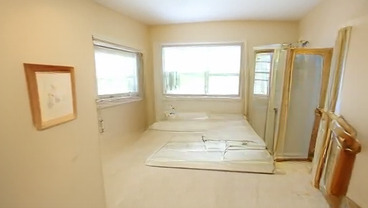}\\%
\includegraphics[width=0.14\linewidth,height=0.08\linewidth]{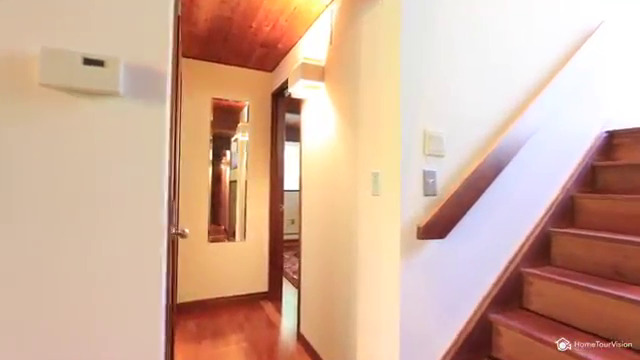}&\includegraphics[width=0.14\linewidth,height=0.08\linewidth]{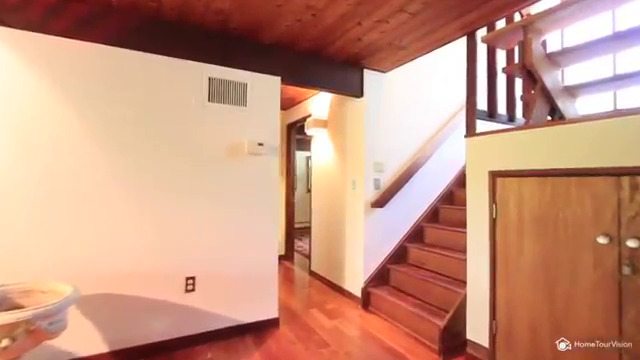}&\includegraphics[width=0.14\linewidth,height=0.08\linewidth]{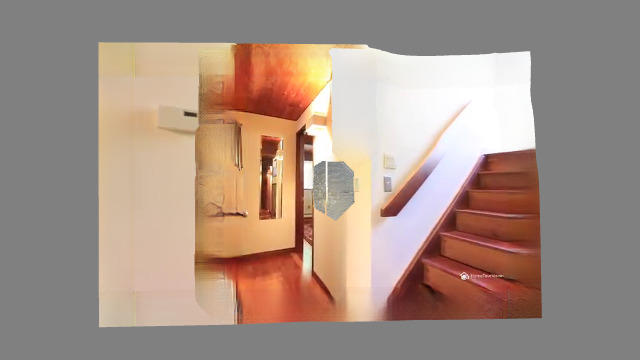}&\includegraphics[width=0.14\linewidth,height=0.08\linewidth]{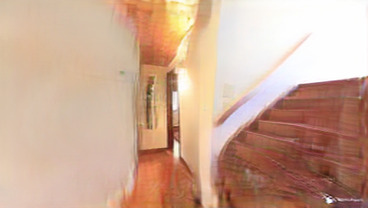}&\includegraphics[width=0.14\linewidth,height=0.08\linewidth]{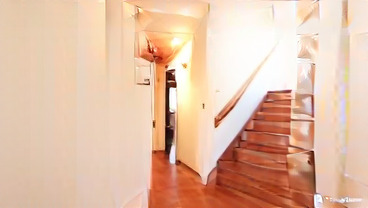}&\includegraphics[width=0.14\linewidth,height=0.08\linewidth]{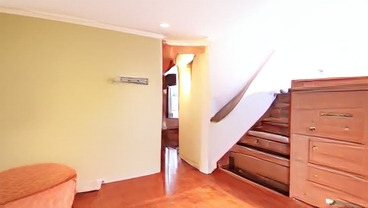}&\includegraphics[width=0.14\linewidth,height=0.08\linewidth]{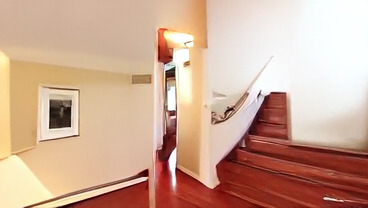}\\%
\includegraphics[width=0.14\linewidth,height=0.08\linewidth]{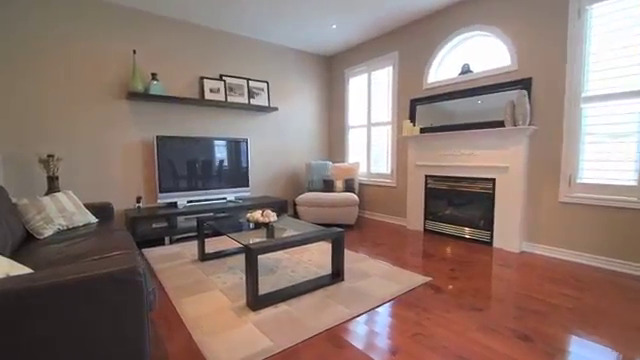}&\includegraphics[width=0.14\linewidth,height=0.08\linewidth]{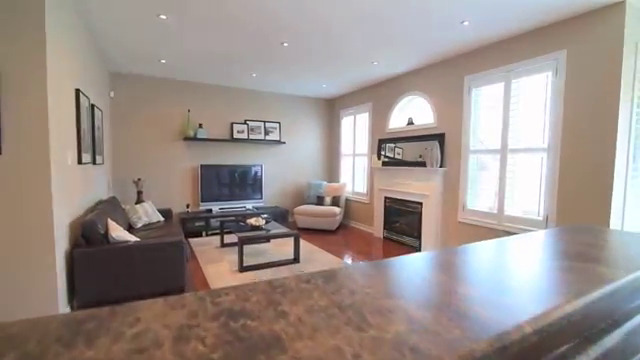}&\includegraphics[width=0.14\linewidth,height=0.08\linewidth]{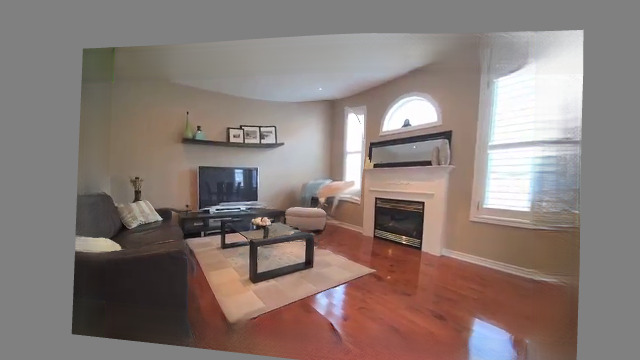}&\includegraphics[width=0.14\linewidth,height=0.08\linewidth]{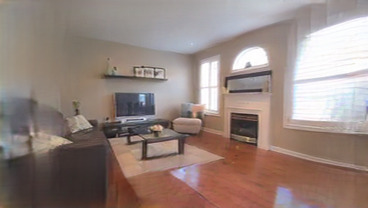}&\includegraphics[width=0.14\linewidth,height=0.08\linewidth]{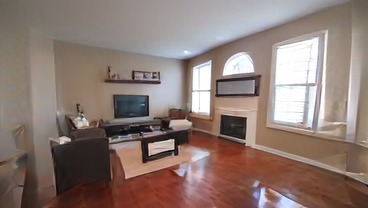}&\includegraphics[width=0.14\linewidth,height=0.08\linewidth]{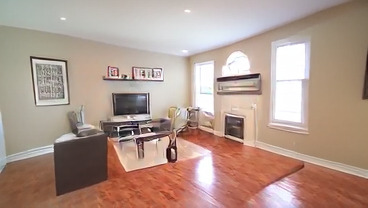}&\includegraphics[width=0.14\linewidth,height=0.08\linewidth]{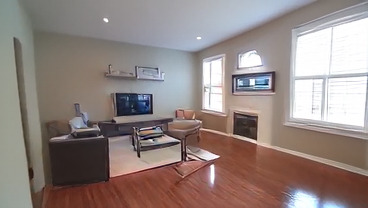}\\%
\includegraphics[width=0.14\linewidth,height=0.08\linewidth]{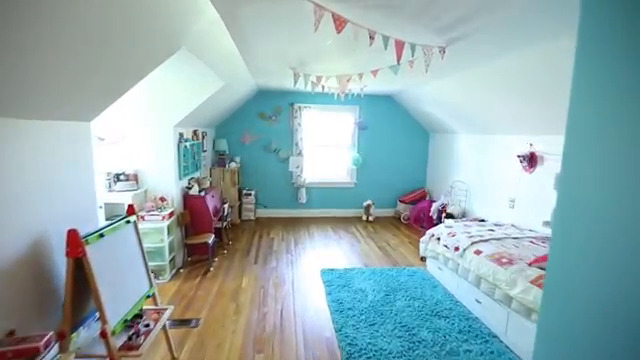}&\includegraphics[width=0.14\linewidth,height=0.08\linewidth]{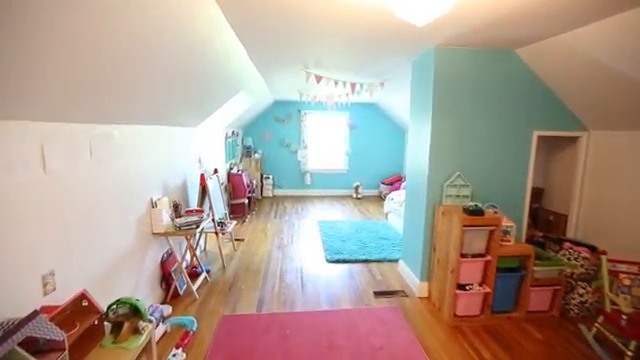}&\includegraphics[width=0.14\linewidth,height=0.08\linewidth]{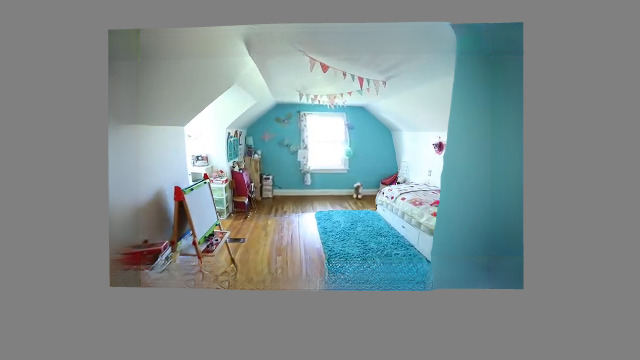}&\includegraphics[width=0.14\linewidth,height=0.08\linewidth]{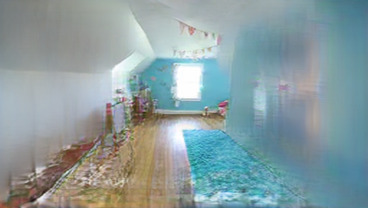}&\includegraphics[width=0.14\linewidth,height=0.08\linewidth]{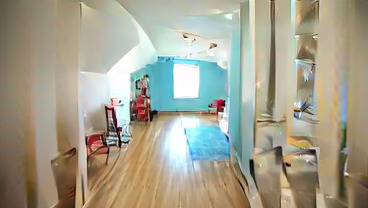}&\includegraphics[width=0.14\linewidth,height=0.08\linewidth]{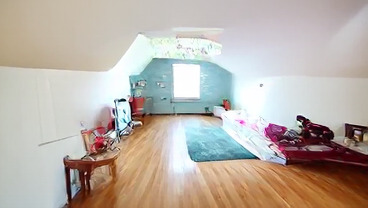}&\includegraphics[width=0.14\linewidth,height=0.08\linewidth]{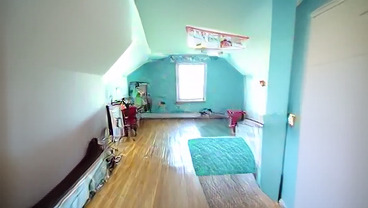}\\%
\includegraphics[width=0.14\linewidth,height=0.08\linewidth]{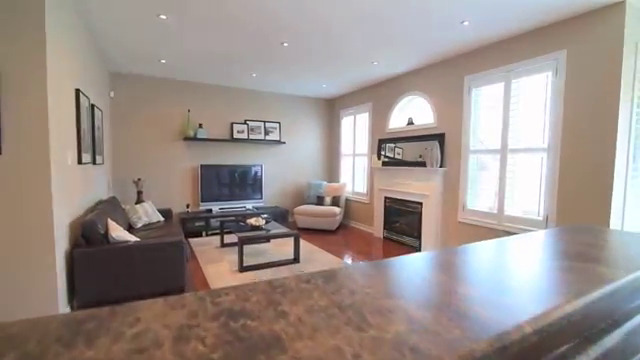}&\includegraphics[width=0.14\linewidth,height=0.08\linewidth]{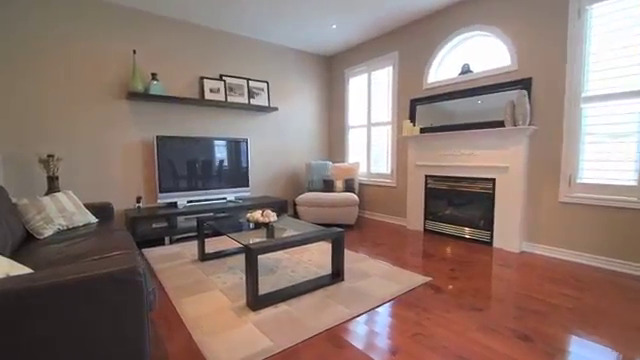}&\includegraphics[width=0.14\linewidth,height=0.08\linewidth]{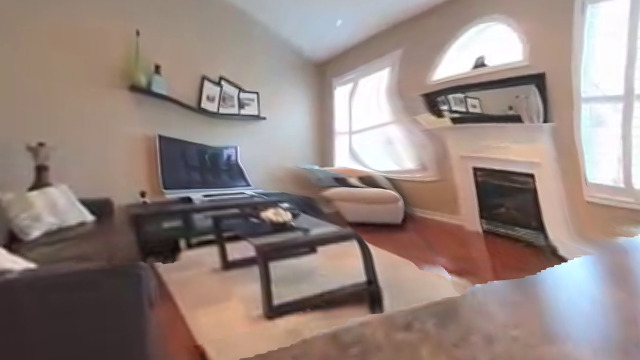}&\includegraphics[width=0.14\linewidth,height=0.08\linewidth]{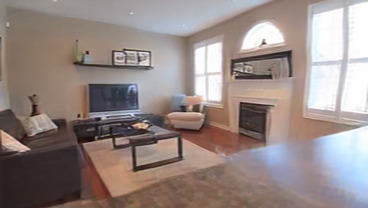}&\includegraphics[width=0.14\linewidth,height=0.08\linewidth]{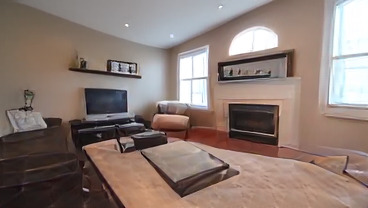}&\includegraphics[width=0.14\linewidth,height=0.08\linewidth]{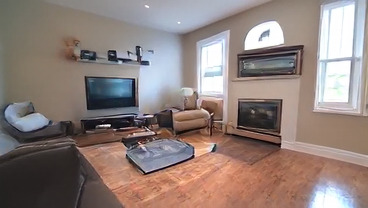}&\includegraphics[width=0.14\linewidth,height=0.08\linewidth]{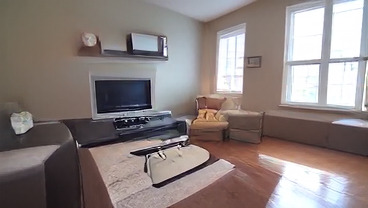}\\\bottomrule%
\end{tabular}
}
   \end{center}
     \vspace{-0.75em}
      \caption{Additional qualitative comparisons on RealEstate10K.
      }
   \label{fig:supprealestate}
   \end{figure*}
}

\newcommand{\figsuppacid}        {
   \begin{figure*}[bthp]
   \begin{center}
     \resizebox{\linewidth}{!}{%
       \setlength{\tabcolsep}{0.1em}%
\begin{tabular}{@{}cc@{\hskip 0.75em}ccc@{\hskip 0.75em}cc@{}}%
\toprule%
Source&Target&\threedp&\infnat{}&\vqwarper{}&\varv{}&\varvi{}\\%
\midrule%
\includegraphics[width=0.14\linewidth,height=0.08\linewidth]{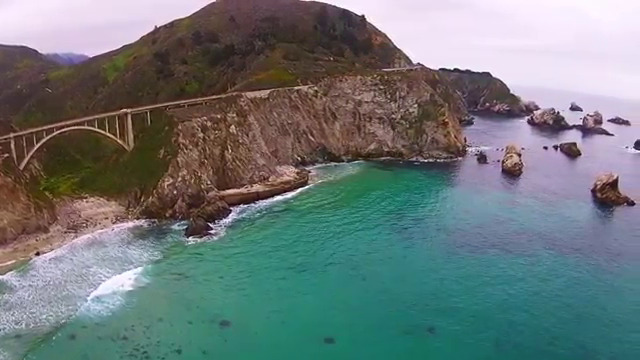}&\includegraphics[width=0.14\linewidth,height=0.08\linewidth]{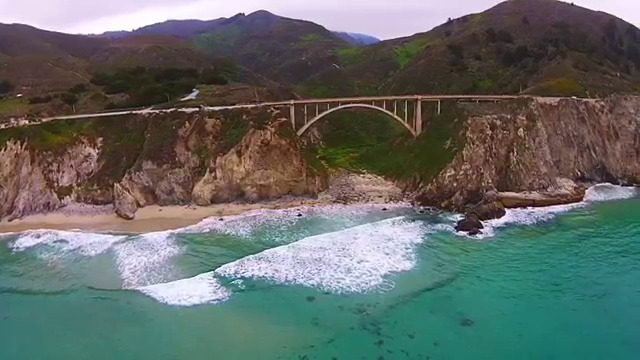}&\includegraphics[width=0.14\linewidth,height=0.08\linewidth]{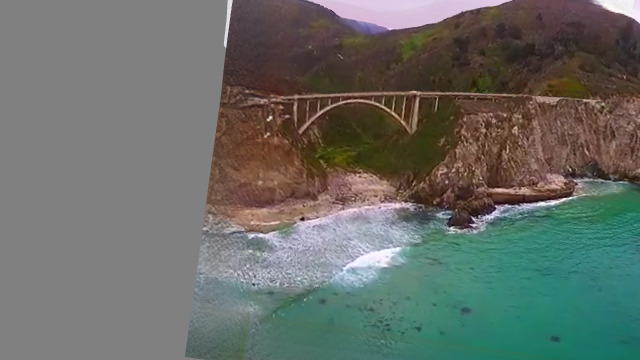}&\includegraphics[width=0.14\linewidth,height=0.08\linewidth]{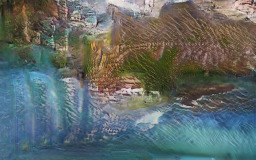}&\includegraphics[width=0.14\linewidth,height=0.08\linewidth]{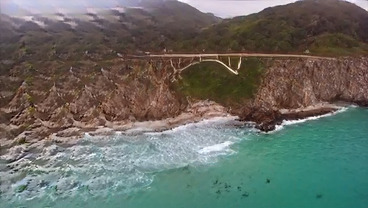}&\includegraphics[width=0.14\linewidth,height=0.08\linewidth]{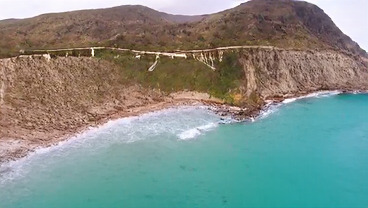}&\includegraphics[width=0.14\linewidth,height=0.08\linewidth]{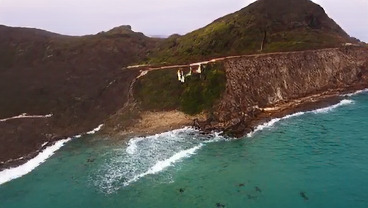}\\%
\includegraphics[width=0.14\linewidth,height=0.08\linewidth]{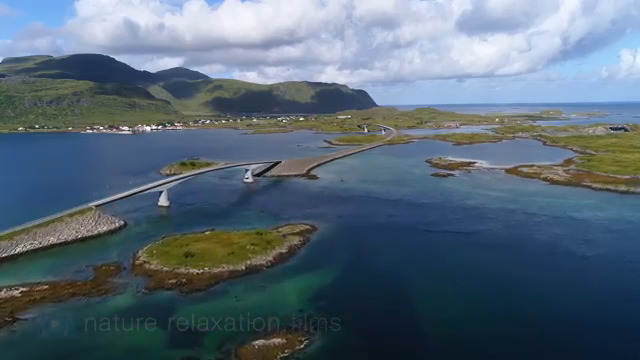}&\includegraphics[width=0.14\linewidth,height=0.08\linewidth]{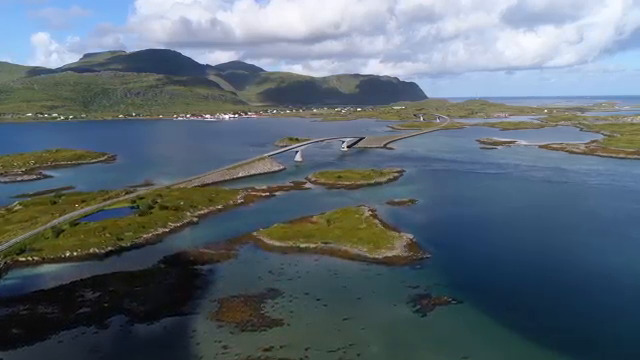}&\includegraphics[width=0.14\linewidth,height=0.08\linewidth]{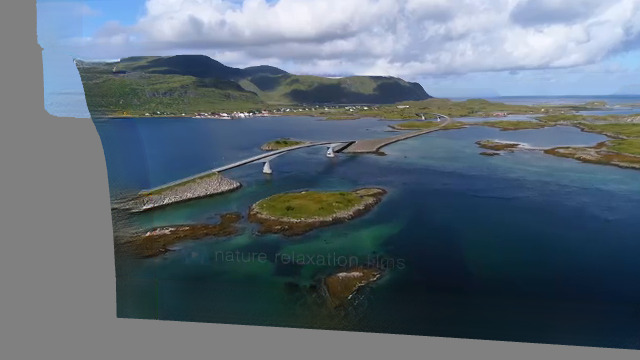}&\includegraphics[width=0.14\linewidth,height=0.08\linewidth]{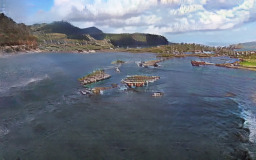}&\includegraphics[width=0.14\linewidth,height=0.08\linewidth]{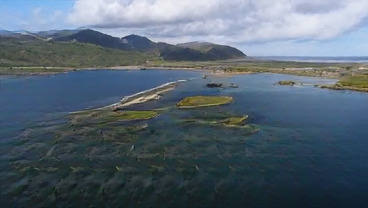}&\includegraphics[width=0.14\linewidth,height=0.08\linewidth]{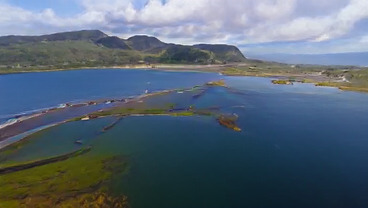}&\includegraphics[width=0.14\linewidth,height=0.08\linewidth]{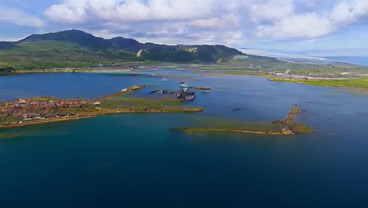}\\%
\includegraphics[width=0.14\linewidth,height=0.08\linewidth]{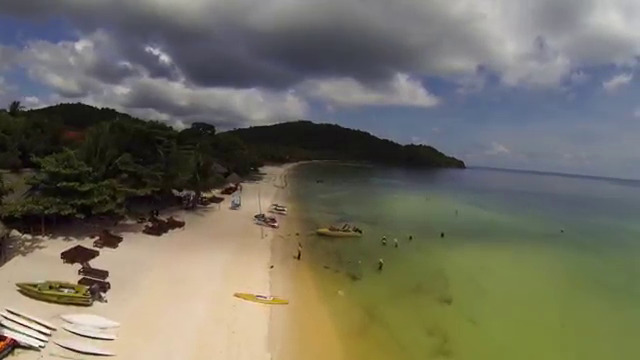}&\includegraphics[width=0.14\linewidth,height=0.08\linewidth]{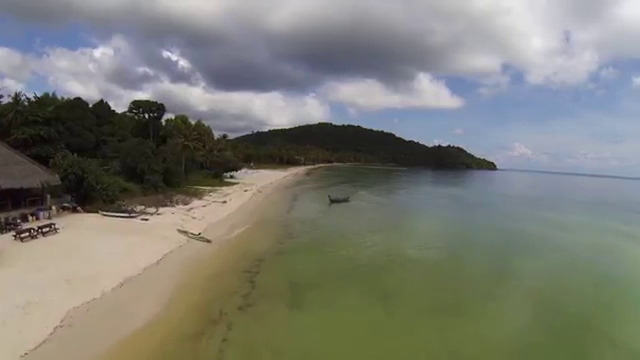}&\includegraphics[width=0.14\linewidth,height=0.08\linewidth]{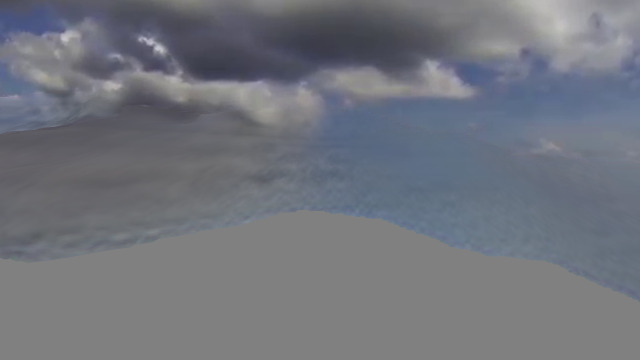}&\includegraphics[width=0.14\linewidth,height=0.08\linewidth]{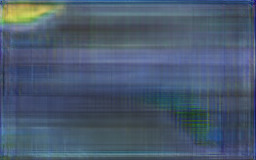}&\includegraphics[width=0.14\linewidth,height=0.08\linewidth]{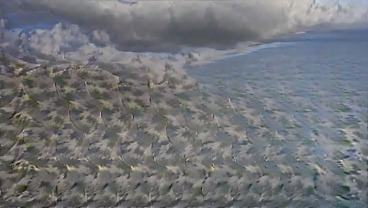}&\includegraphics[width=0.14\linewidth,height=0.08\linewidth]{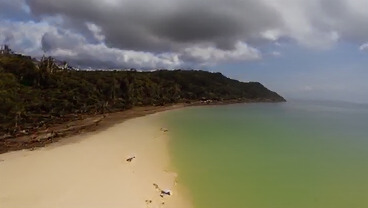}&\includegraphics[width=0.14\linewidth,height=0.08\linewidth]{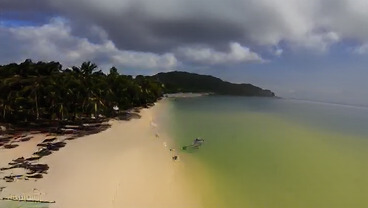}\\%
\includegraphics[width=0.14\linewidth,height=0.08\linewidth]{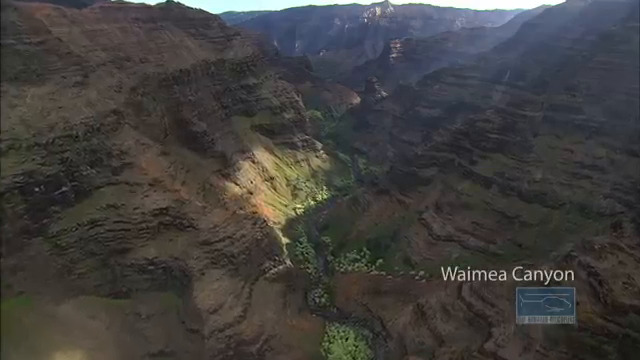}&\includegraphics[width=0.14\linewidth,height=0.08\linewidth]{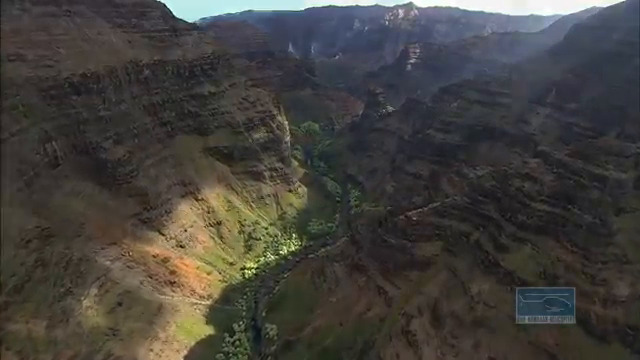}&\includegraphics[width=0.14\linewidth,height=0.08\linewidth]{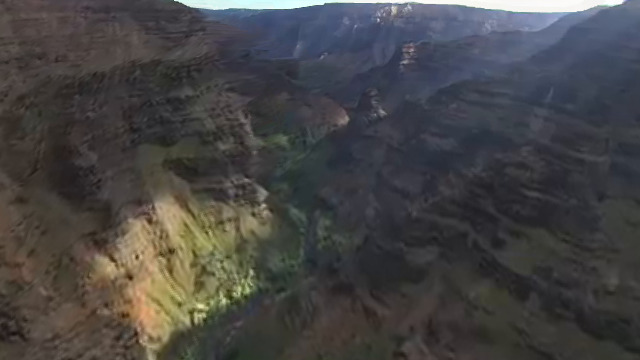}&\includegraphics[width=0.14\linewidth,height=0.08\linewidth]{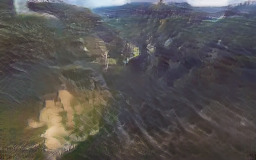}&\includegraphics[width=0.14\linewidth,height=0.08\linewidth]{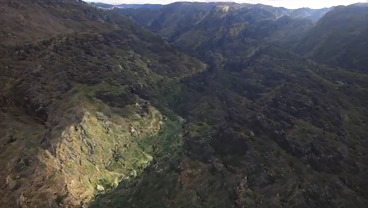}&\includegraphics[width=0.14\linewidth,height=0.08\linewidth]{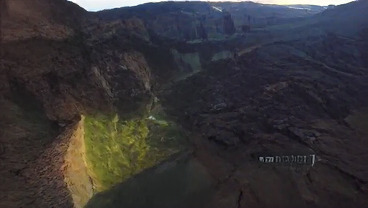}&\includegraphics[width=0.14\linewidth,height=0.08\linewidth]{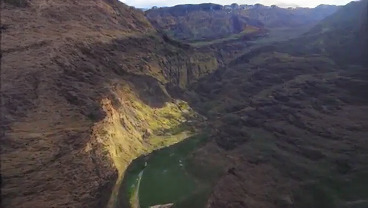}\\%
\includegraphics[width=0.14\linewidth,height=0.08\linewidth]{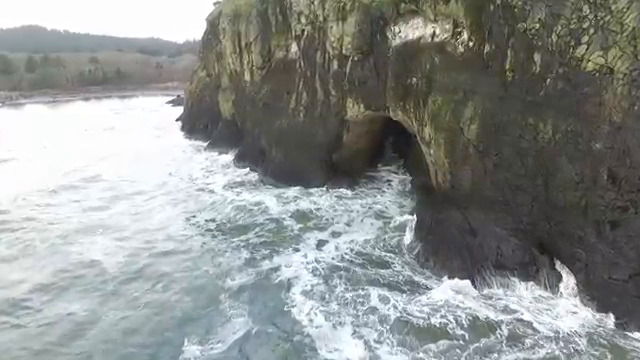}&\includegraphics[width=0.14\linewidth,height=0.08\linewidth]{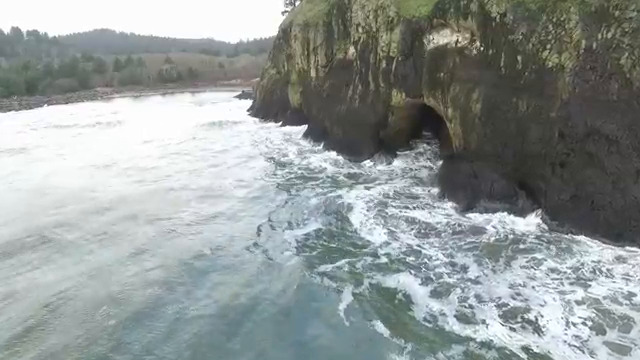}&\includegraphics[width=0.14\linewidth,height=0.08\linewidth]{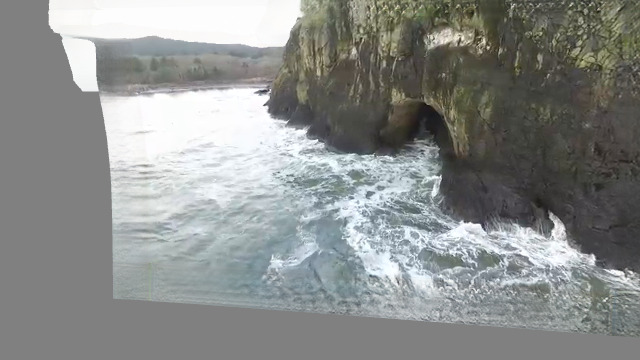}&\includegraphics[width=0.14\linewidth,height=0.08\linewidth]{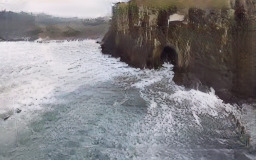}&\includegraphics[width=0.14\linewidth,height=0.08\linewidth]{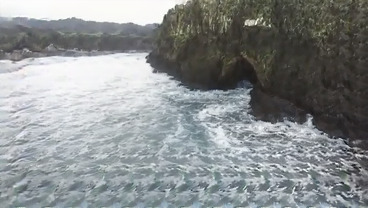}&\includegraphics[width=0.14\linewidth,height=0.08\linewidth]{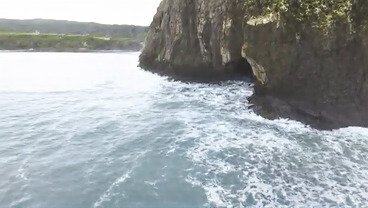}&\includegraphics[width=0.14\linewidth,height=0.08\linewidth]{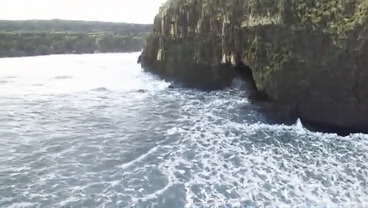}\\%
\includegraphics[width=0.14\linewidth,height=0.08\linewidth]{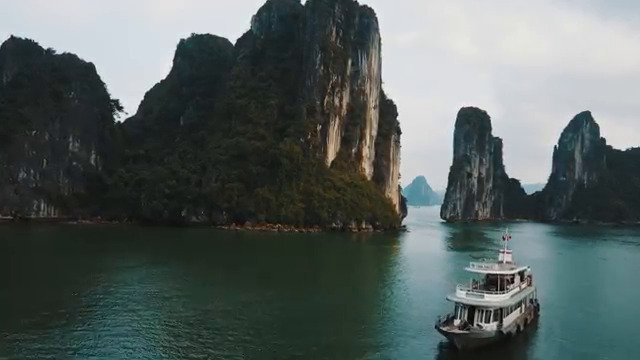}&\includegraphics[width=0.14\linewidth,height=0.08\linewidth]{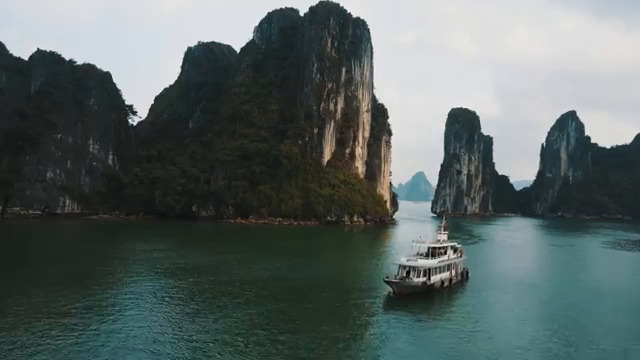}&\includegraphics[width=0.14\linewidth,height=0.08\linewidth]{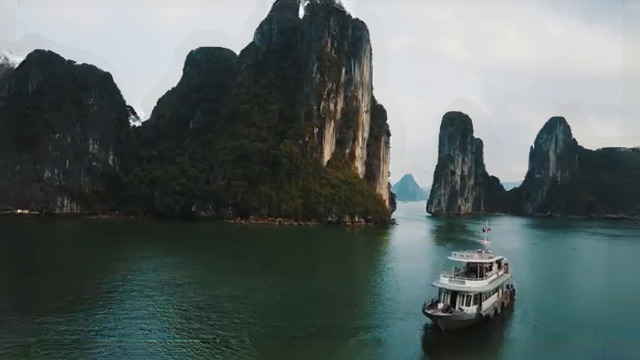}&\includegraphics[width=0.14\linewidth,height=0.08\linewidth]{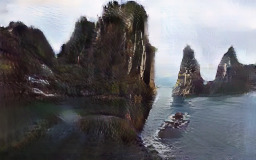}&\includegraphics[width=0.14\linewidth,height=0.08\linewidth]{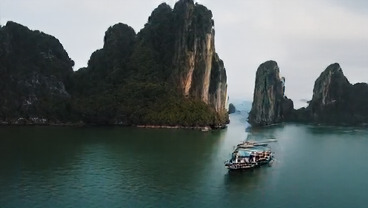}&\includegraphics[width=0.14\linewidth,height=0.08\linewidth]{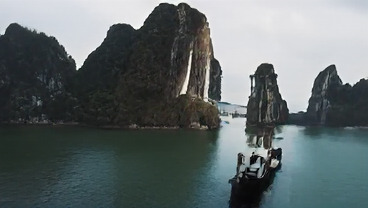}&\includegraphics[width=0.14\linewidth,height=0.08\linewidth]{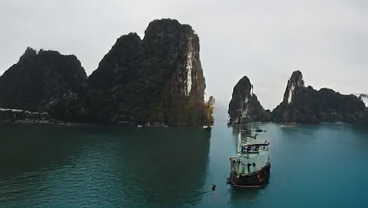}\\%
\includegraphics[width=0.14\linewidth,height=0.08\linewidth]{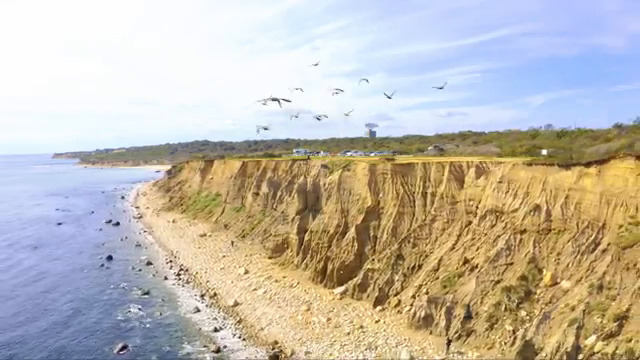}&\includegraphics[width=0.14\linewidth,height=0.08\linewidth]{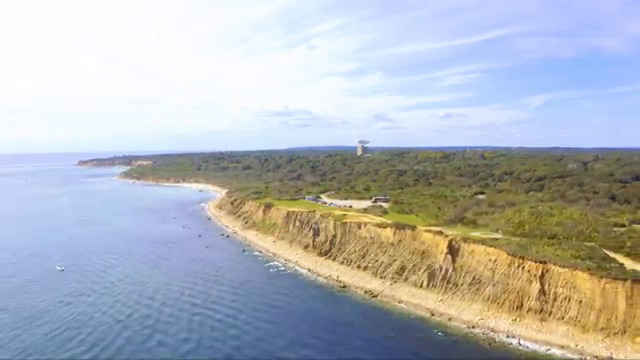}&\includegraphics[width=0.14\linewidth,height=0.08\linewidth]{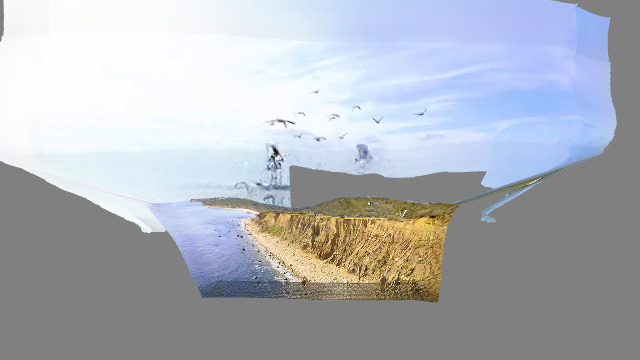}&\includegraphics[width=0.14\linewidth,height=0.08\linewidth]{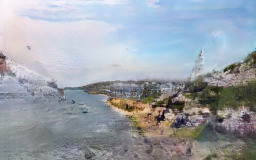}&\includegraphics[width=0.14\linewidth,height=0.08\linewidth]{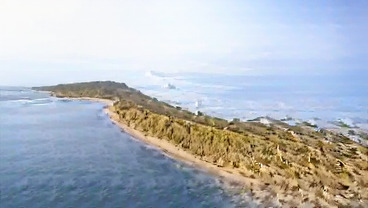}&\includegraphics[width=0.14\linewidth,height=0.08\linewidth]{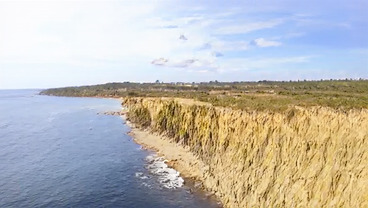}&\includegraphics[width=0.14\linewidth,height=0.08\linewidth]{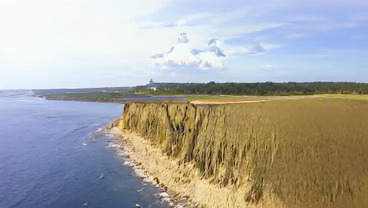}\\%
\includegraphics[width=0.14\linewidth,height=0.08\linewidth]{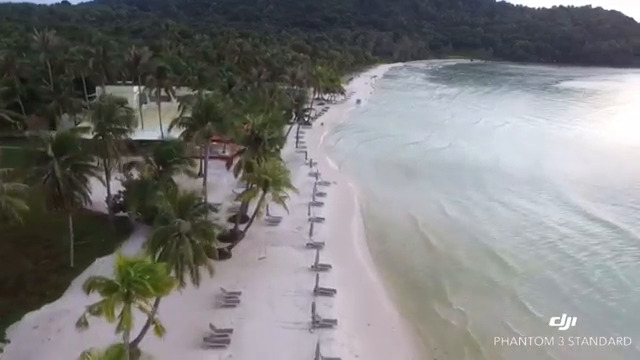}&\includegraphics[width=0.14\linewidth,height=0.08\linewidth]{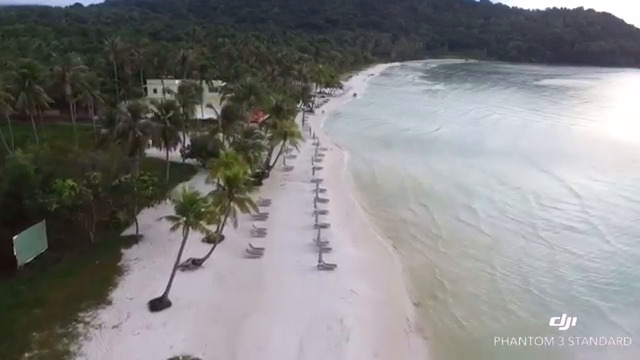}&\includegraphics[width=0.14\linewidth,height=0.08\linewidth]{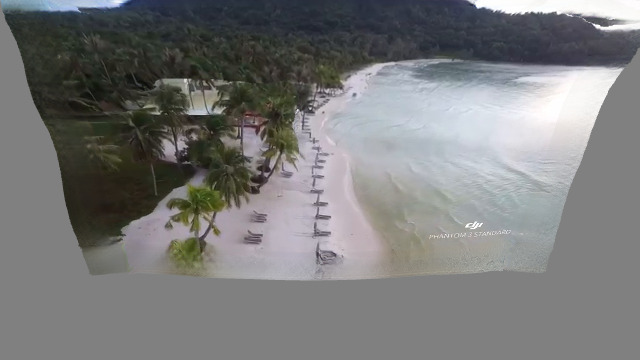}&\includegraphics[width=0.14\linewidth,height=0.08\linewidth]{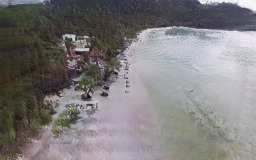}&\includegraphics[width=0.14\linewidth,height=0.08\linewidth]{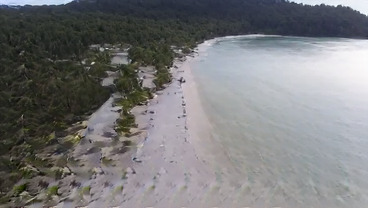}&\includegraphics[width=0.14\linewidth,height=0.08\linewidth]{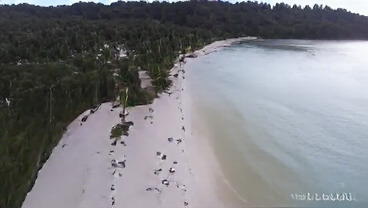}&\includegraphics[width=0.14\linewidth,height=0.08\linewidth]{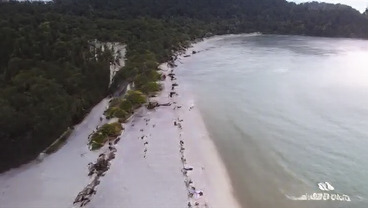}\\%
\includegraphics[width=0.14\linewidth,height=0.08\linewidth]{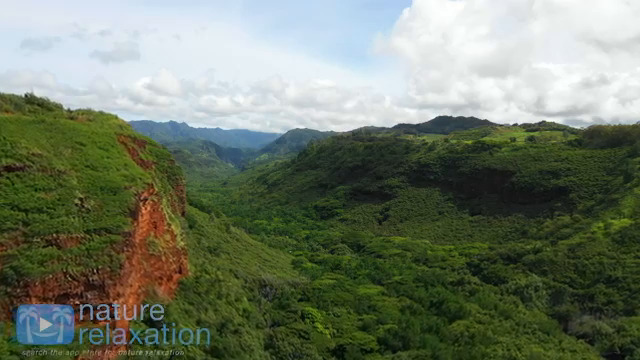}&\includegraphics[width=0.14\linewidth,height=0.08\linewidth]{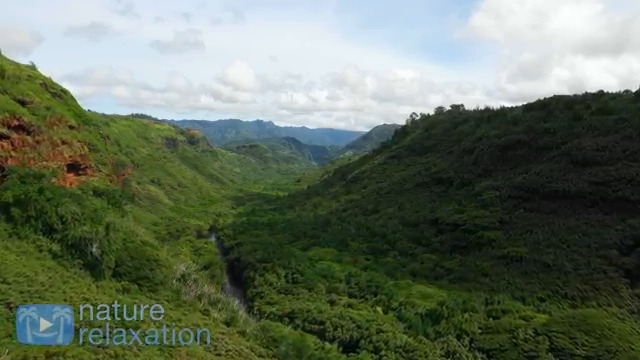}&\includegraphics[width=0.14\linewidth,height=0.08\linewidth]{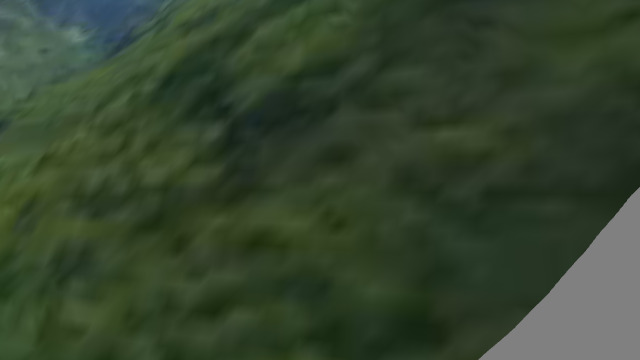}&\includegraphics[width=0.14\linewidth,height=0.08\linewidth]{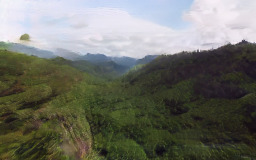}&\includegraphics[width=0.14\linewidth,height=0.08\linewidth]{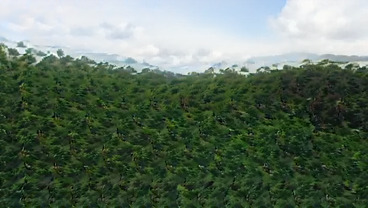}&\includegraphics[width=0.14\linewidth,height=0.08\linewidth]{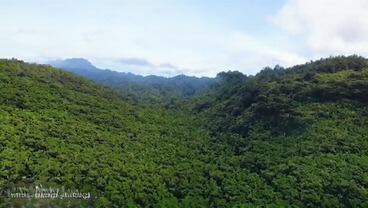}&\includegraphics[width=0.14\linewidth,height=0.08\linewidth]{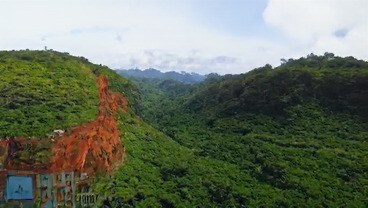}\\%
\includegraphics[width=0.14\linewidth,height=0.08\linewidth]{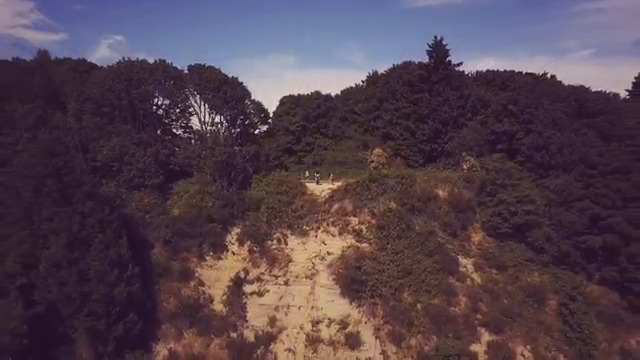}&\includegraphics[width=0.14\linewidth,height=0.08\linewidth]{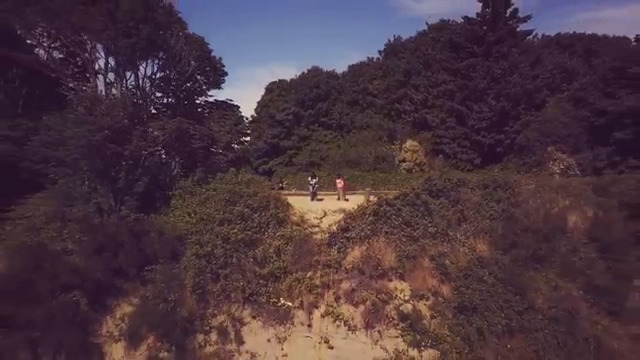}&\includegraphics[width=0.14\linewidth,height=0.08\linewidth]{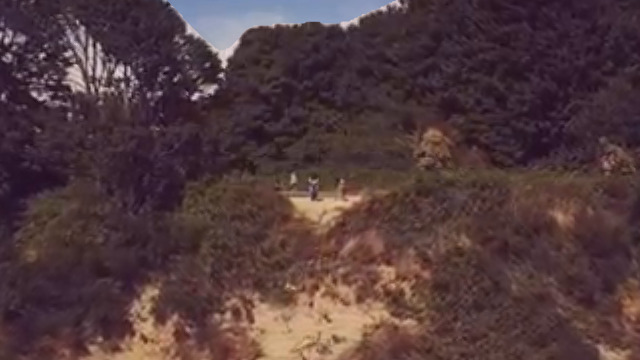}&\includegraphics[width=0.14\linewidth,height=0.08\linewidth]{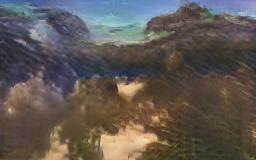}&\includegraphics[width=0.14\linewidth,height=0.08\linewidth]{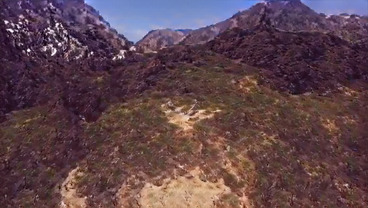}&\includegraphics[width=0.14\linewidth,height=0.08\linewidth]{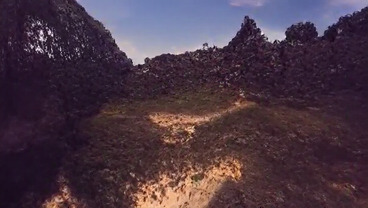}&\includegraphics[width=0.14\linewidth,height=0.08\linewidth]{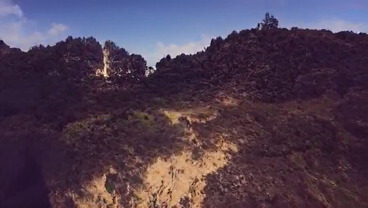}\\%
\includegraphics[width=0.14\linewidth,height=0.08\linewidth]{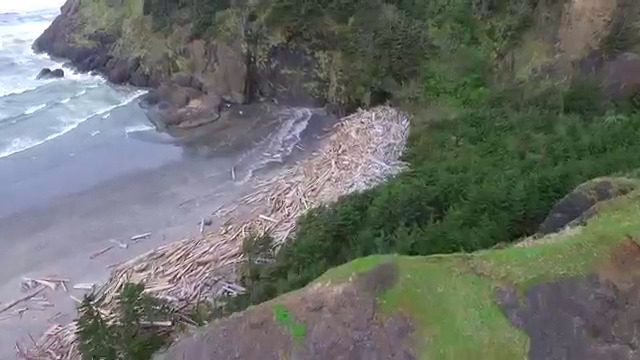}&\includegraphics[width=0.14\linewidth,height=0.08\linewidth]{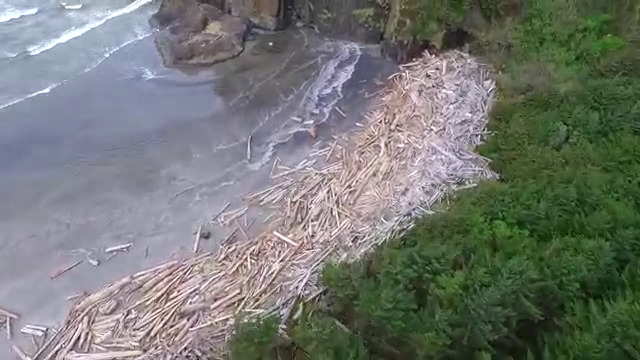}&\includegraphics[width=0.14\linewidth,height=0.08\linewidth]{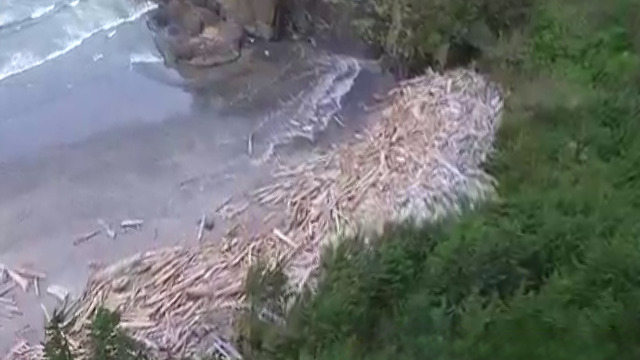}&\includegraphics[width=0.14\linewidth,height=0.08\linewidth]{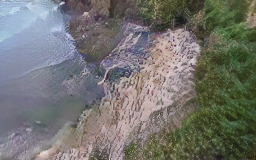}&\includegraphics[width=0.14\linewidth,height=0.08\linewidth]{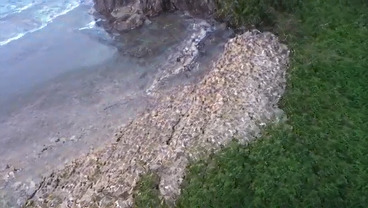}&\includegraphics[width=0.14\linewidth,height=0.08\linewidth]{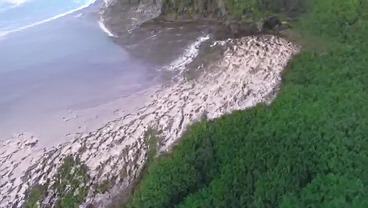}&\includegraphics[width=0.14\linewidth,height=0.08\linewidth]{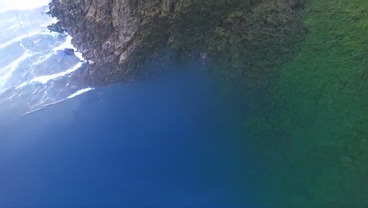}\\%
\includegraphics[width=0.14\linewidth,height=0.08\linewidth]{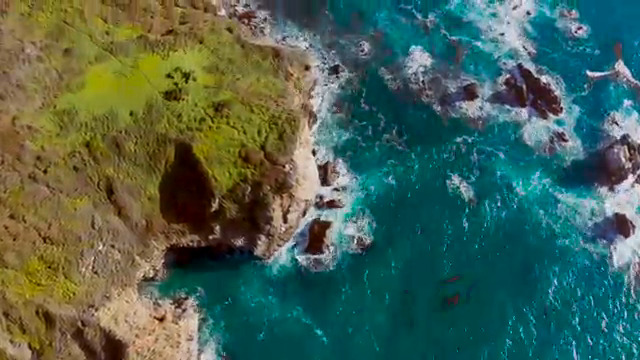}&\includegraphics[width=0.14\linewidth,height=0.08\linewidth]{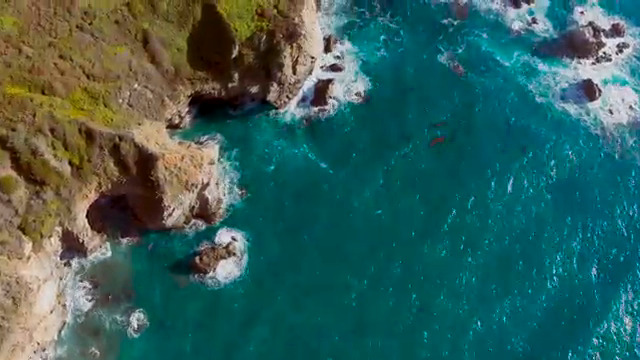}&\includegraphics[width=0.14\linewidth,height=0.08\linewidth]{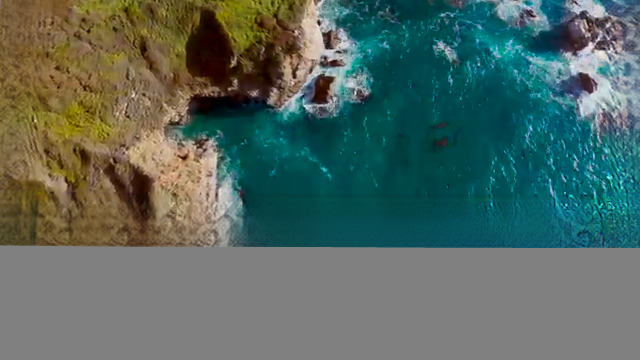}&\includegraphics[width=0.14\linewidth,height=0.08\linewidth]{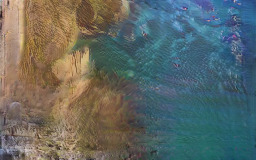}&\includegraphics[width=0.14\linewidth,height=0.08\linewidth]{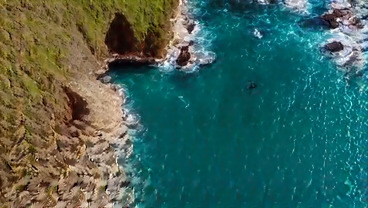}&\includegraphics[width=0.14\linewidth,height=0.08\linewidth]{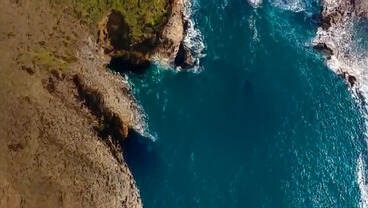}&\includegraphics[width=0.14\linewidth,height=0.08\linewidth]{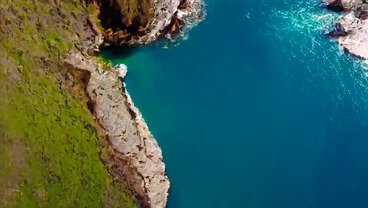}\\%
\includegraphics[width=0.14\linewidth,height=0.08\linewidth]{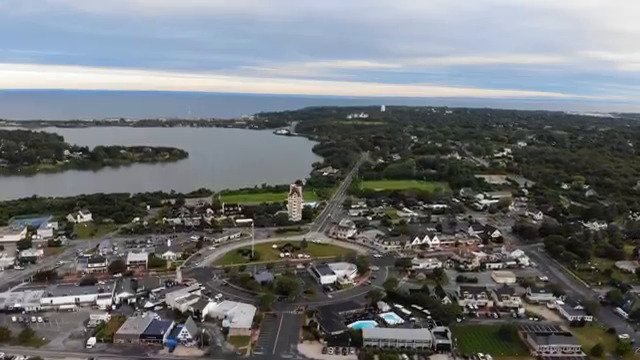}&\includegraphics[width=0.14\linewidth,height=0.08\linewidth]{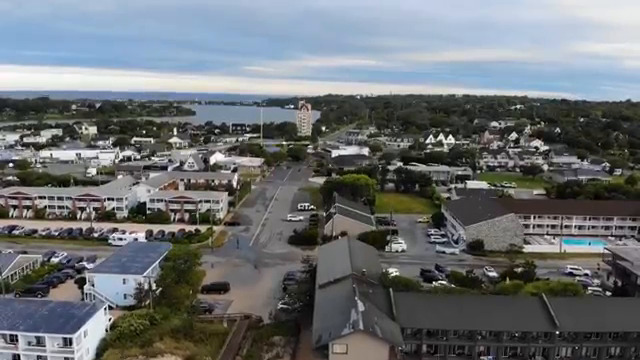}&\includegraphics[width=0.14\linewidth,height=0.08\linewidth]{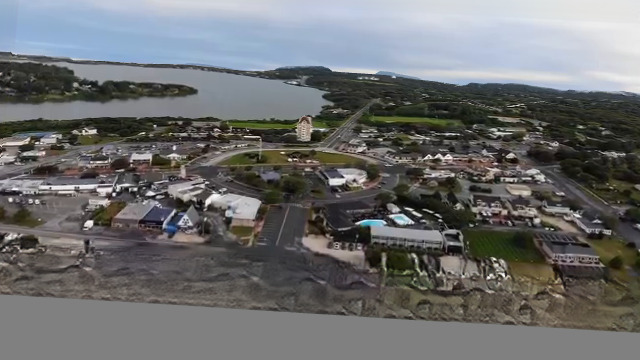}&\includegraphics[width=0.14\linewidth,height=0.08\linewidth]{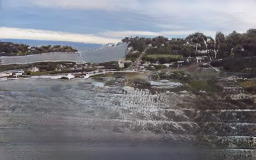}&\includegraphics[width=0.14\linewidth,height=0.08\linewidth]{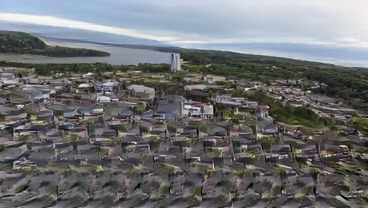}&\includegraphics[width=0.14\linewidth,height=0.08\linewidth]{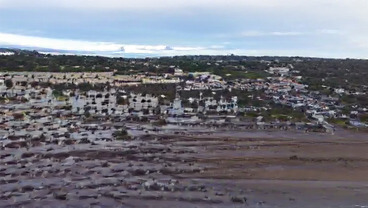}&\includegraphics[width=0.14\linewidth,height=0.08\linewidth]{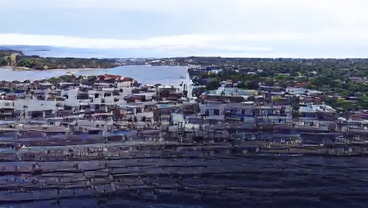}\\%
\includegraphics[width=0.14\linewidth,height=0.08\linewidth]{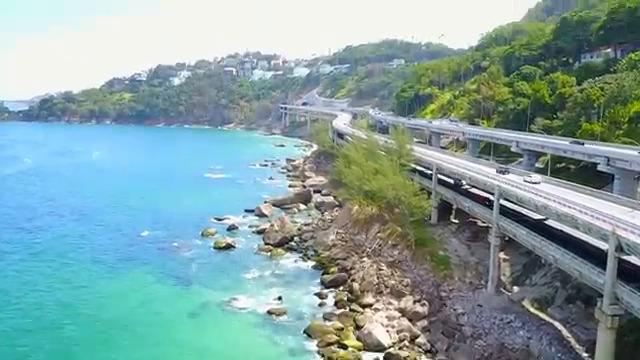}&\includegraphics[width=0.14\linewidth,height=0.08\linewidth]{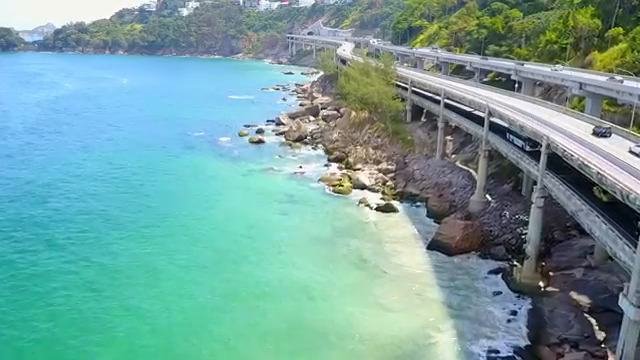}&\includegraphics[width=0.14\linewidth,height=0.08\linewidth]{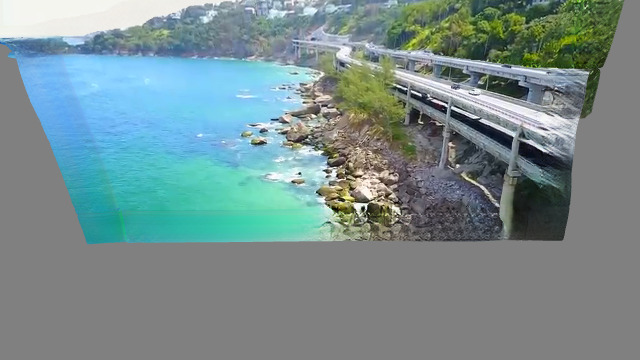}&\includegraphics[width=0.14\linewidth,height=0.08\linewidth]{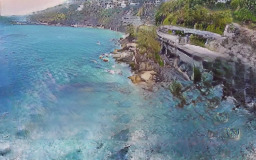}&\includegraphics[width=0.14\linewidth,height=0.08\linewidth]{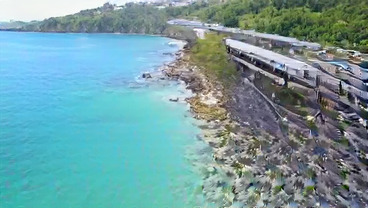}&\includegraphics[width=0.14\linewidth,height=0.08\linewidth]{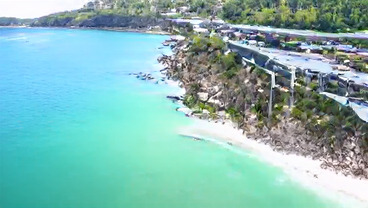}&\includegraphics[width=0.14\linewidth,height=0.08\linewidth]{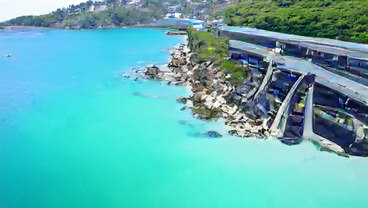}\\\bottomrule%
\end{tabular}
}
   \end{center}
     \vspace{-0.75em}
      \caption{Additional qualitative comparisons on ACID.
      }
   \label{fig:suppacid}
   \end{figure*}
}

\newcommand{\figsupprealestatesamples}        {
   \begin{figure*}[bthp]
   \begin{center}
     \resizebox{\linewidth}{!}{%
       \setlength{\tabcolsep}{0.1em}%
\begin{tabular}{@{}cc@{\hskip 0.75em}ccccc@{}}%
\toprule%
Source&$\sigma$&\multicolumn{5}{c}{samples of variant \varv{}}\\%
\midrule%
\includegraphics[width=0.14\linewidth,height=0.08\linewidth]{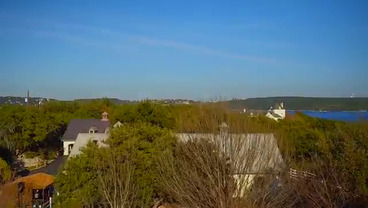}&\includegraphics[width=0.14\linewidth,height=0.08\linewidth]{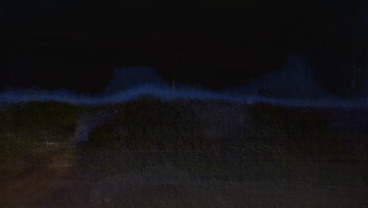}&\includegraphics[width=0.14\linewidth,height=0.08\linewidth]{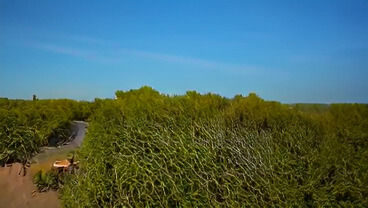}&\includegraphics[width=0.14\linewidth,height=0.08\linewidth]{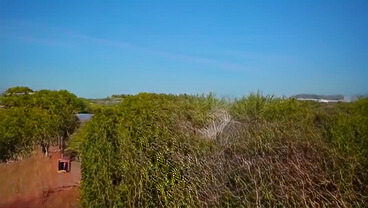}&\includegraphics[width=0.14\linewidth,height=0.08\linewidth]{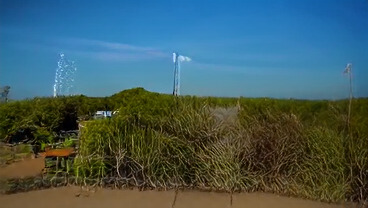}&\includegraphics[width=0.14\linewidth,height=0.08\linewidth]{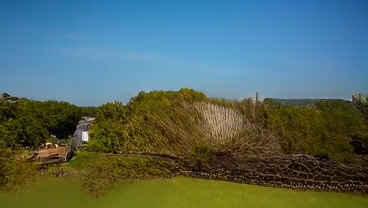}&\includegraphics[width=0.14\linewidth,height=0.08\linewidth]{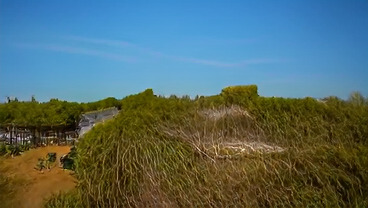}\\%
\includegraphics[width=0.14\linewidth,height=0.08\linewidth]{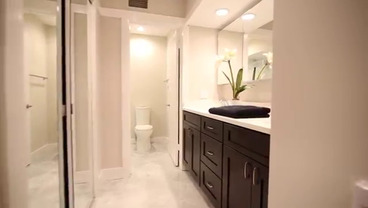}&\includegraphics[width=0.14\linewidth,height=0.08\linewidth]{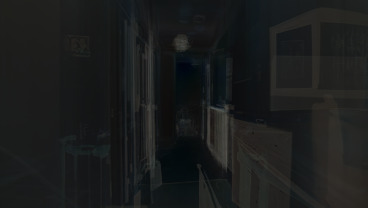}&\includegraphics[width=0.14\linewidth,height=0.08\linewidth]{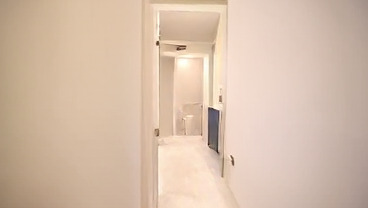}&\includegraphics[width=0.14\linewidth,height=0.08\linewidth]{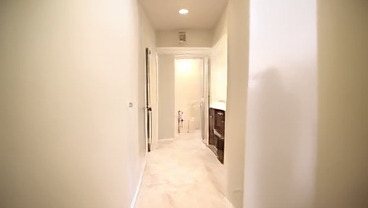}&\includegraphics[width=0.14\linewidth,height=0.08\linewidth]{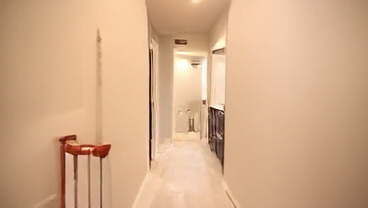}&\includegraphics[width=0.14\linewidth,height=0.08\linewidth]{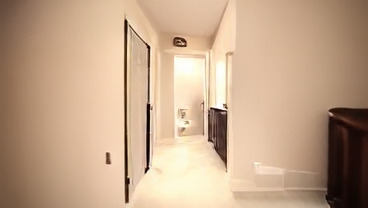}&\includegraphics[width=0.14\linewidth,height=0.08\linewidth]{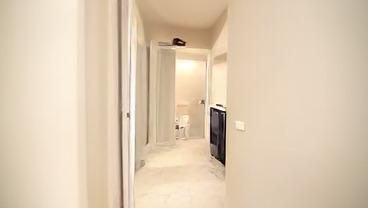}\\%
\includegraphics[width=0.14\linewidth,height=0.08\linewidth]{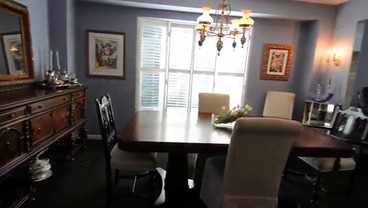}&\includegraphics[width=0.14\linewidth,height=0.08\linewidth]{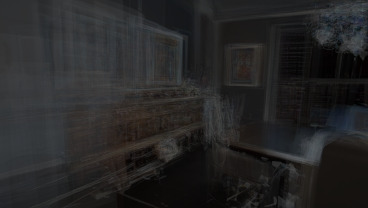}&\includegraphics[width=0.14\linewidth,height=0.08\linewidth]{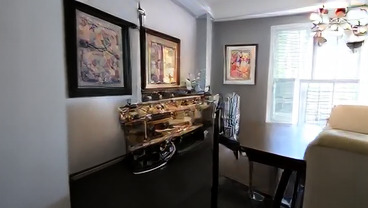}&\includegraphics[width=0.14\linewidth,height=0.08\linewidth]{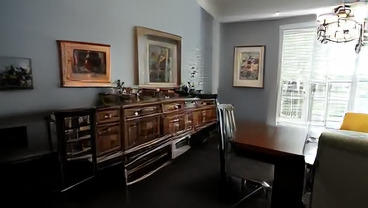}&\includegraphics[width=0.14\linewidth,height=0.08\linewidth]{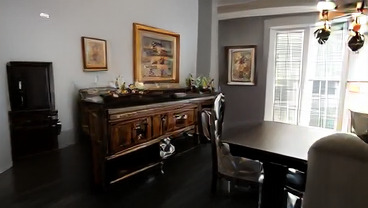}&\includegraphics[width=0.14\linewidth,height=0.08\linewidth]{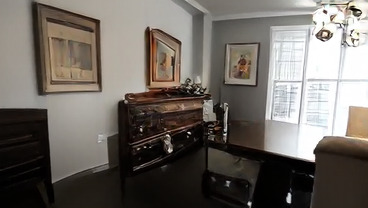}&\includegraphics[width=0.14\linewidth,height=0.08\linewidth]{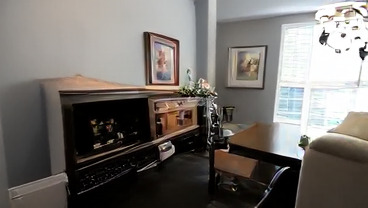}\\%
\includegraphics[width=0.14\linewidth,height=0.08\linewidth]{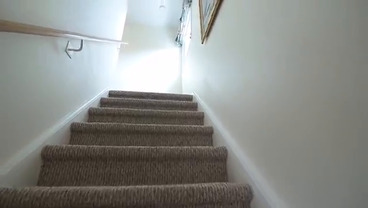}&\includegraphics[width=0.14\linewidth,height=0.08\linewidth]{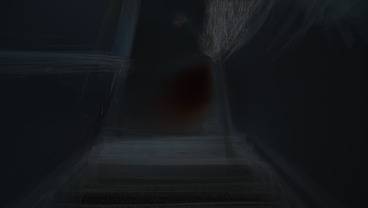}&\includegraphics[width=0.14\linewidth,height=0.08\linewidth]{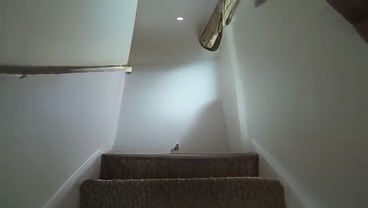}&\includegraphics[width=0.14\linewidth,height=0.08\linewidth]{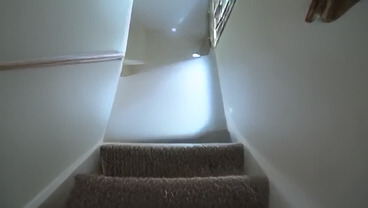}&\includegraphics[width=0.14\linewidth,height=0.08\linewidth]{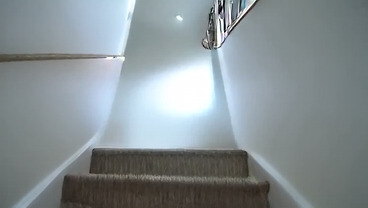}&\includegraphics[width=0.14\linewidth,height=0.08\linewidth]{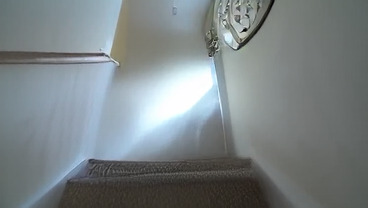}&\includegraphics[width=0.14\linewidth,height=0.08\linewidth]{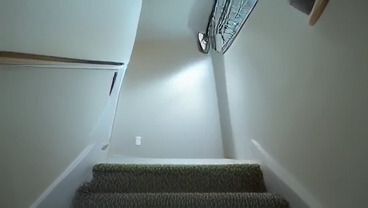}\\%
\includegraphics[width=0.14\linewidth,height=0.08\linewidth]{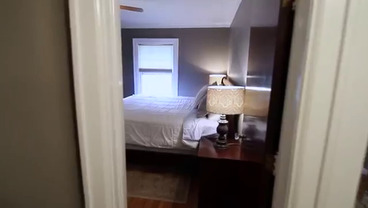}&\includegraphics[width=0.14\linewidth,height=0.08\linewidth]{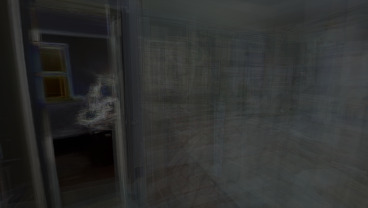}&\includegraphics[width=0.14\linewidth,height=0.08\linewidth]{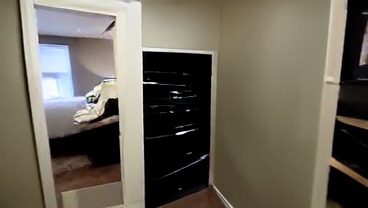}&\includegraphics[width=0.14\linewidth,height=0.08\linewidth]{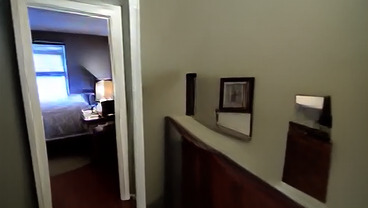}&\includegraphics[width=0.14\linewidth,height=0.08\linewidth]{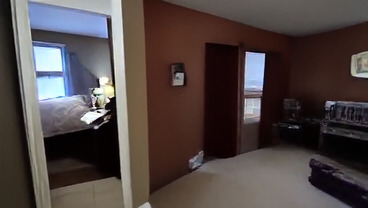}&\includegraphics[width=0.14\linewidth,height=0.08\linewidth]{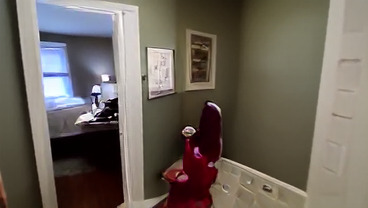}&\includegraphics[width=0.14\linewidth,height=0.08\linewidth]{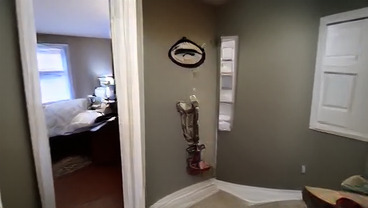}\\%
\includegraphics[width=0.14\linewidth,height=0.08\linewidth]{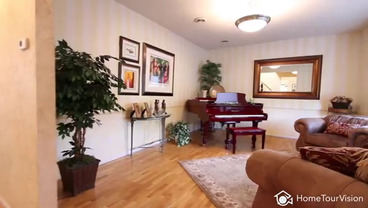}&\includegraphics[width=0.14\linewidth,height=0.08\linewidth]{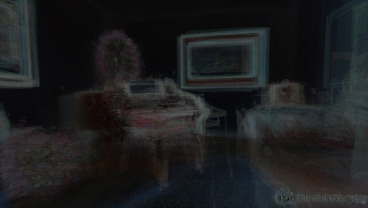}&\includegraphics[width=0.14\linewidth,height=0.08\linewidth]{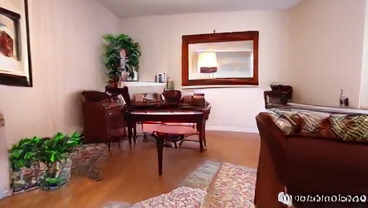}&\includegraphics[width=0.14\linewidth,height=0.08\linewidth]{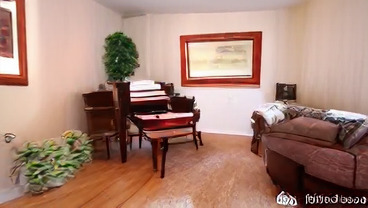}&\includegraphics[width=0.14\linewidth,height=0.08\linewidth]{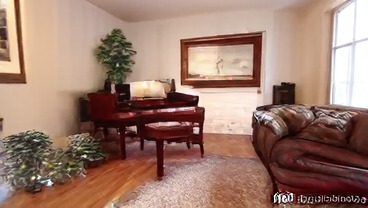}&\includegraphics[width=0.14\linewidth,height=0.08\linewidth]{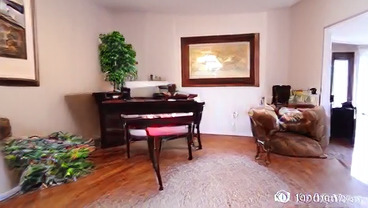}&\includegraphics[width=0.14\linewidth,height=0.08\linewidth]{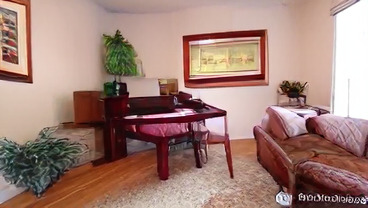}\\%
\includegraphics[width=0.14\linewidth,height=0.08\linewidth]{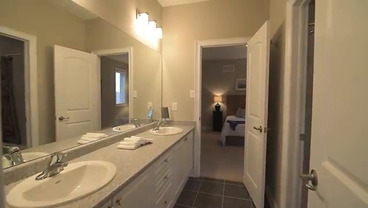}&\includegraphics[width=0.14\linewidth,height=0.08\linewidth]{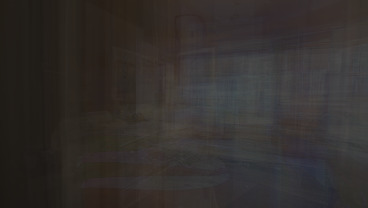}&\includegraphics[width=0.14\linewidth,height=0.08\linewidth]{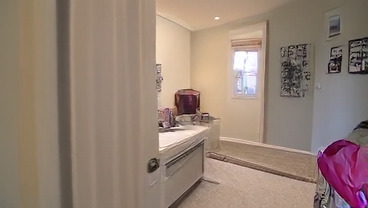}&\includegraphics[width=0.14\linewidth,height=0.08\linewidth]{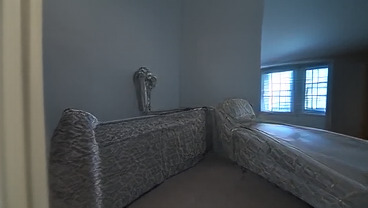}&\includegraphics[width=0.14\linewidth,height=0.08\linewidth]{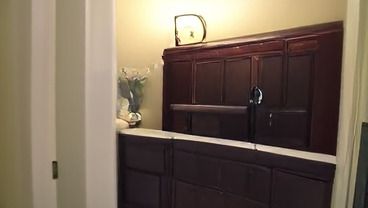}&\includegraphics[width=0.14\linewidth,height=0.08\linewidth]{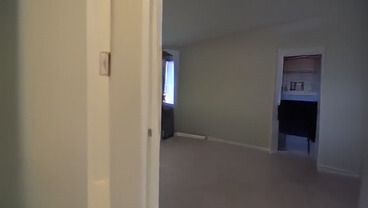}&\includegraphics[width=0.14\linewidth,height=0.08\linewidth]{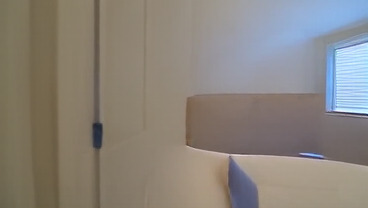}\\%
\includegraphics[width=0.14\linewidth,height=0.08\linewidth]{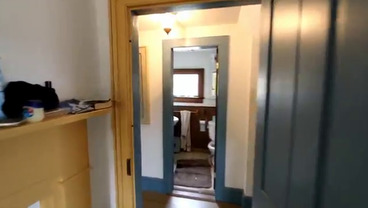}&\includegraphics[width=0.14\linewidth,height=0.08\linewidth]{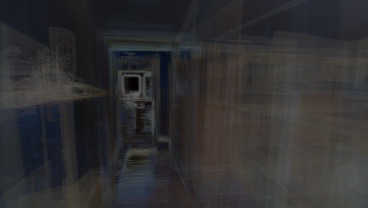}&\includegraphics[width=0.14\linewidth,height=0.08\linewidth]{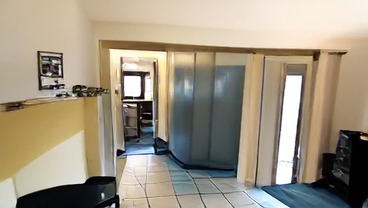}&\includegraphics[width=0.14\linewidth,height=0.08\linewidth]{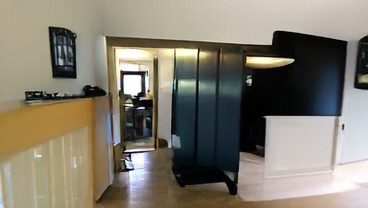}&\includegraphics[width=0.14\linewidth,height=0.08\linewidth]{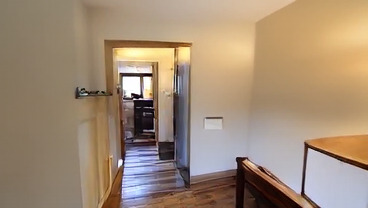}&\includegraphics[width=0.14\linewidth,height=0.08\linewidth]{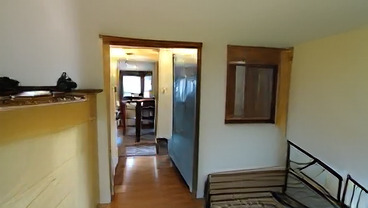}&\includegraphics[width=0.14\linewidth,height=0.08\linewidth]{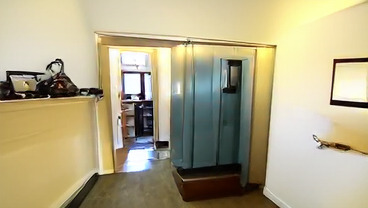}\\%
\includegraphics[width=0.14\linewidth,height=0.08\linewidth]{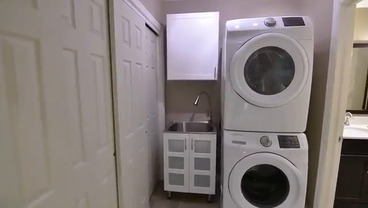}&\includegraphics[width=0.14\linewidth,height=0.08\linewidth]{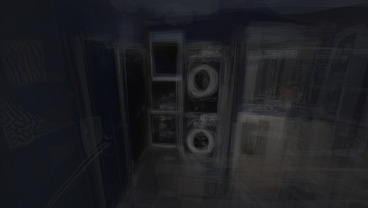}&\includegraphics[width=0.14\linewidth,height=0.08\linewidth]{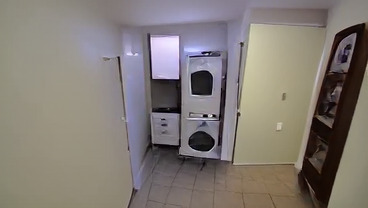}&\includegraphics[width=0.14\linewidth,height=0.08\linewidth]{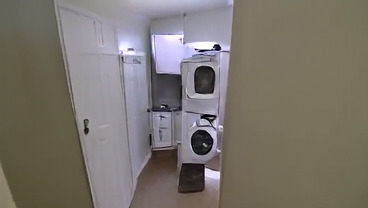}&\includegraphics[width=0.14\linewidth,height=0.08\linewidth]{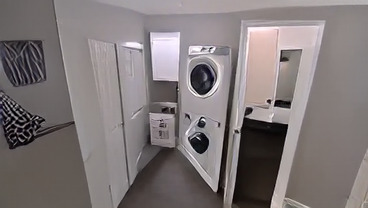}&\includegraphics[width=0.14\linewidth,height=0.08\linewidth]{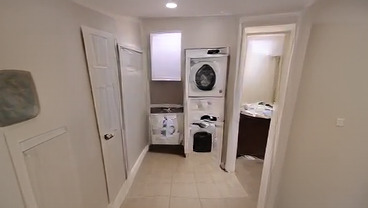}&\includegraphics[width=0.14\linewidth,height=0.08\linewidth]{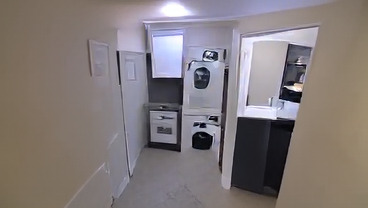}\\%
\includegraphics[width=0.14\linewidth,height=0.08\linewidth]{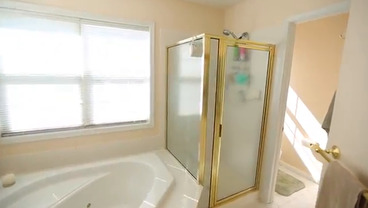}&\includegraphics[width=0.14\linewidth,height=0.08\linewidth]{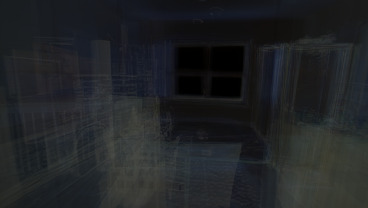}&\includegraphics[width=0.14\linewidth,height=0.08\linewidth]{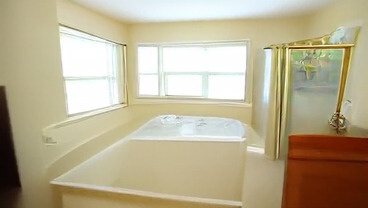}&\includegraphics[width=0.14\linewidth,height=0.08\linewidth]{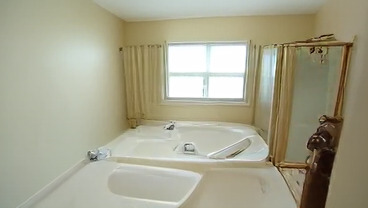}&\includegraphics[width=0.14\linewidth,height=0.08\linewidth]{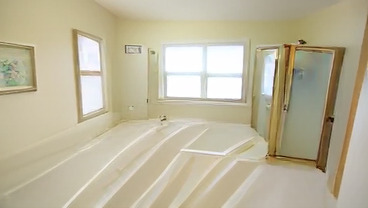}&\includegraphics[width=0.14\linewidth,height=0.08\linewidth]{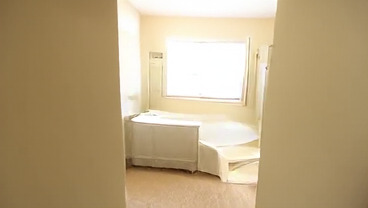}&\includegraphics[width=0.14\linewidth,height=0.08\linewidth]{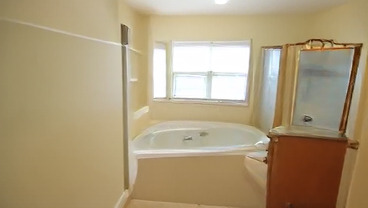}\\%
\includegraphics[width=0.14\linewidth,height=0.08\linewidth]{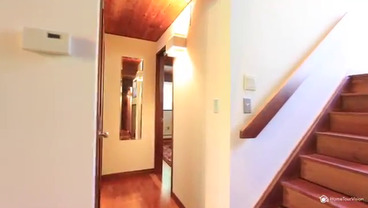}&\includegraphics[width=0.14\linewidth,height=0.08\linewidth]{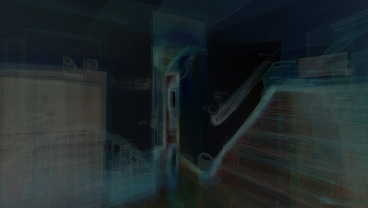}&\includegraphics[width=0.14\linewidth,height=0.08\linewidth]{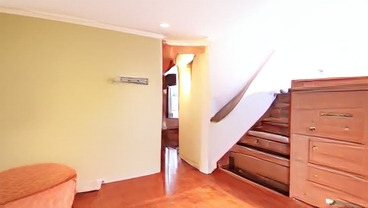}&\includegraphics[width=0.14\linewidth,height=0.08\linewidth]{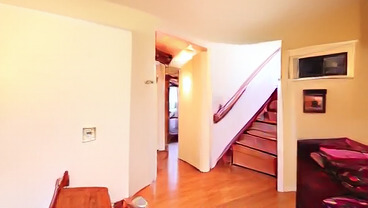}&\includegraphics[width=0.14\linewidth,height=0.08\linewidth]{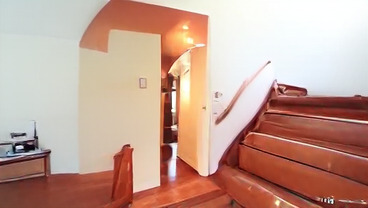}&\includegraphics[width=0.14\linewidth,height=0.08\linewidth]{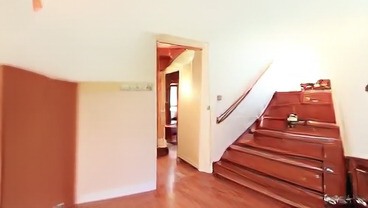}&\includegraphics[width=0.14\linewidth,height=0.08\linewidth]{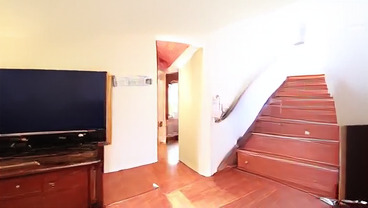}\\%
\includegraphics[width=0.14\linewidth,height=0.08\linewidth]{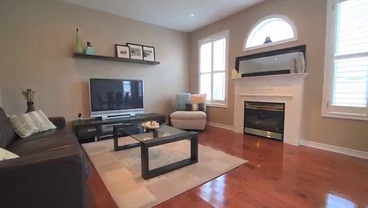}&\includegraphics[width=0.14\linewidth,height=0.08\linewidth]{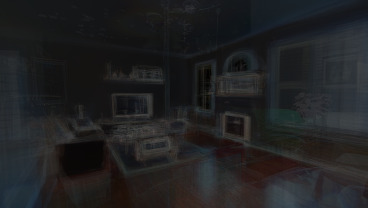}&\includegraphics[width=0.14\linewidth,height=0.08\linewidth]{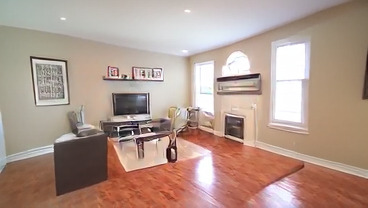}&\includegraphics[width=0.14\linewidth,height=0.08\linewidth]{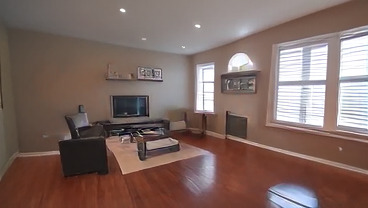}&\includegraphics[width=0.14\linewidth,height=0.08\linewidth]{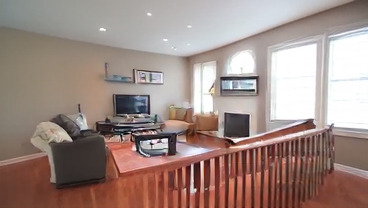}&\includegraphics[width=0.14\linewidth,height=0.08\linewidth]{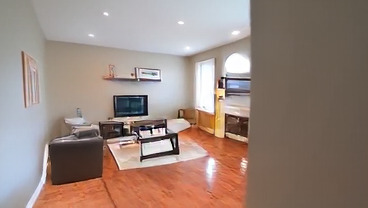}&\includegraphics[width=0.14\linewidth,height=0.08\linewidth]{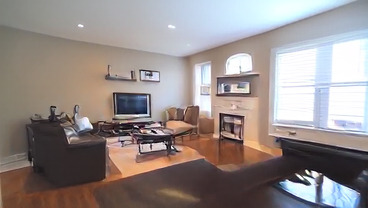}\\%
\includegraphics[width=0.14\linewidth,height=0.08\linewidth]{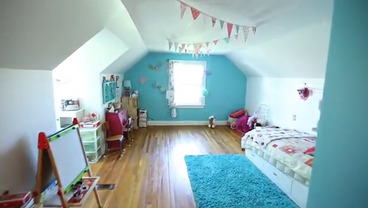}&\includegraphics[width=0.14\linewidth,height=0.08\linewidth]{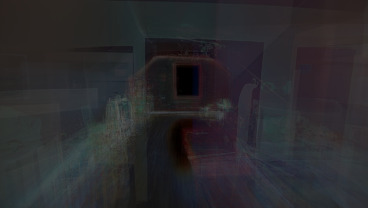}&\includegraphics[width=0.14\linewidth,height=0.08\linewidth]{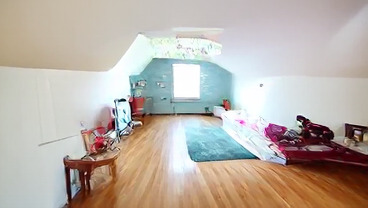}&\includegraphics[width=0.14\linewidth,height=0.08\linewidth]{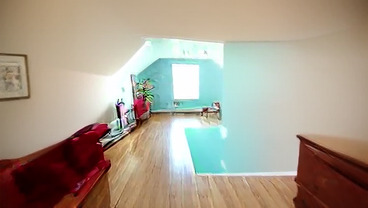}&\includegraphics[width=0.14\linewidth,height=0.08\linewidth]{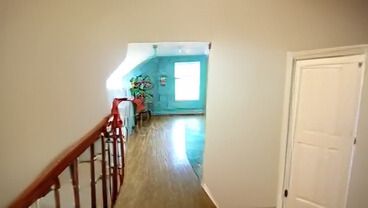}&\includegraphics[width=0.14\linewidth,height=0.08\linewidth]{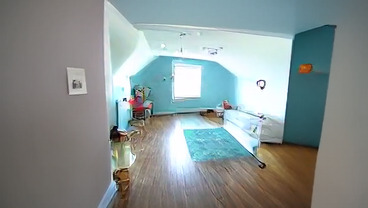}&\includegraphics[width=0.14\linewidth,height=0.08\linewidth]{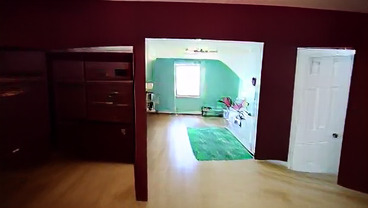}\\%
\includegraphics[width=0.14\linewidth,height=0.08\linewidth]{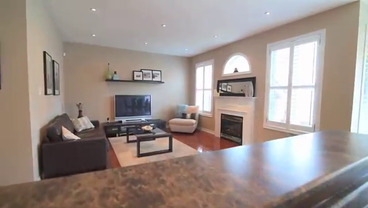}&\includegraphics[width=0.14\linewidth,height=0.08\linewidth]{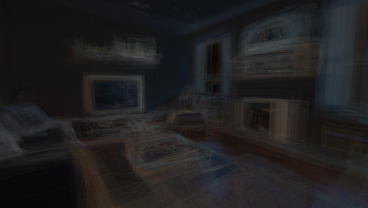}&\includegraphics[width=0.14\linewidth,height=0.08\linewidth]{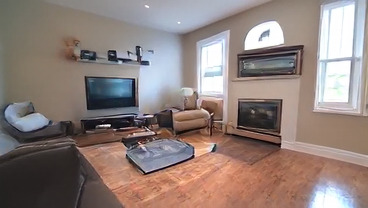}&\includegraphics[width=0.14\linewidth,height=0.08\linewidth]{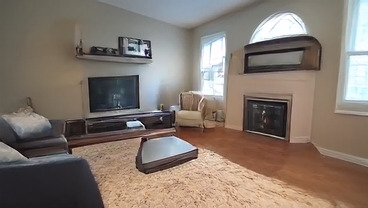}&\includegraphics[width=0.14\linewidth,height=0.08\linewidth]{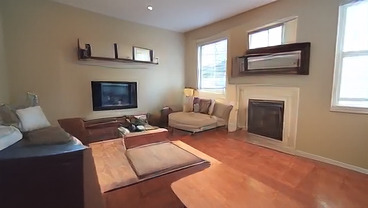}&\includegraphics[width=0.14\linewidth,height=0.08\linewidth]{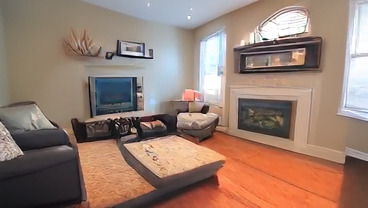}&\includegraphics[width=0.14\linewidth,height=0.08\linewidth]{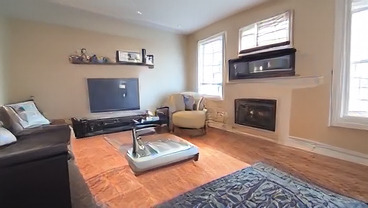}\\\bottomrule%
\end{tabular}
}
   \end{center}
     \vspace{-0.75em}
      \caption{Additional samples on RealEstate10K. The second column depicts the pixel-wise 
      standard deviation $\sigma$ obtained from $n=32$ samples. 
      }
   \label{fig:supprealestatesamples}
   \end{figure*}
}

\newcommand{\figsuppacidsamples}        {
   \begin{figure*}[bthp]
   \begin{center}
     \resizebox{\linewidth}{!}{%
       \setlength{\tabcolsep}{0.1em}%
\begin{tabular}{@{}cc@{\hskip 0.75em}ccccc@{}}%
\toprule%
Source&$\sigma$&\multicolumn{5}{c}{samples of variant \varv{}}\\%
\midrule%
\includegraphics[width=0.14\linewidth,height=0.08\linewidth]{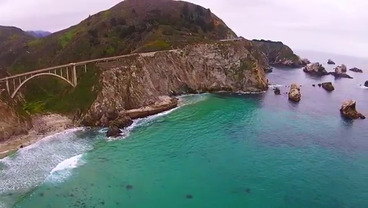}&\includegraphics[width=0.14\linewidth,height=0.08\linewidth]{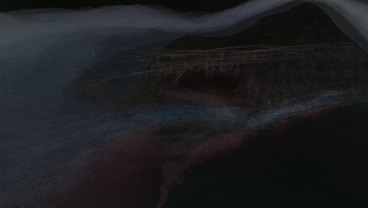}&\includegraphics[width=0.14\linewidth,height=0.08\linewidth]{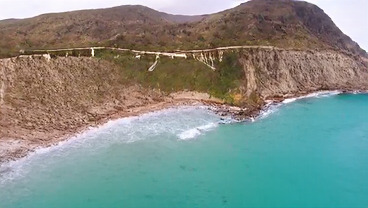}&\includegraphics[width=0.14\linewidth,height=0.08\linewidth]{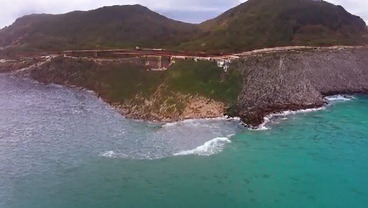}&\includegraphics[width=0.14\linewidth,height=0.08\linewidth]{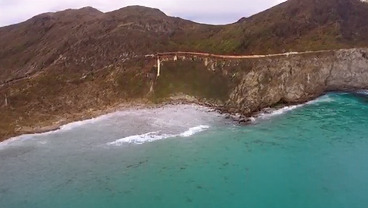}&\includegraphics[width=0.14\linewidth,height=0.08\linewidth]{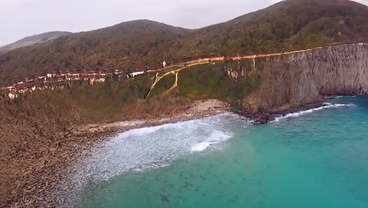}&\includegraphics[width=0.14\linewidth,height=0.08\linewidth]{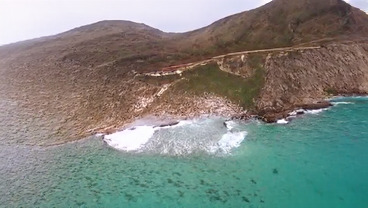}\\%
\includegraphics[width=0.14\linewidth,height=0.08\linewidth]{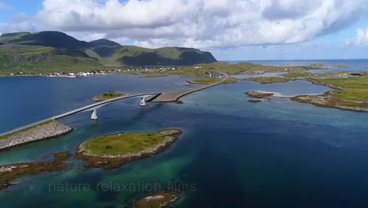}&\includegraphics[width=0.14\linewidth,height=0.08\linewidth]{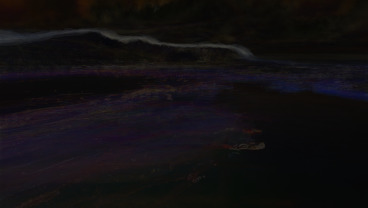}&\includegraphics[width=0.14\linewidth,height=0.08\linewidth]{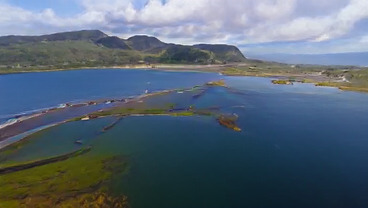}&\includegraphics[width=0.14\linewidth,height=0.08\linewidth]{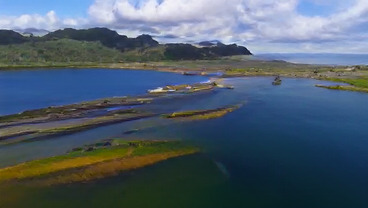}&\includegraphics[width=0.14\linewidth,height=0.08\linewidth]{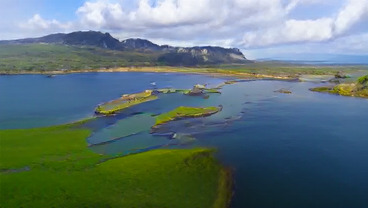}&\includegraphics[width=0.14\linewidth,height=0.08\linewidth]{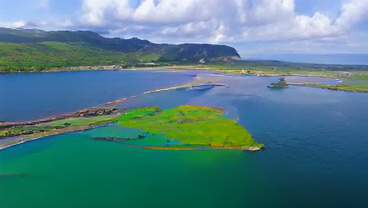}&\includegraphics[width=0.14\linewidth,height=0.08\linewidth]{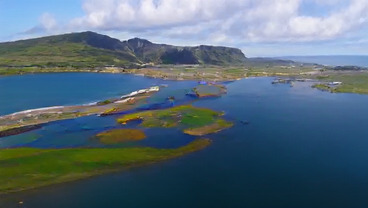}\\%
\includegraphics[width=0.14\linewidth,height=0.08\linewidth]{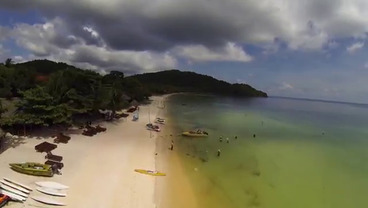}&\includegraphics[width=0.14\linewidth,height=0.08\linewidth]{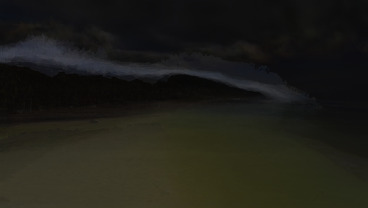}&\includegraphics[width=0.14\linewidth,height=0.08\linewidth]{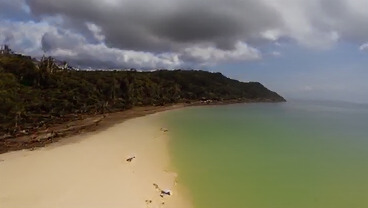}&\includegraphics[width=0.14\linewidth,height=0.08\linewidth]{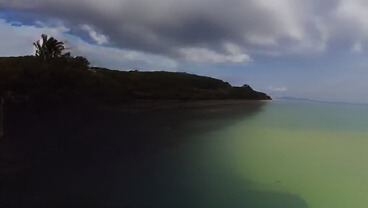}&\includegraphics[width=0.14\linewidth,height=0.08\linewidth]{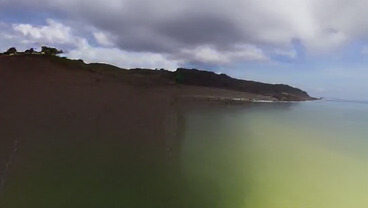}&\includegraphics[width=0.14\linewidth,height=0.08\linewidth]{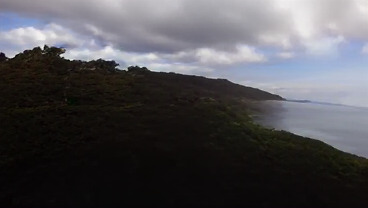}&\includegraphics[width=0.14\linewidth,height=0.08\linewidth]{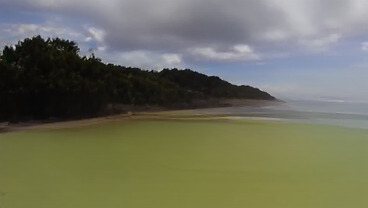}\\%
\includegraphics[width=0.14\linewidth,height=0.08\linewidth]{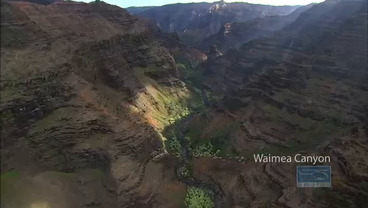}&\includegraphics[width=0.14\linewidth,height=0.08\linewidth]{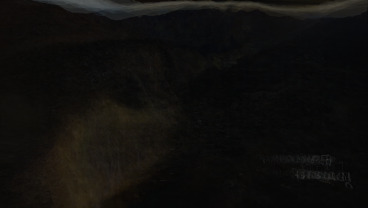}&\includegraphics[width=0.14\linewidth,height=0.08\linewidth]{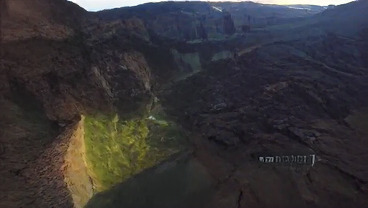}&\includegraphics[width=0.14\linewidth,height=0.08\linewidth]{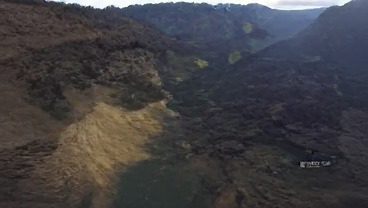}&\includegraphics[width=0.14\linewidth,height=0.08\linewidth]{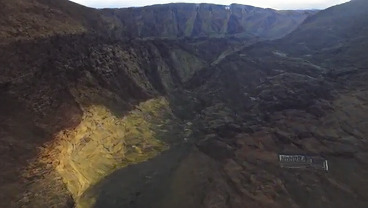}&\includegraphics[width=0.14\linewidth,height=0.08\linewidth]{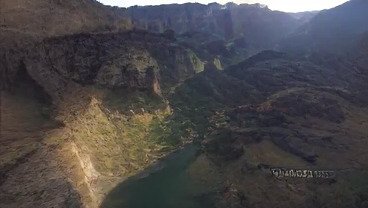}&\includegraphics[width=0.14\linewidth,height=0.08\linewidth]{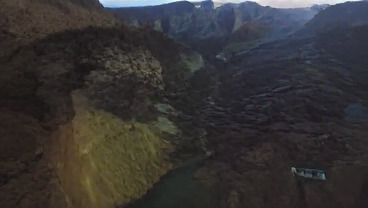}\\%
\includegraphics[width=0.14\linewidth,height=0.08\linewidth]{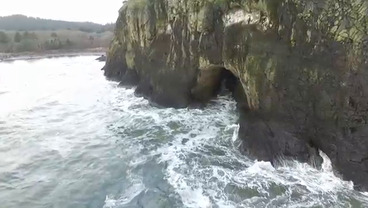}&\includegraphics[width=0.14\linewidth,height=0.08\linewidth]{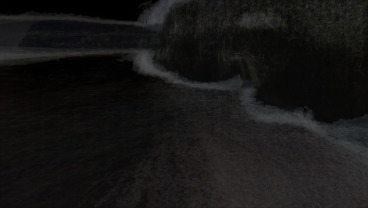}&\includegraphics[width=0.14\linewidth,height=0.08\linewidth]{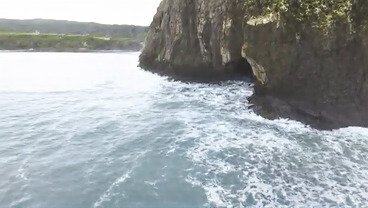}&\includegraphics[width=0.14\linewidth,height=0.08\linewidth]{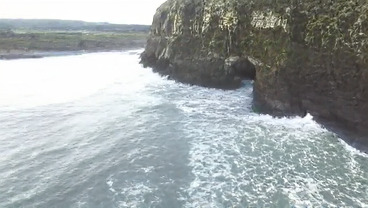}&\includegraphics[width=0.14\linewidth,height=0.08\linewidth]{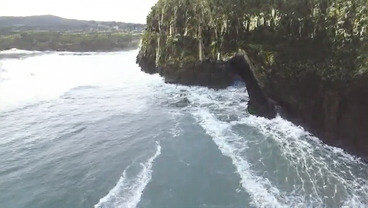}&\includegraphics[width=0.14\linewidth,height=0.08\linewidth]{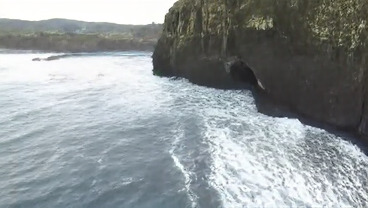}&\includegraphics[width=0.14\linewidth,height=0.08\linewidth]{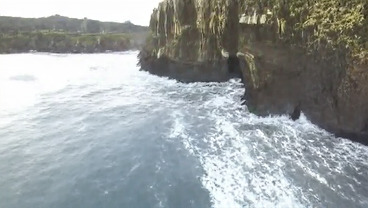}\\%
\includegraphics[width=0.14\linewidth,height=0.08\linewidth]{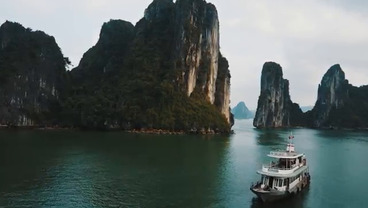}&\includegraphics[width=0.14\linewidth,height=0.08\linewidth]{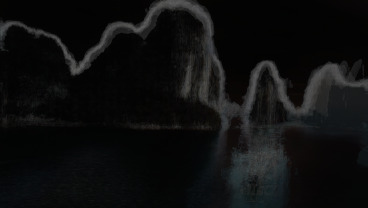}&\includegraphics[width=0.14\linewidth,height=0.08\linewidth]{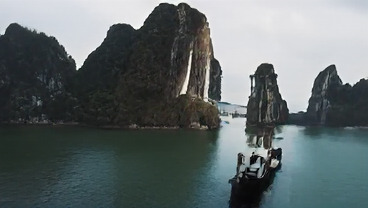}&\includegraphics[width=0.14\linewidth,height=0.08\linewidth]{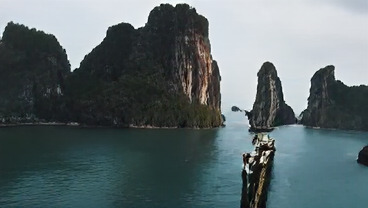}&\includegraphics[width=0.14\linewidth,height=0.08\linewidth]{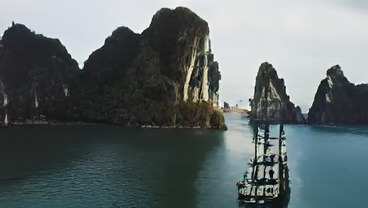}&\includegraphics[width=0.14\linewidth,height=0.08\linewidth]{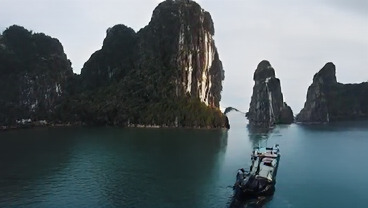}&\includegraphics[width=0.14\linewidth,height=0.08\linewidth]{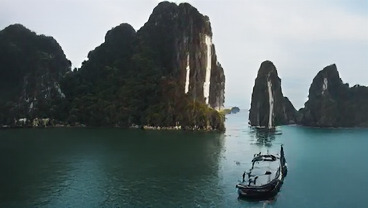}\\%
\includegraphics[width=0.14\linewidth,height=0.08\linewidth]{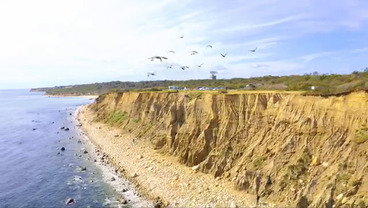}&\includegraphics[width=0.14\linewidth,height=0.08\linewidth]{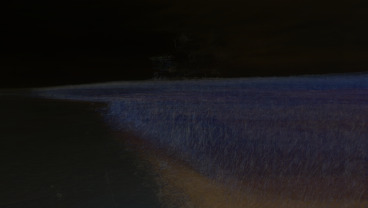}&\includegraphics[width=0.14\linewidth,height=0.08\linewidth]{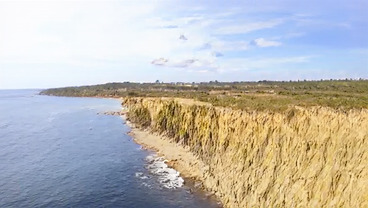}&\includegraphics[width=0.14\linewidth,height=0.08\linewidth]{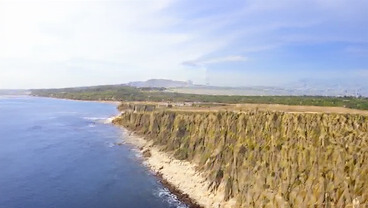}&\includegraphics[width=0.14\linewidth,height=0.08\linewidth]{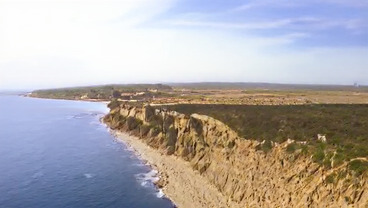}&\includegraphics[width=0.14\linewidth,height=0.08\linewidth]{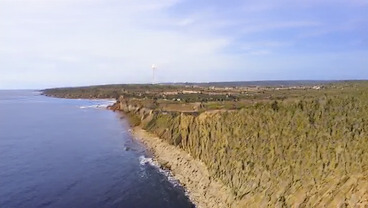}&\includegraphics[width=0.14\linewidth,height=0.08\linewidth]{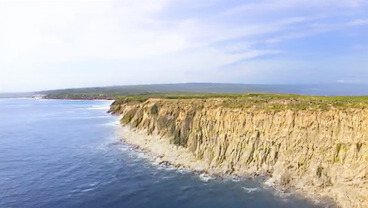}\\%
\includegraphics[width=0.14\linewidth,height=0.08\linewidth]{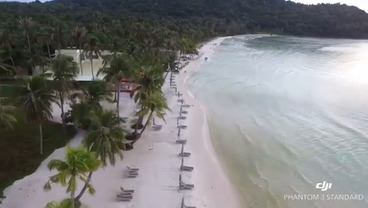}&\includegraphics[width=0.14\linewidth,height=0.08\linewidth]{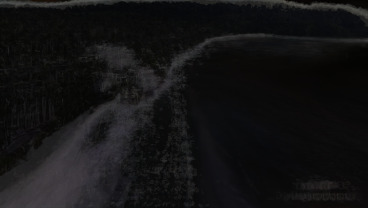}&\includegraphics[width=0.14\linewidth,height=0.08\linewidth]{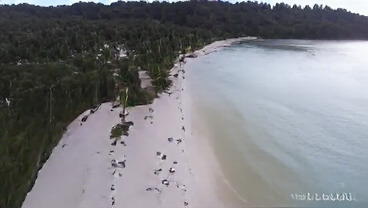}&\includegraphics[width=0.14\linewidth,height=0.08\linewidth]{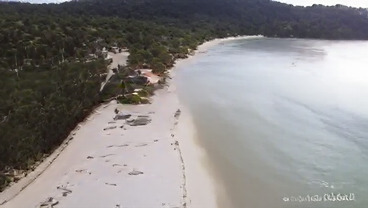}&\includegraphics[width=0.14\linewidth,height=0.08\linewidth]{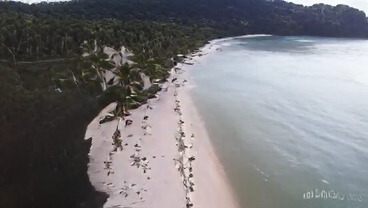}&\includegraphics[width=0.14\linewidth,height=0.08\linewidth]{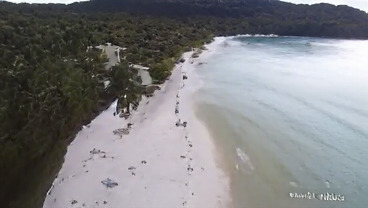}&\includegraphics[width=0.14\linewidth,height=0.08\linewidth]{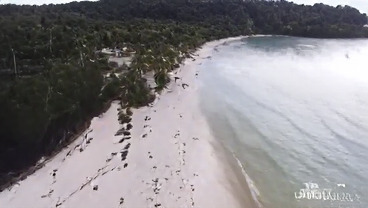}\\%
\includegraphics[width=0.14\linewidth,height=0.08\linewidth]{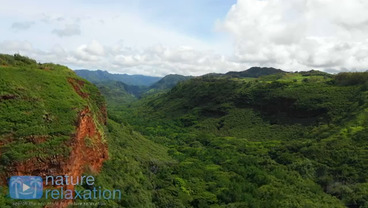}&\includegraphics[width=0.14\linewidth,height=0.08\linewidth]{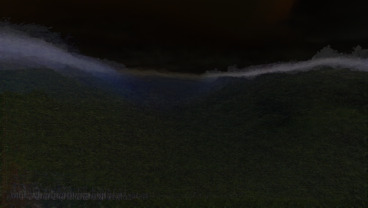}&\includegraphics[width=0.14\linewidth,height=0.08\linewidth]{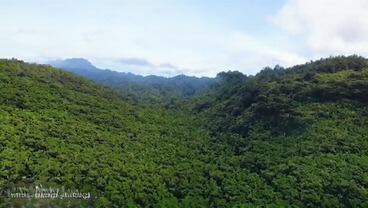}&\includegraphics[width=0.14\linewidth,height=0.08\linewidth]{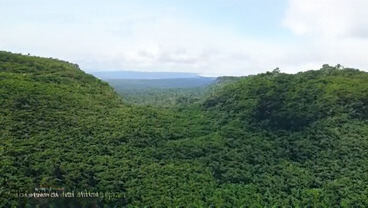}&\includegraphics[width=0.14\linewidth,height=0.08\linewidth]{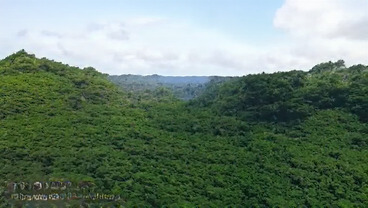}&\includegraphics[width=0.14\linewidth,height=0.08\linewidth]{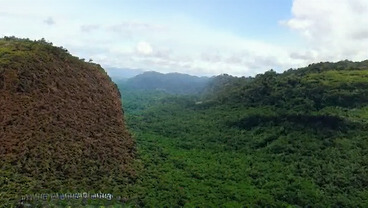}&\includegraphics[width=0.14\linewidth,height=0.08\linewidth]{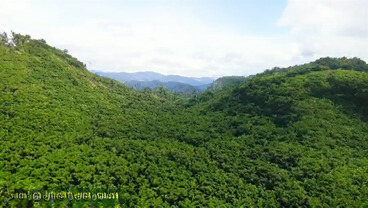}\\%
\includegraphics[width=0.14\linewidth,height=0.08\linewidth]{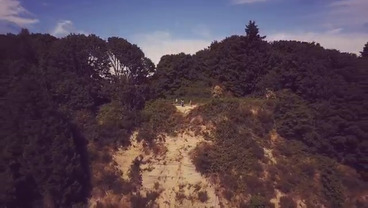}&\includegraphics[width=0.14\linewidth,height=0.08\linewidth]{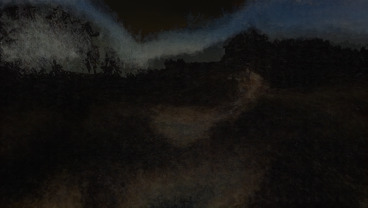}&\includegraphics[width=0.14\linewidth,height=0.08\linewidth]{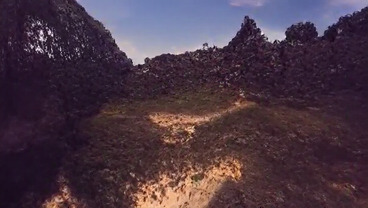}&\includegraphics[width=0.14\linewidth,height=0.08\linewidth]{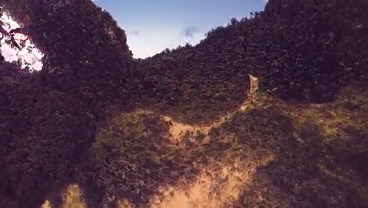}&\includegraphics[width=0.14\linewidth,height=0.08\linewidth]{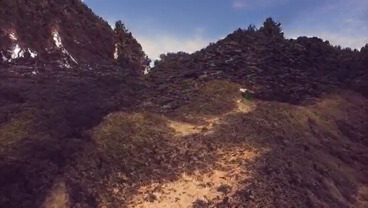}&\includegraphics[width=0.14\linewidth,height=0.08\linewidth]{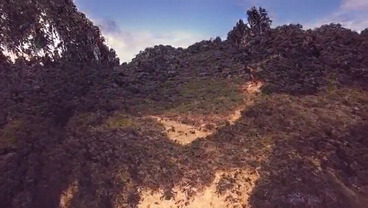}&\includegraphics[width=0.14\linewidth,height=0.08\linewidth]{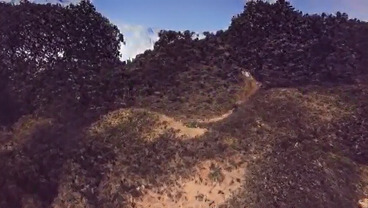}\\%
\includegraphics[width=0.14\linewidth,height=0.08\linewidth]{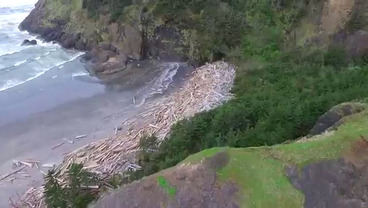}&\includegraphics[width=0.14\linewidth,height=0.08\linewidth]{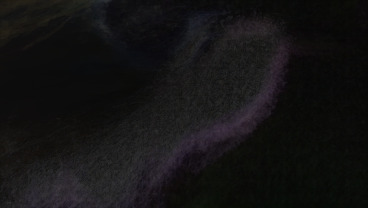}&\includegraphics[width=0.14\linewidth,height=0.08\linewidth]{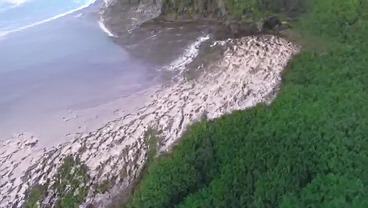}&\includegraphics[width=0.14\linewidth,height=0.08\linewidth]{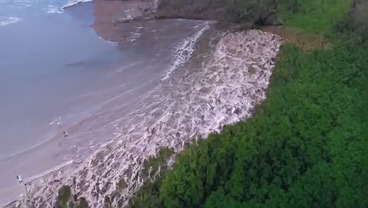}&\includegraphics[width=0.14\linewidth,height=0.08\linewidth]{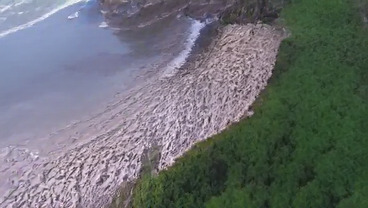}&\includegraphics[width=0.14\linewidth,height=0.08\linewidth]{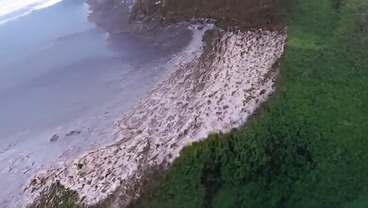}&\includegraphics[width=0.14\linewidth,height=0.08\linewidth]{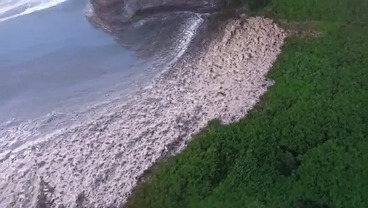}\\%
\includegraphics[width=0.14\linewidth,height=0.08\linewidth]{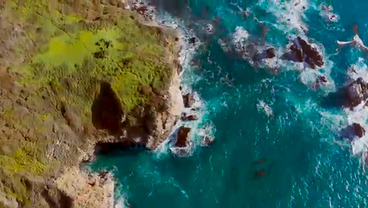}&\includegraphics[width=0.14\linewidth,height=0.08\linewidth]{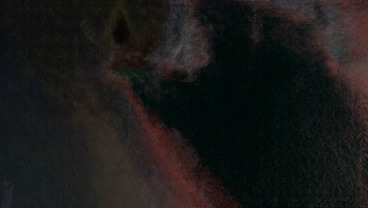}&\includegraphics[width=0.14\linewidth,height=0.08\linewidth]{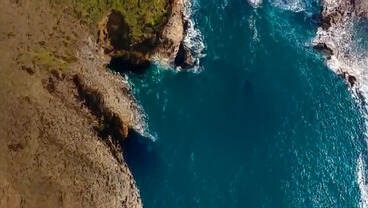}&\includegraphics[width=0.14\linewidth,height=0.08\linewidth]{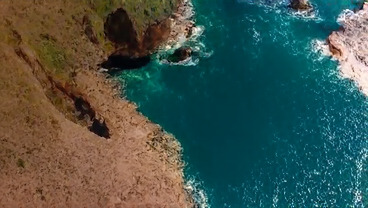}&\includegraphics[width=0.14\linewidth,height=0.08\linewidth]{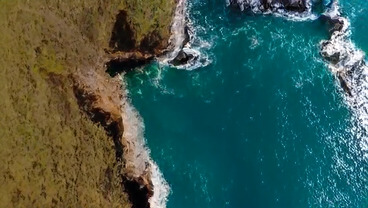}&\includegraphics[width=0.14\linewidth,height=0.08\linewidth]{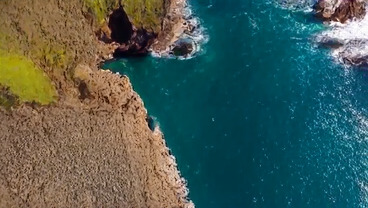}&\includegraphics[width=0.14\linewidth,height=0.08\linewidth]{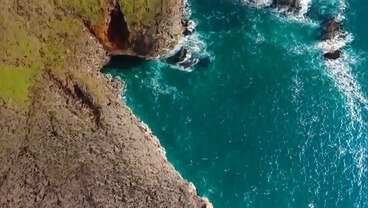}\\%
\includegraphics[width=0.14\linewidth,height=0.08\linewidth]{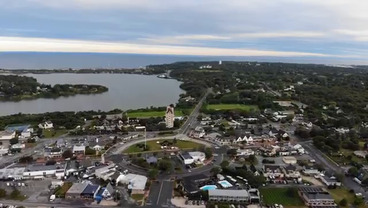}&\includegraphics[width=0.14\linewidth,height=0.08\linewidth]{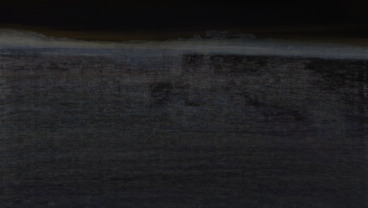}&\includegraphics[width=0.14\linewidth,height=0.08\linewidth]{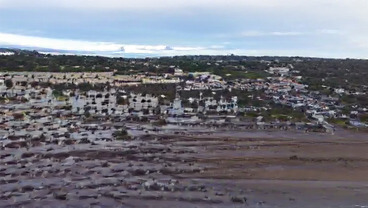}&\includegraphics[width=0.14\linewidth,height=0.08\linewidth]{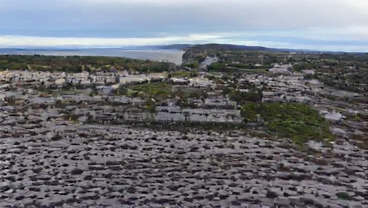}&\includegraphics[width=0.14\linewidth,height=0.08\linewidth]{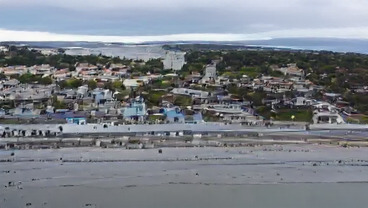}&\includegraphics[width=0.14\linewidth,height=0.08\linewidth]{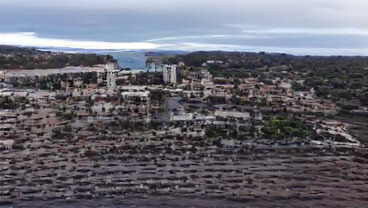}&\includegraphics[width=0.14\linewidth,height=0.08\linewidth]{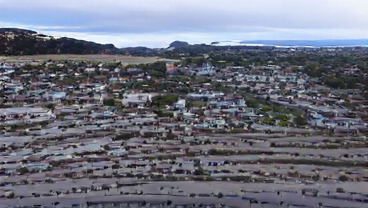}\\%
\includegraphics[width=0.14\linewidth,height=0.08\linewidth]{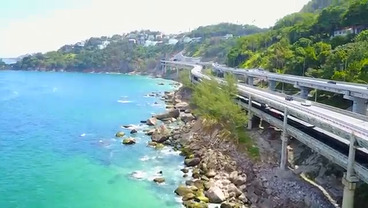}&\includegraphics[width=0.14\linewidth,height=0.08\linewidth]{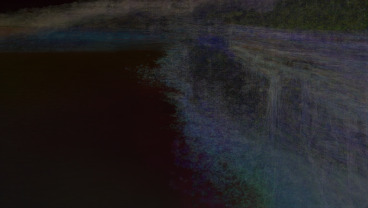}&\includegraphics[width=0.14\linewidth,height=0.08\linewidth]{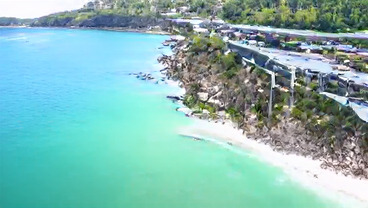}&\includegraphics[width=0.14\linewidth,height=0.08\linewidth]{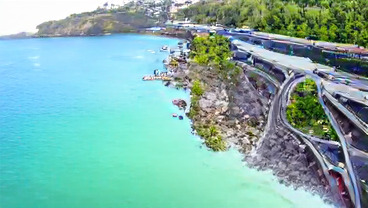}&\includegraphics[width=0.14\linewidth,height=0.08\linewidth]{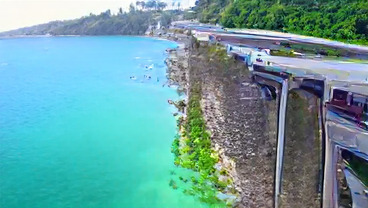}&\includegraphics[width=0.14\linewidth,height=0.08\linewidth]{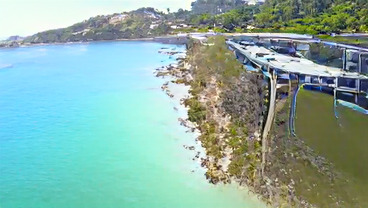}&\includegraphics[width=0.14\linewidth,height=0.08\linewidth]{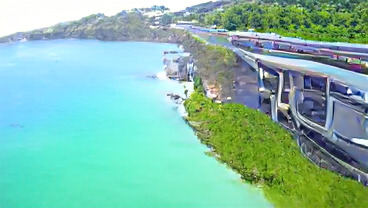}\\\bottomrule%
\end{tabular}
}
   \end{center}
     \vspace{-0.75em}
      \caption{Additional samples on ACID. The second column depicts the pixel-wise 
      standard deviation $\sigma$ obtained from $n=32$ samples. 
      }
   \label{fig:suppacidsamples}
   \end{figure*}
}

\newcommand{\figdepthprobesqualitativesupp}{
\begin{figure*}[t]
  \renewcommand{\impath}[1]{img/depthprobes/img/layer0/src_image/##1}
  \renewcommand{\impatha}[1]{img/depthprobes/img/layer0/depth_prediction/##1}
  \renewcommand{\impathb}[1]{img/depthprobes/img/layer1/depth_prediction/##1}
  \renewcommand{\impathc}[1]{img/depthprobes/img/layer4/depth_prediction/##1}
  \renewcommand{\impathe}[1]{img/depthprobes/img/layer5/depth_prediction/##1}
  \renewcommand{\impathd}[1]{img/depthprobes/img/layer4/depth_reconstruction/##1}
  \renewcommand{\imwidth}{0.16\textwidth}
  \setlength{\tabcolsep}{0.1pt}
\centering
\begin{tabular}{c cccc c}
\footnotesize{Input} & \footnotesize{Layer \#0} & \footnotesize{Layer \#1} & \footnotesize{\textbf{Layer \#4}} & \footnotesize{Layer \#5} & \footnotesize{Depth Rec.} \\
\toprule
	\includegraphics[width=\imwidth, align=c]{\impath{115_src_image_000276}} &	         
	\includegraphics[width=\imwidth, align=c]{\impatha{115_depth_prediction_000276}} &	         
	\includegraphics[width=\imwidth, align=c]{\impathb{115_depth_prediction_000276}} &	         
	\includegraphics[width=\imwidth, align=c]{\impathc{115_depth_prediction_000276}} &	         
	\includegraphics[width=\imwidth, align=c]{\impathe{115_depth_prediction_000276}} &	         
	\includegraphics[width=\imwidth, align=c]{\impathd{115_depth_reconstruction_000276}} \\

	\includegraphics[width=\imwidth, align=c]{\impath{123_src_image_000212}} &	         
	\includegraphics[width=\imwidth, align=c]{\impatha{123_depth_prediction_000212}} &	         
	\includegraphics[width=\imwidth, align=c]{\impathb{123_depth_prediction_000212}} &	         
	\includegraphics[width=\imwidth, align=c]{\impathc{123_depth_prediction_000212}} &	         
	\includegraphics[width=\imwidth, align=c]{\impathe{123_depth_prediction_000212}} &	         
	\includegraphics[width=\imwidth, align=c]{\impathd{123_depth_reconstruction_000212}} \\

	\includegraphics[width=\imwidth, align=c]{\impath{140_src_image_000095}} &	         
	\includegraphics[width=\imwidth, align=c]{\impatha{140_depth_prediction_000095}} &	         
	\includegraphics[width=\imwidth, align=c]{\impathb{140_depth_prediction_000095}} &	         
	\includegraphics[width=\imwidth, align=c]{\impathc{140_depth_prediction_000095}} &	         
	\includegraphics[width=\imwidth, align=c]{\impathe{140_depth_prediction_000095}} &	         
	\includegraphics[width=\imwidth, align=c]{\impathd{140_depth_reconstruction_000095}} \\

	\includegraphics[width=\imwidth, align=c]{\impath{158_src_image_000330}} &	         
	\includegraphics[width=\imwidth, align=c]{\impatha{158_depth_prediction_000330}} &	         
	\includegraphics[width=\imwidth, align=c]{\impathb{158_depth_prediction_000330}} &	         
	\includegraphics[width=\imwidth, align=c]{\impathc{158_depth_prediction_000330}} &	         
	\includegraphics[width=\imwidth, align=c]{\impathe{158_depth_prediction_000330}} &	         
	\includegraphics[width=\imwidth, align=c]{\impathd{158_depth_reconstruction_000330}} \\
	
	\includegraphics[width=\imwidth, align=c]{\impath{116_src_image_000422}} &	         
	\includegraphics[width=\imwidth, align=c]{\impatha{116_depth_prediction_000422}} &	         
	\includegraphics[width=\imwidth, align=c]{\impathb{116_depth_prediction_000422}} &	         
	\includegraphics[width=\imwidth, align=c]{\impathc{116_depth_prediction_000422}} &	         
	\includegraphics[width=\imwidth, align=c]{\impathe{116_depth_prediction_000422}} &	         
	\includegraphics[width=\imwidth, align=c]{\impathd{116_depth_reconstruction_000422}} \\
	
	\includegraphics[width=\imwidth, align=c]{\impath{121_src_image_000583}} &	         
	\includegraphics[width=\imwidth, align=c]{\impatha{121_depth_prediction_000583}} &	         
	\includegraphics[width=\imwidth, align=c]{\impathb{121_depth_prediction_000583}} &	         
	\includegraphics[width=\imwidth, align=c]{\impathc{121_depth_prediction_000583}} &	         
	\includegraphics[width=\imwidth, align=c]{\impathe{121_depth_prediction_000583}} &	         
	\includegraphics[width=\imwidth, align=c]{\impathd{121_depth_reconstruction_000583}} \\
\end{tabular}
\caption{
  Additional results on linearly probed depth maps for different transformer layers as in Fig.~\ref{fig:figdepthprobesqualitative}. See Sec.~\ref{subsec:extractthedepth}.
}
\label{fig:figdepthprobesqualitativesupp}
\end{figure*}
}

\newcommand{\tabapproachtable}{
\begin{table}
\begin{center}
\caption{Summarizing our approach to encoding inductive biases.}
\label{tab:approachtable}
\begin{footnotesize}
\resizebox{.5\textwidth}{!}{%
\begin{tabular}{l c c c c}
\toprule
method & \makecell{explicit \\ warp} & \makecell{requires \\ depth} & \makecell{warped \\ features} & $f(\xsrc, \cam)$ \\
\midrule
\vari{} & \cmark & \cmark & $\xsrc$ & Eq.\eqref{eq:fvari} \\
\varii{} & \cmark & \cmark & $\encoder(\xsrc)$ & Eq.\eqref{eq:fvarii} \\
\variii{} & \cmark & \cmark & $e(\encoder(\xsrc)),\,e^{\text{pos}}$ & Eq.\eqref{eq:fvariii} \\
\variv{} & \xmark & \cmark & -- & Eq.\eqref{eq:fvariv} \\
\varv{} & \xmark & \cmark & -- & Eq.\eqref{eq:fvarv} \\
\varvi{} & \xmark & \xmark & -- & Eq.\eqref{eq:fvarvi} \\
\varvii{} & \cmark & \cmark & $e(\encoder(\xsrc)),\,e^{\text{pos}}$ & Eq.\eqref{eq:fvariii}+Eq.\eqref{eq:fvarv} \\
\bottomrule
\end{tabular}%
}
\end{footnotesize}
\end{center}
\end{table}
}

\newcommand{\tabrealallcomp}{
\begin{table}
\centering
\caption{All RealEstate.}
\label{tab:realallcomp}
\begin{tabular}{@{}lrrrrr@{}}%
\toprule%
method&FID ↓&IS ↑&PSIM ↓&SSIM ↑&PSNR ↑\\%
\midrule%
geo nodepth&\textbf{\underline{48.59}}&4.24 $\pm$0.30&3.17 $\pm$0.43&0.42 $\pm$0.13&12.16 $\pm$2.54\\%
geo depth emb&\textbf{48.84}&4.17 $\pm$0.33&\underline{3.09} $\pm$0.46&0.44 $\pm$0.13&12.51 $\pm$2.69\\%
geo depth&\underline{49.15}&4.17 $\pm$0.52&\textbf{3.05} $\pm$0.46&0.44 $\pm$0.13&12.66 $\pm$2.68\\%
wrp image&49.63&4.26 $\pm$0.38&3.18 $\pm$0.46&0.43 $\pm$0.13&12.11 $\pm$2.66\\%
geo catdepth&50.04&4.36 $\pm$0.33&3.10 $\pm$0.45&0.43 $\pm$0.13&12.43 $\pm$2.66\\%
wrp emb&50.35&4.04 $\pm$0.39&3.15 $\pm$0.45&0.43 $\pm$0.13&12.30 $\pm$2.66\\%
wrp feature&54.82&4.17 $\pm$0.53&3.31 $\pm$0.43&0.41 $\pm$0.13&11.75 $\pm$2.58\\%
vqwarper&66.66&\underline{4.47} $\pm$0.49&\textbf{\underline{2.97}} $\pm$0.55&0.42 $\pm$0.15&\textbf{\underline{13.60}} $\pm$2.56\\%
3D Photo&85.43&\textbf{5.10} $\pm$0.39&3.20 $\pm$0.54&\textbf{\underline{0.49}} $\pm$0.12&12.80 $\pm$2.33\\%
SynSin&113.88&3.70 $\pm$0.30&3.30 $\pm$0.51&\textbf{0.47} $\pm$0.13&\underline{12.87} $\pm$2.46\\%
MiDaS&132.13&\textbf{\underline{5.63}} $\pm$0.77&3.38 $\pm$0.56&\underline{0.46} $\pm$0.15&\textbf{13.09} $\pm$2.16\\\bottomrule%
\end{tabular}
\end{table*}
}

\newcommand{\tabmergedbiascomp}{
\begin{table*}[t]
\centering
  \caption{To assess the effect of encoding different degrees of 3D prior knowledge,
  we evaluate all variants 
  on RealEstate and ACID
  using
   negative log-likelihood (NLL), 
  FID \cite{DBLP:conf/nips/HeuselRUNH17} and PSIM \cite{DBLP:conf/cvpr/ZhangIESW18}, PSNR and 
  SSIM \cite{DBLP:journals/tip/WangBSS04}. We highlight
  \textbf{\underline{best}}, \textbf{second best} and \underline{third best}
  scores.
  }
\label{tab:mergedbiascomp}
\begin{footnotesize}
  \begin{tabular}{lrrrrr rrrrr}%
\toprule%
&\multicolumn{5}{c}{RealEstate10K}&\multicolumn{5}{c}{ACID} \\%
\cmidrule(r{0.5em}){2-6}
\cmidrule(l{0.5em}){7-11}
method&FID ↓&NLL ↓&PSIM ↓&SSIM ↑&PSNR ↑ &FID ↓&NLL ↓&PSIM ↓&SSIM ↑&PSNR ↑\\%
\midrule%

\varvi{}&\textbf{\underline{48.59}}&4.956 &3.17 \tiny{$\pm$0.43}&0.42 \tiny{$\pm$0.13}&12.16 \tiny{$\pm$2.54}
& \textbf{\underline{42.88}}&5.365 \tiny{$\pm$0.007}&2.90 \tiny{$\pm$0.53}&0.40 \tiny{$\pm$0.15}&15.17 \tiny{$\pm$3.40} \\%

\varvii{}&\textbf{48.84}& \underline{4.913} &\textbf{3.09} \tiny{$\pm$0.46}&\textbf{0.44} \tiny{$\pm$0.13}&\textbf{12.51} \tiny{$\pm$2.69}&\underline{44.47}&\textbf{\underline{5.341}} \tiny{$\pm$0.008}&\textbf{\underline{2.83}} \tiny{$\pm$0.54}&0.41 \tiny{$\pm$0.15}&\textbf{\underline{15.54}} \tiny{$\pm$3.52}\\%

\varv{}&\underline{49.15}&\textbf{\underline{4.836}}&\textbf{\underline{3.05}} \tiny{$\pm$0.46}&\textbf{\underline{0.44}} \tiny{$\pm$0.13}&\textbf{\underline{12.66}} \tiny{$\pm$2.68}
&\textbf{42.93}&\underline{5.353} \tiny{$\pm$0.011}&\textbf{2.86} \tiny{$\pm$0.52}&\underline{0.41} \tiny{$\pm$0.15}&\underline{15.53} \tiny{$\pm$3.34}\\%

\vari{}&49.63&4.924&3.18 \tiny{$\pm$0.46}&0.43 \tiny{$\pm$0.13}&12.11 \tiny{$\pm$2.66}
&47.72&5.414 \tiny{$\pm$0.006}&3.00 \tiny{$\pm$0.51}&0.40 \tiny{$\pm$0.14}&14.83 \tiny{$\pm$3.20}\\%

\variv{}&50.04&\textbf{4.860}&\underline{3.10} \tiny{$\pm$0.45}&\underline{0.43} \tiny{$\pm$0.13}&\underline{12.43} \tiny{$\pm$2.66}
&47.44&\textbf{5.350} \tiny{$\pm$0.004}&\underline{2.86} \tiny{$\pm$0.55}&\textbf{\underline{0.42}} \tiny{$\pm$0.15}&\textbf{15.54} \tiny{$\pm$3.57}\\%
\variii{}&50.35 & 5.004 & 3.15 \tiny{$\pm$0.45}&0.43 \tiny{$\pm$0.13}&12.30 \tiny{$\pm$2.66}
&47.08&5.416 \tiny{$\pm$0.007}&2.88 \tiny{$\pm$0.54}&\textbf{0.42} \tiny{$\pm$0.15}&15.45 \tiny{$\pm$3.61}\\%

\varii{}&54.82&5.159 &3.31 \tiny{$\pm$0.43}&0.41 \tiny{$\pm$0.13}&11.75 \tiny{$\pm$2.58} & 
52.65&5.657 \tiny{$\pm$0.003}&3.14 \tiny{$\pm$0.52}&0.38 \tiny{$\pm$0.15}&14.06 \tiny{$\pm$3.28}\\

\bottomrule%
\end{tabular}
\end{footnotesize}
\end{table*}
}

\newcommand{\tabrealbiascomp}{
\begin{table}
\centering
  \caption{RealEstate Bias Comparison. We highlight the
  \textbf{\underline{best}}, \textbf{second best} and \underline{third best}
  scores in each column.}
\label{tab:realbiascomp}
\begin{footnotesize}
  \begin{tabular}{@{}lrrrrr@{}}%
\toprule%
method&FID ↓&IS ↑&PSIM ↓&SSIM ↑&PSNR ↑\\%
\midrule%
\varvi{}&\textbf{\underline{48.59}}&\underline{4.24} \tiny{$\pm$0.30}&3.17 \tiny{$\pm$0.43}&0.42 \tiny{$\pm$0.13}&12.16 \tiny{$\pm$2.54}\\%
\varvii{}&\textbf{48.84}&4.17 \tiny{$\pm$0.33}&\textbf{3.09} \tiny{$\pm$0.46}&\textbf{0.44} \tiny{$\pm$0.13}&\textbf{12.51} \tiny{$\pm$2.69}\\%
\varv{}&\underline{49.15}&4.17 \tiny{$\pm$0.52}&\textbf{\underline{3.05}} \tiny{$\pm$0.46}&\textbf{\underline{0.44}} \tiny{$\pm$0.13}&\textbf{\underline{12.66}} \tiny{$\pm$2.68}\\%
\vari{}&49.63&\textbf{4.26} \tiny{$\pm$0.38}&3.18 \tiny{$\pm$0.46}&0.43 \tiny{$\pm$0.13}&12.11 \tiny{$\pm$2.66}\\%
\variv{}&50.04&\textbf{\underline{4.36}} \tiny{$\pm$0.33}&\underline{3.10} \tiny{$\pm$0.45}&\underline{0.43} \tiny{$\pm$0.13}&\underline{12.43} \tiny{$\pm$2.66}\\%
\variii&50.35&4.04 \tiny{$\pm$0.39}&3.15 \tiny{$\pm$0.45}&0.43 \tiny{$\pm$0.13}&12.30 \tiny{$\pm$2.66}\\%
\varii{}&54.82&4.17 \tiny{$\pm$0.53}&3.31 \tiny{$\pm$0.43}&0.41 \tiny{$\pm$0.13}&11.75 \tiny{$\pm$2.58}\\\bottomrule%
\end{tabular}
\end{footnotesize}
\end{table}
}

\newcommand{\tabrealsotacomp}{
\begin{table}
\centering
\caption{Quantitative comparison on RealEstate. Reconstruction
  metrics are reported with 32 samples, see
  Fig.~\ref{fig:reconstructionbothdepth} for other values.
  }
\label{tab:realsotacomp}
\begin{footnotesize}
  \setlength{\tabcolsep}{0.5em}
  \begin{tabular}{@{}lrrrrr@{}}%
\toprule%
method&FID ↓&IS ↑&PSIM ↓&SSIM ↑&PSNR ↑\\%
\midrule%
\varvi{}&\textbf{\underline{48.59}}&4.24 \tiny{$\pm$0.30}&\textbf{2.95} \tiny{$\pm$0.43}&\textbf{0.49} \tiny{$\pm$0.12}&\textbf{14.06} \tiny{$\pm$2.41}\\%
\varv{}&\textbf{49.15}&4.17 \tiny{$\pm$0.52}&\textbf{\underline{2.86}} \tiny{$\pm$0.45}&\textbf{\underline{0.50}} \tiny{$\pm$0.12}&\textbf{\underline{14.47}} \tiny{$\pm$2.51}\\%
\midrule%
\vqwarper{}&\underline{66.66}&\underline{4.47} \tiny{$\pm$0.49}&\underline{2.97} \tiny{$\pm$0.55}&0.42 \tiny{$\pm$0.15}&\underline{13.60} \tiny{$\pm$2.56}\\%
\threedp{}&85.43&\textbf{5.10} \tiny{$\pm$0.39}&3.20 \tiny{$\pm$0.54}&\underline{0.49} \tiny{$\pm$0.12}&12.80 \tiny{$\pm$2.33}\\%
\synsin{}&113.88&3.70 \tiny{$\pm$0.30}&3.30 \tiny{$\pm$0.51}&0.47 \tiny{$\pm$0.13}&12.87 \tiny{$\pm$2.46}\\%
\midas{}&132.13&\textbf{\underline{5.63}} \tiny{$\pm$0.77}&3.38 \tiny{$\pm$0.56}&0.46 \tiny{$\pm$0.15}&13.09 \tiny{$\pm$2.16}\\\bottomrule%
\end{tabular}
\end{footnotesize}
  \vspace{-1.25em}
\end{table}
}

\newcommand{\tabacidallcomp}{
\begin{table*}
\centering
\caption{All ACID.}
\label{tab:acidallcomp}
\begin{footnotesize}
\begin{tabular}{@{}lrrrrr@{}}%
\toprule%
method&FID ↓&IS ↑&PSIM ↓&SSIM ↑&PSNR ↑\\%
\midrule%
geo nodepth&\textbf{\underline{42.88}}&2.63 $\pm$0.14&2.90 $\pm$0.53&0.40 $\pm$0.15&15.17 $\pm$3.40\\%
geo depth&\textbf{42.93}&2.62 $\pm$0.23&\underline{2.86} $\pm$0.52&0.41 $\pm$0.15&15.53 $\pm$3.34\\%
geo depth emb&\underline{44.47}&2.67 $\pm$0.23&\textbf{2.83} $\pm$0.54&0.41 $\pm$0.15&\textbf{15.54} $\pm$3.52\\%
wrp emb&47.08&2.70 $\pm$0.25&2.88 $\pm$0.54&\underline{0.42} $\pm$0.15&15.45 $\pm$3.61\\%
geo catdepth&47.44&2.77 $\pm$0.31&2.86 $\pm$0.55&\textbf{0.42} $\pm$0.15&\underline{15.54} $\pm$3.57\\%
wrp image&47.72&\underline{2.79} $\pm$0.25&3.00 $\pm$0.51&0.40 $\pm$0.14&14.83 $\pm$3.20\\%
wrp feature&52.65&\textbf{2.84} $\pm$0.27&3.14 $\pm$0.52&0.38 $\pm$0.15&14.06 $\pm$3.28\\%
vqwarper&53.77&2.60 $\pm$0.18&\textbf{\underline{2.72}} $\pm$0.56&0.41 $\pm$0.16&\textbf{\underline{16.60}} $\pm$3.43\\%
infnat s5&76.07&2.44 $\pm$0.21&3.28 $\pm$0.47&0.39 $\pm$0.15&15.24 $\pm$2.87\\%
infnat s1&79.00&2.71 $\pm$0.23&3.11 $\pm$0.58&0.42 $\pm$0.15&15.35 $\pm$3.50\\%
infnat s10&88.81&2.52 $\pm$0.20&3.44 $\pm$0.41&0.35 $\pm$0.14&14.32 $\pm$2.55\\%
MiDaS&106.10&\textbf{\underline{3.62}} $\pm$0.36&3.11 $\pm$0.68&\textbf{\underline{0.45}} $\pm$0.15&14.82 $\pm$2.85\\\bottomrule%
\end{tabular}
\end{footnotesize}
\end{table*}
}

\newcommand{\tabacidbiascomp}{
\begin{table}
\centering
\caption{ACID Bias.}
\label{tab:acidbiascomp}
\begin{footnotesize}
  \begin{tabular}{@{}lrrrrr@{}}%
\toprule%
method&FID ↓&IS ↑&PSIM ↓&SSIM ↑&PSNR ↑\\%
\midrule%
\varvi{}&\textbf{\underline{42.88}}&2.63 \tiny{$\pm$0.14}&2.90 \tiny{$\pm$0.53}&0.40 \tiny{$\pm$0.15}&15.17 \tiny{$\pm$3.40}\\%
\varv{}&\textbf{42.93}&2.62 \tiny{$\pm$0.23}&\textbf{2.86} \tiny{$\pm$0.52}&\underline{0.41} \tiny{$\pm$0.15}&\underline{15.53} \tiny{$\pm$3.34}\\%
\varvii{}&\underline{44.47}&2.67 \tiny{$\pm$0.23}&\textbf{\underline{2.83}} \tiny{$\pm$0.54}&0.41 \tiny{$\pm$0.15}&\textbf{\underline{15.54}} \tiny{$\pm$3.52}\\%
\variii{}&47.08&2.70 \tiny{$\pm$0.25}&2.88 \tiny{$\pm$0.54}&\textbf{0.42} \tiny{$\pm$0.15}&15.45 \tiny{$\pm$3.61}\\%
\variv{}&47.44&\underline{2.77} \tiny{$\pm$0.31}&\underline{2.86} \tiny{$\pm$0.55}&\textbf{\underline{0.42}} \tiny{$\pm$0.15}&\textbf{15.54} \tiny{$\pm$3.57}\\%
\vari{}&47.72&\textbf{2.79} \tiny{$\pm$0.25}&3.00 \tiny{$\pm$0.51}&0.40 \tiny{$\pm$0.14}&14.83 \tiny{$\pm$3.20}\\%
\varii{}&52.65&\textbf{\underline{2.84}} \tiny{$\pm$0.27}&3.14 \tiny{$\pm$0.52}&0.38 \tiny{$\pm$0.15}&14.06 \tiny{$\pm$3.28}\\\bottomrule%
\end{tabular}
\end{footnotesize}
\end{table}
}

\newcommand{\tabacidsotacomp}{
\begin{table}
\centering
\caption{Quantitative comparison on ACID using 32 samples for reconstruction
  metrics. $(n)$: number of steps used for \infnat{}.
  }
\label{tab:acidsotacomp}
\begin{footnotesize}
  \setlength{\tabcolsep}{0.5em}
  \begin{tabular}{@{}lrrrrr@{}}%
\toprule%
method&FID ↓&IS ↑&PSIM ↓&SSIM ↑&PSNR ↑\\%
\midrule%
\varvi{}&\textbf{\underline{42.88}}&2.63 \tiny{$\pm$0.14}&\underline{2.77} \tiny{$\pm$0.54}&\textbf{0.46} \tiny{$\pm$0.14}&\underline{16.49} \tiny{$\pm$3.33}\\%
\varv{}&\textbf{42.93}&2.62 \tiny{$\pm$0.23}&\textbf{2.73} \tiny{$\pm$0.53}&\textbf{\underline{0.46}} \tiny{$\pm$0.14}&\textbf{\underline{16.80}} \tiny{$\pm$3.24}\\%
\midrule%
\vqwarper{}&\underline{53.77}&2.60 \tiny{$\pm$0.18}&\textbf{\underline{2.72}} \tiny{$\pm$0.56}&0.41 \tiny{$\pm$0.16}&\textbf{16.60} \tiny{$\pm$3.43}\\%
    \infnat{}(5)&76.07&2.44 \tiny{$\pm$0.21}&3.28 \tiny{$\pm$0.47}&0.39 \tiny{$\pm$0.15}&15.24 \tiny{$\pm$2.87}\\%
\threedp{}&76.17&\textbf{3.50} \tiny{$\pm$0.47}&3.01 \tiny{$\pm$0.64}&\underline{0.45} \tiny{$\pm$0.14}&14.87 \tiny{$\pm$3.08}\\%
    \infnat{}(1)&79.00&\underline{2.71} \tiny{$\pm$0.23}&3.11 \tiny{$\pm$0.58}&0.42 \tiny{$\pm$0.15}&15.35 \tiny{$\pm$3.50}\\%
    \infnat{}(10)&88.81&2.52 \tiny{$\pm$0.20}&3.44 \tiny{$\pm$0.41}&0.35 \tiny{$\pm$0.14}&14.32 \tiny{$\pm$2.55}\\%
\midas{}&106.10&\textbf{\underline{3.62}} \tiny{$\pm$0.36}&3.11 \tiny{$\pm$0.68}&0.45 \tiny{$\pm$0.15}&14.82 \tiny{$\pm$2.85}\\\bottomrule%
\end{tabular}
\end{footnotesize}
  \vspace{-2.00em}
\end{table}
}

\newcommand{\todo}[1]{\textcolor{red}{#1}}
\newcommand{\threedp}{3DPhoto \cite{DBLP:conf/cvpr/ShihSKH20}}
\newcommand{\synsin}{SynSin \cite{DBLP:conf/cvpr/WilesGS020}}

\twocolumn[{%
\vspace*{-0.5cm}
\maketitle%
\figfirstpagefigure%
}]

%

%
\begin{abstract}
Is a geometric model required to synthesize novel views from a single image?
  Being bound to local convolutions, CNNs need explicit 3D biases to model
  geometric transformations. In contrast, we demonstrate that a
  transformer-based model can synthesize entirely novel views without any
  hand-engineered 3D biases.
  This is achieved by (i) a global attention mechanism for implicitly learning
  long-range 3D correspondences between source and target views, and (ii) a
  probabilistic formulation necessary to capture the ambiguity inherent in
  predicting novel views from a single image, thereby overcoming the
  limitations of previous approaches that are restricted to relatively small
  viewpoint changes.
  We evaluate various ways to integrate 3D priors into a transformer
  architecture.
  However, our experiments show that no such geometric priors are required and
  that the transformer is capable of implicitly learning 3D relationships
  between images.
  Furthermore, this approach outperforms the state of the art in terms of visual quality
  while covering the full distribution of possible realizations.
 \vspace{-1.5em}
\end{abstract}

\newcommand{\tdst}{\text{dst}}
\newcommand{\tsrc}{\text{src}}
\newcommand{\twrp}{\text{wrp}}
\newcommand{\temb}{\text{emb}}
\newcommand{\tcam}{\text{cam}}
\newcommand{\tpos}{\text{pos}}
\newcommand{\tdepth}{\text{d}}
\newcommand{\warp}{F^{\tsrc \rightarrow \tdst}}
\newcommand{\splatting}{\mathcal{S}}
\newcommand{\xdst}{x^{\text{dst}}}
\newcommand{\xsrc}{x^{\text{src}}}
\newcommand{\zdst}{z^{\tdst}}
\newcommand{\zsrc}{z^{\tsrc}}
\newcommand{\emb}{e}

\newcommand{\sdst}{s^{\tdst}}
\newcommand{\ssrc}{s^{\tsrc}}

\newcommand{\RR}{\mathbb{R}}
\newcommand{\EE}{\mathbb{E}}
\newcommand{\encoder}{E}
\newcommand{\decoder}{G}
\newcommand{\codebook}{\mathcal{Z}}

\newcommand{\transformer}{\mathcal{T}}
\newcommand{\cam}{T}

\newcommand{\conv}{W}

\newcommand{\vari}{\emph{expl.-img}}
\newcommand{\varii}{\emph{expl.-feat}}
\newcommand{\variii}{\emph{expl.-emb}}
\newcommand{\variv}{\emph{impl.-catdepth}}
\newcommand{\varv}{\emph{impl.-depth}}
\newcommand{\varvi}{\emph{impl.-nodepth}}
\newcommand{\varvii}{\emph{hybrid}}

\newcommand{\infnat}{InfNat \cite{DBLP:journals/corr/abs-2012-09855}}
\newcommand{\vqwarper}{\emph{expl.-det}}
\newcommand{\midas}{MiDaS \cite{Ranftl2020}}

\section{Introduction}
\label{sec:intro}

\noindent
Imagine looking through an open doorway. Most of the room on the other side is
invisible. Nevertheless, we can estimate how the room \emph{likely} looks. The
few visible features enable an informed guess about the height of the ceiling,
the position of walls and lighting etc. Given this limited information, we can
then imagine several plausible realizations of the room on the other side. This 3D geometric reasoning and the ability to predict what the world will look like \emph{before} we move is critical to orient ourselves in a world with three spatial dimensions. Therefore, we address the problem of novel view synthesis (NVS) \cite{DBLP:conf/siggraph/LevoyH96,
DBLP:conf/siggraph/GortlerGSC96, DBLP:conf/siggraph/DebevecTM96} based on a single initial image and a desired change in viewpoint. In particular, we aim at specifically modeling
\emph{large} camera transformations, \eg rotating the camera by $90\degree$ and
looking at previously unseen scenery. As this is an underdetermined problem, we
present a probabilistic generative model that learns the distribution of
possible target images and synthesizes them at high fidelity. Solving this task
has the potential to transform the passive experience of viewing images into an
interactive, 3D exploration of the depicted scene. This requires an approach 
that both understands the geometry of the scene and, when rendering novel views of an input, %
considers their semantic relationships to the visible content.

\noindent
\textbf{Interpolation vs. Extrapolation}
Recently, impressive synthesis results have been obtained with
geometry-focused approaches in the multi-view setting
\cite{DBLP:conf/eccv/RieglerK20, DBLP:journals/corr/abs-2011-07233,
DBLP:conf/eccv/MildenhallSTBRN20}, where not just a single but a large number
of images or a video of a scene are available such that the task is closer to a
view interpolation than a synthesis of genuinely novel views. 
In contrast, if only a single image is available, the synthesis of novel views is always
an extrapolation task. Solving this task is appealing because
it allows a 3D exploration of a scene starting from only a single
picture.

\noindent 
While existing approaches for single-view synthesis
make small camera transformations, such as a rotation by a few degrees,
possible,
we aim at expanding the possible camera changes to include \emph{large} transformations.
The latter necessitates a probabilistic framework:
Especially when applying large transformation, the problem is underdetermined
because there are many possible target images which are consistent with the
source image and camera pose.  This task cannot be solved with a reconstruction
objective alone, as it will either lead to averaging, and hence blurry
synthesis results, or, when combined with an adversarial objective, cause a
significant mode-dropping when modeling the target distribution.
To remedy these issues, we propose to model this task
with a powerful, autoregressive transformer, trained to
maximize the likelihood of the target data. \\
\textbf{Explicit vs. Implicit Geometry}
The success of transformers is often attributed to the fact that they enforce
less inductive biases compared to convolutional neural networks (CNNs), which
are biased towards local context. Relying mainly on CNNs, this
locality-bias required previous approaches for NVS to explicitly model the overall %
geometric transformation, thereby enforcing yet another inductive bias regarding the
three dimensional structure. In contrast, by modeling interactions between
far-flung
regions of source and target images,
transformers have the potential to learn to represent the required
geometric transformation implicitly without requiring such hand engineered
operations. %
This raises the question whether it is at all necessary to explicitly include
such biases in a transformer model. To address this question, we perform
several experiments with varying degrees of inductive bias and find that our
autoregressively trained transformer model is indeed capable of learning this
transformation completely without built-in priors and can even learn to predict
depth in an unsupervised fashion. \\
\textbf{To summarize our contributions,}
we \emph{(i)} propose to learn a probabilistic model for single view synthesis
that properly takes into account the uncertainties inherent in the task and show
that this leads to significant benefits over previous state-of-the-art
approaches when modeling large camera transformations; see Fig.~\ref{fig:firstpagefigure}.
We \emph{(ii)} also analyze the need for explicit 3D inductive biases in transformer
architectures for the task of NVS with large viewpoint changes 
and find that transformers make it obsolete to explicitly
code 3D transformations into the model and instead can learn the
required transformation implicitly themselves.
We also \emph{(iii)} find that the benefits of providing them geometric information 
in the form of explicit \emph{depth maps} are relatively small, and investigate the ability to recover an
explicit depth representation from the layers of a transformer which
has learned to represent the geometric transformation implicitly and without any 
depth supervision.

\section{Related Work}
\vspace*{-0.35em}
\noindent
\textbf{Novel View Synthesis (NVS)}
We can identify three seminal works which illustrate different levels of reliance
on geometry to synthesize novel views. \cite{DBLP:conf/siggraph/LevoyH96}
describes an approach which requires no geometric model, but requires a large
number of structured input views.
\cite{DBLP:conf/siggraph/GortlerGSC96} describes a similar approach but shows
that unstructured input views suffice if geometric information in the form of
a coarse volumetric estimate is employed.
\cite{DBLP:conf/siggraph/DebevecTM96} can work with a sparse set of views but requires an accurate
photogrammetric model.
Subsequent work also analyzed the commonalities and
trade-offs of these approaches \cite{DBLP:conf/siggraph/BuehlerBMGC01}.
Ideally, an approach could synthesize novel views from a single image without
having to rely on accurate geometric models of the scene and
early works on deep learning for NVS explored
the possibility to directly predict novel views
\cite{DBLP:conf/cvpr/DosovitskiySB15, DBLP:journals/pami/DosovitskiySTB17,
DBLP:conf/nips/KulkarniWKT15, DBLP:conf/nips/YangRYL15,
DBLP:conf/eccv/TatarchenkoDB16} or their appearance flows
\cite{DBLP:conf/eccv/ZhouTSME16, DBLP:conf/cvpr/ParkYYCB17,
DBLP:conf/eccv/SunHLZL18} with convolutional neural networks (CNNs).
However, results of these methods were limited to simple or synthetic data
and subsequent works combined geometric approaches with CNNs.

\noindent
Among these deep learning approaches that explicitly model geometry, we can
distinguish between approaches relying on a proxy geometry to perform a
warping into the target view,
and approaches predicting a 3D representation that can subsequently be rendered
in novel views.
For the proxy geometry, \cite{DBLP:conf/cvpr/MeshryGKHPSM19} relies on point
clouds obtained from
structure from motion (SfM)
\cite{DBLP:conf/iccv/AgarwalSSSS09, DBLP:conf/cvpr/SchonbergerF16} and
multi-view stereo (MVS) \cite{DBLP:conf/eccv/SchonbergerZFP16, DBLP:journals/pami/FurukawaP10}.
To perform the warping, \cite{DBLP:journals/corr/FlynnNPS15, DBLP:journals/tog/XuBSHSR19} use
plane-sweep volumes, \cite{DBLP:journals/tog/KalantariWR16}
estimates depth at novel views and \cite{DBLP:conf/iccv/ChoiGT0K19, DBLP:conf/eccv/XieGF16} a depth
probability volume.
\cite{DBLP:conf/eccv/RieglerK20, DBLP:journals/corr/abs-2011-07233} post-process MVS results to a global mesh and \cite{DBLP:journals/tog/HedmanPPFDB18} relies on
per-view meshes \cite{DBLP:journals/tog/HedmanRDB16}.
Other approaches learn 3D features per scene, which are associated with a point
cloud \cite{DBLP:conf/eccv/AlievSKUL20} or UV maps \cite{DBLP:journals/tog/ThiesZN19}, and decoded to the target image using a CNN.
However, all of these approaches rely on multi-view inputs to obtain an
estimate for the proxy geometry.

\noindent
Approaches which predict 3D representations mainly utilize layered
representations such as layered depth images (LDIs)
\cite{DBLP:conf/siggraph/ShadeGHS98, DBLP:journals/tog/HedmanASK17, DBLP:journals/tog/HedmanK18}, multi-plane images
(MPIs) \cite{DBLP:journals/ijcv/SzeliskiG99, DBLP:journals/tog/ZhouTFFS18,
DBLP:conf/cvpr/SrinivasanTBRNS19, DBLP:conf/cvpr/FlynnBDDFOST19} and variants thereof
\cite{DBLP:journals/tog/PennerZ17, DBLP:conf/eccv/LiXDS20}. While this allows an efficient rendering
of novel views from the obtained representations, their layered nature
limits the range of novel views that can be synthesized with them. Another
emerging approach \cite{DBLP:conf/eccv/MildenhallSTBRN20} represents a five dimensional light field directly with a
multi-layer-perceptron (MLP), but still requires a large number of input views
to correctly learn this MLP.
\figapproachmerged

\noindent
In the case of NVS from a single view, SfM approaches cannot
be used to estimate proxy geometries and early works relied on human
interaction to obtain a scene model \cite{DBLP:conf/siggraph/HorryAA97}.
\cite{DBLP:conf/iccv/SrinivasanWSRN17} uses a large scale, scene-specific light
field dataset to learn CNNs
which predict light fields from a single image. \cite{DBLP:conf/cvpr/LiuHS18}
assumes that scenes can be represented by a fixed set of planar surfaces. To
handle more general scenes, most methods rely on monocular depth estimation
\cite{Ranftl2020, DBLP:conf/eccv/GargKC016, DBLP:conf/cvpr/GodardAB17,
DBLP:conf/cvpr/ZhouBSL17, DBLP:conf/iccv/GodardAFB19} to predict warps
\cite{DBLP:journals/tog/NiklausMYL19, DBLP:conf/cvpr/WilesGS020,
DBLP:journals/corr/abs-2012-09855} or LDIs \cite{DBLP:journals/prl/DhamoTLNT19,
DBLP:journals/tog/KopfMAQGCPFWYZH20, DBLP:conf/cvpr/ShihSKH20}.
\cite{DBLP:conf/cvpr/TuckerS20} directly predicts
an MPI, and \cite{hu2021worldsheet} a mesh.
To handle disocclusions, most of these methods rely on adversarial losses,
inspired by generative adversarial networks (GANs)
\cite{DBLP:conf/nips/GoodfellowPMXWOCB14}, to perform inpainting in
these regions.
However, the quality of these approaches quickly degrades for larger viewpoint
changes
because they do not model the uncertainty of the task. While adversarial losses
can remedy an averaging effect over multiple possible
realizations to some degree, our empirical results show the advantages
of properly modeling the probabilistic nature of NVS from a
single image.

\noindent
\textbf{Self-Attention and Transformers}
The \emph{transformer} \cite{DBLP:conf/nips/VaswaniSPUJGKP17} is a sequence-to-sequence model that models interactions between learned representations of sequence elements by the so-called attention mechanism \cite{DBLP:journals/corr/BahdanauCB14, DBLP:conf/emnlp/ParikhT0U16}. Importantly, this mechanism does not introduce locality biases
such as those present in \eg CNNs, as the importance and interactions of sequence
elements are weighed regardless of their relative positioning. We build our autogressive transformer from the
GPT-2 architecture \cite{radford2019language}, \ie multiple blocks of multihead
self-attention, layer norm \cite{DBLP:journals/corr/BaKH16} and position-wise MLP.

\noindent
\textbf{Generative Two Stage Approaches}
Our approach is based on work in conditional generative modeling combined with neural discrete representation learning (VQVAE) \cite{DBLP:conf/nips/OordVK17}. The latter
aims to learn discrete, compressed representations through either vector quantization
or soft relaxation of the discrete assignment \cite{DBLP:conf/iclr/MaddisonMT17, DBLP:conf/iclr/JangGP17}.
\noindent
This paradigm provides a suitable space~\cite{DBLP:conf/iclr/SalimansK0K17, dieleman2020typicality, DBLP:conf/iclr/0022KSDDSSA17} to train autoregressive (AR) likelihood models on the latent representations and has been utilized to train generative models for hierarchical, class-conditional image synthesis \cite{DBLP:conf/nips/RazaviOV19}, text-controlled image synthesis \cite{DBLP:journals/corr/abs-2102-12092} and music generation \cite{DBLP:journals/corr/abs-2005-00341}, 
and continuous analogues using VAEs \cite{DBLP:conf/iclr/DaiW19} or normalizing flows \cite{DBLP:conf/cvpr/EsserRO20} exist.
Recently, \cite{DBLP:journals/corr/abs-2012-09841} demonstrated that adversarial training of the VQVAE improves
compression while retaining high-fidelity reconstructions, subsequently enabling efficient training of an
AR transformer model on the learned latent space (yielding a so-called VQGAN). 
We directly build on this work and use VQGANs to represent both source and target views and, when needed, 
depth maps. 
Concurrent to our work, \cite{rockwell2021pixelsynth} develop an approach to NVS which uses a VQVAE and PixelCNN++ \cite{DBLP:conf/iclr/SalimansK0K17} to outpaint large viewpoint changes. 

\section{Approach}

\noindent
To render a given image $\xsrc$ experienceable in a 3D manner, we allow the
specification of arbitrary new viewpoints, including in particular \emph{large}
camera transformations $\cam$. As a result we expect multiple plausible
realizations $\xdst$ for the novel view, which are all consistent with the
input, since this problem is highly underdetermined. Consequently, we follow a
probabilistic approach and sample novel views from the distribution
\begin{equation}
\xdst \sim p(\xdst \vert \xsrc, \cam).
\label{eq:dstdistr}
\end{equation}
To solve this task, a model must explicitly or implicitly learn the 3D relationship between both images and $\cam$.
In contrast to most previous work that tries to solve this task with CNNs and
therefore oftentimes includes an explicit 3D transformation, we
want to use the expressive transformer architecture and investigate to what
extent the explicit specification of such a 3D model \emph{is necessary at
all}. 

\noindent
Sec.~\ref{subsec:transformertraining} describes how to train
a transformer model in the latent space of a
VQGAN.
Next, Sec.~\ref{subsec:encodingbiases}
shows how inductive biases
can be build
into the transformer and describes all bias-variants that we
analyze.
Finally, Sec.~\ref{subsec:depthreadout} presents our
approach %
to extract geometric information from a transformer
where no
3D bias has been explicitly specified.
\subsection{Probabilistic View Synthesis in Latent Space}
\label{subsec:transformertraining}
\noindent 
Learning the distribution in Eq.~\eqref{eq:dstdistr}
requires a model
which can capture long-range interactions between source and target view to implicitly represent geometric
transformations.
Transformer architectures naturally meet these requirements, since they are not confined to short-range relations such
as CNNs with their convolutional kernels and exhibit 
state-of-the-art performance
\cite{DBLP:conf/nips/VaswaniSPUJGKP17}.
Since likelihood-based models have been shown
\cite{DBLP:conf/iclr/SalimansK0K17} to spend too much capacity on short-range
interactions of pixels when modeling images directly in pixel space, we follow \cite{DBLP:journals/corr/abs-2012-09841} and employ a two-stage
training. The first stage performs adversarially guided
discrete representation learning (VQGAN), obtaining an abstract
latent space %
that has proved to be well-suited for efficiently training generative transformers
\cite{DBLP:journals/corr/abs-2012-09841}. %
\newcommand{\tokenemb}{g}
\noindent
\textbf{Modeling Conditional Image Likelihoods}
VQGAN consists of an encoder $\encoder$, a decoder $\decoder$ and a
codebook $\codebook = \{z_i\}_{i=1}^{\vert \codebook \vert}$ of discrete
representations $z_i \in \RR^{d_z}$. 
The trained VQGAN allows to encode any $x \in \RR^{H \times W \times 3}$
into the discrete latent space as $\encoder(x) \in \RR^{h\times w
\times d_z}$\footnote{This includes the vector quantization step as described
in \cite{DBLP:conf/nips/OordVK17}}.  Unrolled in raster-scan order, this latent
representation corresponds to a sequence $s \in \RR^{h\cdot w \times d_z}$ and
can be equivalently expressed as a sequence of integers which index the learned
codebook $\codebook$. Following the usual designation
\cite{DBLP:conf/nips/VaswaniSPUJGKP17} we refer to the sequence elements as
``tokens''. An embedding function $\tokenemb=\tokenemb(s)\in\RR^{h\cdot w
\times d_e}$ maps each of these tokens into the embedding space of the
transformer $\transformer$ and adds learnable positional encodings.
Similarly, to encode the input view $\xsrc$ and the camera transformation
$\cam$, both are mapped into the embedding space by a function $f$:

\begin{equation}
  f: (\xsrc, \cam) \mapsto f(\xsrc, \cam) \in \RR^{n \times d_e}, %
\end{equation}
where $n$ denotes the length of the conditioning sequence.
By using different functions $f$ various inductive biases can be incorporated into the architecture as described in Sec.~\ref{subsec:encodingbiases}.
The transformer $\transformer$ then processes the concatenated sequence 
$[f(\xsrc, T), \tokenemb(\sdst)]$  to learn the distribution of plausible novel views conditioned on $\xsrc$ and $\cam$,  
\begin{equation}
  p_{\transformer}\Big(\sdst | f(\xsrc, \cam) \Big) = \prod_i p_{\transformer}\Big(\sdst_i \vert \sdst_{<i}, f(\xsrc, \cam)\Big).
\end{equation}
\noindent Hence, to train an autoregressive transformer by next-token prediction 
$p_{\transformer}(s_i \vert s_{<i},f(\xsrc, \cam))$
we maximize the log-likelihood of the data, leading to the training objective %
\begin{equation}
\mathcal{L}_{\transformer} = \EE_{\xsrc, \xdst \sim p(\xsrc, \xdst)}
\Big[-\log p_{\transformer}\big(\sdst \vert f(\xsrc, \cam)\big) \Big].
\end{equation}
\vspace{-1.75em}
\subsection{Encoding Inductive Biases}
\vspace{-0.35em}
\label{subsec:encodingbiases}
\noindent 
Besides achieving high-quality NVS, we aim
to investigate to what extent transformers depend on
a 3D inductive bias. To this end, we compare approaches where a geometric
transformation is built \emph{explicitly} into the
conditioning function $f$, and approaches where no such
transformation is used. In the latter case, the transformer
itself must learn the required relationship between source and target view. If
successful, the transformation will be described
\emph{implicitly} by the transformer.

\noindent 
\textbf{Geometric Image Warping}
We first describe how an explicit geometric transformation results from the 3D
relation of source and target images.
For this, pixels of the source image are back-projected to three
dimensional coordinates, which can then be re-projected into the target view.
We assume a pinhole
camera model, such that the projection of 3D points
to homoegenous pixel coordinates is determined
through the intrinsic camera matrix $K$.
The transformation %
between
source and target coordinates is given by a
rigid motion, consisting of a rotation $R$ and a translation $t$.
Together, these parameters specify the desired control over the novel view to
be generated, \ie $T=(K,R,t)$.
\tabmergedbiascomp

To project pixels back to 3D coordinates, we require information
about their depth $d$, since this information has been discarded by their
projection onto the camera plane.
Since we assume access to only a single source view, we require a monocular
depth estimate. 
Following by previous works \cite{DBLP:conf/cvpr/ShihSKH20,
DBLP:journals/corr/abs-2012-09855}, we use MiDaS \cite{Ranftl2020} in all of
our experiments which require monocular depth information.

The transformation can now be described as a mapping of
pixels $i \in \{1,\dots,H\}, j \in \{1,\dots,W\}$ in the source image
$\xsrc \in \RR^{H\times W\times 3}$ to pixels $i', j'$ in the target image.
In homogeneous coordinates, their relationship is given by
\begin{equation}
  \label{eq:flow}
  \begin{pmatrix} j' \\ i' \\ 1 \end{pmatrix} \simeq K\left( R K^{-1} d(i,j)
    \begin{pmatrix} j \\ i \\ 1 \end{pmatrix} + t\right)
\end{equation}
This relationship
defines a forward flow field
$\warp=\warp(K, R, t, d) \in \RR^{H\times W\times 3}$ from source to target as
a function of depth and camera parameters.
The flow field can then be used to warp the source image $\xsrc$ into the
target view with a warping operation $\splatting$:
\begin{equation}
x^\twrp=\splatting(\warp, x^\tsrc).
\end{equation}
Because the target pixels obtained from the flow are not necessarily integer
valued, we follow \cite{DBLP:conf/cvpr/Niklaus020} and implement
$\splatting$ by bilinearly splatting features across the four
closest target pixels.  When multiple source pixels
map to the same target pixels, we use
their relative depth
to
give points closer to the camera more weight---a soft
variant of z-buffering.

In the simplest case, we can now describe the difference between explicit and
implicit approaches in the way that they receive information about the source
image and the desired target view.
Here, explicit approaches receive
source information warped
using the camera parameters,
whereas implicit approaches receive the original source image and the
camera parameters themselves, \ie
\begin{align}
  &\text{\emph{explicit:}} & \splatting(\warp(K, R, t, d), \xsrc) \\
  &\text{\emph{implicit:}} & (K, R, t, d, \xsrc)
\end{align}
Thus, in explicit approaches we enforce an inductive bias on the 3D
relationship between source and target by making this relationship explicit,
while implicit approaches have to learn it on their own.
Next, we introduce a number of different variants for each, which are
summarized in Fig.~\ref{fig:approachmerged}.

\noindent
\textbf{Explicit Geometric Transformations}
\newcommand{\learnemb}{e}
\newcommand{\learnpos}{e^\tpos}
In the following, we describe all considered variants in terms of the transformer's conditioning function $f$.
Furthermore, $\learnemb$ denotes a learnable embedding mapping the discrete
VQGAN codes $\encoder(x)$ into the embedding space of the
transformer. Similarly, $\learnpos \in \RR^{n\times d_e}$ denotes a learnable positional encoding. 
The flow field $\warp(K, R, t, d)$ is always computed from $\xsrc$
and, to improve readability, we omit it from the arguments of the warping
operation, \ie $\splatting(\cdot)=\splatting(\warp(K, R, t, d), \cdot)$.

\emph{(I)} Our first explicit variant, \vari{}, warps the source image and
encodes it
in the same way as the target image:
\begin{equation}
  \label{eq:fvari}
  f(\xsrc, \cam) = \learnemb(\encoder(\splatting(\xsrc)))+\learnpos
\end{equation}

\emph{(II)} Inspired by previous works \cite{DBLP:conf/eccv/RieglerK20,DBLP:conf/eccv/AlievSKUL20}
we include
a \varii{} variant which first encodes the original source image, and subsequently
applies the warping on top of these features. We again use the VQGAN encoder $\encoder$ to
obtain
\begin{equation}
  \label{eq:fvarii}
  f(\xsrc, \cam) = \learnemb(\splatting(\encoder(\xsrc)))+\learnpos
\end{equation}

\emph{(III)} To account for the fact that the warped features in Eq.~\eqref{eq:fvarii}
remain fixed (due to $\encoder$ being frozen), we also consider 
a \variii{} variant that warps the \emph{learnable} embeddings 
and positional encodings of the transformer model.
More precisely, we concatenate original embeddings with their warped variants
and merge them with a learnable matrix.
Doing this for both the embeddings of the codes and for the
positional encodings using matrices $W^\temb, W^\tpos \in \RR^{d_e \times 2\cdot d_e}$, 
the conditioning function $f$ then reads:
\begin{align}
  \label{eq:fvariii}
  f(\xsrc, \cam) = &W^\temb[
    \learnemb(\encoder(\xsrc)),
    \splatting(\learnemb(\encoder(\xsrc)))
  ] +\nonumber\\
  &W^\tpos[
    \learnpos, \splatting(\learnpos)
  ]
\end{align}

\noindent
\textbf{Implicit Geometric Transformations}
Next, we describe implicit variants that we use to analyze if
transformers---with their ability to attend to all positions equally
well---require an explicit geometric transformation built into the model.
We use the same notation as for the explicit variants.

\emph{(IV)} The
first variant, \variv{}, provides the transformer with all the same components
which are used in the explicit variants: Camera parameters $K, R, t$, estimated
depth $d$ and source image $\xsrc$. Camera parameters are flattened and
concatenated to $\hat{\cam}$, which is mapped
via
$\conv^\tcam \in \RR^{d_e \times 1}$
to the embedding space.
Depth and source images are encoded by VQGAN encoders $\encoder^\tdepth$ and
$\encoder$ to obtain
\begin{equation}
  \label{eq:fvariv}
  f(\xsrc, \cam) = [\conv^\tcam \hat{\cam},
  \learnemb(\encoder^\tdepth(d)),
  \learnemb(\encoder(\xsrc))] + \learnpos
\end{equation}
Compared to the other variants, this sequence
is roughly $\frac{3}{2}$ times longer, resulting in twice the
computational costs.
\figreconstructionbothdepth

\emph{(V)} Therefore, we also include a \varv{} variant, which concatenates the discrete codes of
depth and source image, and maps them with a matrix $W \in \RR^{d_e \times
2\cdot d_z}$ to the embedding space to avoid an increase in sequence length:
\begin{equation}
  \label{eq:fvarv}
  f(\xsrc, \cam) = [\conv^\tcam \hat{\cam},
  W[\encoder^\tdepth(d),
  \encoder(\xsrc)]] +\learnpos
\end{equation}

\emph{(VI)} Implicit approaches offer an intriguing possibility: Because they do
not need an explicit estimate of the depth to perform the warping operation
$\splatting$, they hold the potential to solve the task without such a depth
estimate.
Thus, \varvi{} uses
only camera parameters and source image---the bare minimum according to
our task description.
\begin{equation}
  \label{eq:fvarvi}
  f(\xsrc, \cam) = [\conv^\tcam \hat{\cam},
  \learnemb(\encoder(\xsrc))] + \learnpos
\end{equation}

\emph{(VII)} Finally, we analyze if explicit and implicit approaches offer
complementary strengths.
Thus, we add a \varvii{}
 variant whose
conditioning function is %
the sum of the $f$'s of
\variii{} in Eq.~\eqref{eq:fvariii} and \varv{} in Eq.~\eqref{eq:fvarv}.

\subsection{Depth Readout}
\vspace{-0.35em}
\label{subsec:depthreadout}
\noindent
To investigate the ability to learn an implicit model of the geometric
relationship between different views, we propose to extract an explicit estimate of
depth from a trained model. To do so, 
we use linear probing \cite{DBLP:conf/icml/ChenRC0JLS20}, which is commonly used to investigate
the feature quality of unsupervised approaches. More specifically, we assume a
transformer model consisting of $L$ layers and of type \varvi{},
which is conditioned on source frame and transformation parameters only. Next, we specify a certain layer 
$0 \leq l \leq L$ (where $l=0$ denotes the input) and extract its latent representation $e^l$, 
corresponding to the positions of the provided source frame $\xsrc$.
We then train a position-wise linear classifier $W$ to predict 
the discrete, latent representation
of the depth-encoder $\encoder^\tdepth$ (see Sec.~\ref{subsec:encodingbiases}) 
via a cross-entropy objective from $e^l$.
Note that both the weights of the transformer and the VQGANs remain fixed.

\section{Experiments}
\vspace{-0.35em}
\label{sec:experiments}
\noindent 
\enlargethispage{\baselineskip}
First, Sec.~\ref{subsec:implicitvsexplicit} integrates the different explicit and implicit inductive biases into the transformer to judge if such geometric biases are needed at all. 
Following up, %
Sec.~\ref{subsec:comparesota} compares implicit variants to previous work and 
evaluates both the visual quality and fidelity of synthesized novel views.
Finally, we evaluate the ability of the least biased variant, \varvi{},
to implicitly represent scene geometry,  %
observing that they indeed capture 
such 3D information.

\vspace*{-0.35em}
\subsection{Comparing Implicit and Explicit Transformers}
\vspace{-0.35em}
\label{subsec:implicitvsexplicit}
\noindent 
To investigates if transformers need (or benefit from) an explicit warping
between source and target view we first compare how well the different variants
from Sec.~\ref{subsec:encodingbiases} (see also Fig.~\ref{fig:approachmerged})
can learn a probabilistic model for NVS.
We then
evaluate both the quality and fidelity of their samples.%

To prepare, we first train VQGANs on frames of the RealEstate10K
\cite{DBLP:journals/tog/ZhouTFFS18} and ACID
\cite{DBLP:journals/corr/abs-2012-09855} datasets, whose preparation is
described in the supplementary.
We then train the various transformer variants %
on the latent space of the respective first stage models. Note that this procedure 
ensures comparability of different settings within a given dataset, as the space in which the likelihood is measured remains fixed.

\noindent
\textbf{Comparing Density Estimation Quality}
A basic measure for the performance of probabilistic models is the likelihood
assigned to validation data. Hence, we begin our evaluation 
of the different variants by comparing their (minimal) negative log-likelihood (NLL)
on RealEstate and ACID. %
\noindent
Based on the results in Tab.~\ref{tab:mergedbiascomp}, we can identify three groups with
significant performance differences on ACID: The implicit variants \variv{}, \varv{}, and
\varvi{} and \varvii{} achieve the best performance,
which indicates an advantage over
the purely explicit variants. Adding an explicit warping as in the \varvii{}
model does not help significantly.
\noindent 
Moreover, %
\varii{} is unfavorable, possibly due to
the features $\encoder(\xsrc)$
remaining fixed while training the transformer.
The \emph{learnable} features which are warped in variant \variii{} obtain a lower NLL and thereby confirm the former hypothesis. Still there are no improvements of warped features over warped pixels as in variant \vari{}.
\figentropy
\figrealestate
\noindent 
The results on RealEstate look similar
but
in this case the implicit
variant without depth, \varvi{}, performs a bit worse than \vari{}. %
Presumably,
accurate depth information obtained from a supervised, monocular depth
estimation model are much more beneficial in the indoor setting of
RealEstate compared to the outdoor setting of ACID.

\enlargethispage{\baselineskip}
\noindent
\textbf{Visualizing Entropy of Predictions}
The NLL %
measures
the ability of the transformer to predict target views. 
The entropy of the predicted distribution over the codebook entries for each position captures the prediction uncertainty of the model. See Fig.~\ref{fig:entropy} for a visualization
of variant \varvi{}.
The model is more confident in its predictions for regions which are visible in the source image. This indicates that it is indeed
able to relate source and target via their geometry instead of simply predicting an arbitrary novel view.

\noindent
\textbf{Measuring Image Quality and Fidelity}
Since NLL does not necessarily reflect the visual quality of the images
\cite{DBLP:journals/corr/TheisOB15}, we evaluate the latter also directly. Comparing predictions
with ground-truth helps to judge how well the model respects the geometry.
However, for large camera movements, large parts of the target image
are not visible
in the source view. Thus, we must also evaluate the quality of the
content imagined by the model, which 
might be fairly different from
that of
the
ground-truth, since the latter is just one of many possible realizations of
the real-world.

\noindent
To evaluate the image quality without a direct comparison to the ground-truth,
we report FID scores \cite{DBLP:conf/nips/HeuselRUNH17}. To evaluate the fidelity to the ground-truth, we report
the low-level similarity metrics SSIM \cite{DBLP:journals/tip/WangBSS04} and PSNR, and the high-level similarity
metric PSIM \cite{DBLP:conf/cvpr/ZhangIESW18}, which better represents human assessments of
visual similarity.
\noindent
Tab.~\ref{tab:mergedbiascomp} contains the results for RealEstate10K and ACID. 
In general, they reflect the findings from the NLL
values: Image quality and fidelity of implicit variants with access to depth are superior to explicit variants. 
The implicit variant without depth (\varvi{}) consistently achieves the same good FID
scores as the implicit variants with depth (\variv{} \& \varv{}), 
but cannot achieve quite the same level
of performance in terms of reconstruction fidelity. However, it is on par with the
explicit variants, albeit requiring no depth supervision.
\vspace{-0.75em}

\subsection{Comparison to Previous Approaches}
\vspace{-0.35em}
\label{subsec:comparesota}
\tabrealsotacomp
\enlargethispage{1\baselineskip}
\noindent
Next, we compare our best performing variants \varv{} and \varvi{} to previous
approaches for NVS: \threedp{}, \synsin{} and
\infnat{}. \threedp{} has been trained on MSCOCO \cite{DBLP:conf/eccv/LinMBHPRDZ14} to work on arbitrary
scenes, whereas \synsin{} and \infnat{} have been trained 
on
RealEstate and ACID, respectively. 

\noindent
To assess the effect of formulating the problem probabilistically, we introduce another baseline to compare probabilistic and deterministic models
with otherwise equal architectures. Specifically, we use
the same VQGAN architecture as described in
Sec.~\ref{subsec:transformertraining}. However, it is not trained as an
autoencoder, but instead the encoder receives the warped source image
$x^\twrp$, and the decoder predicts the target image $x^\tdst$. 
This model, denote by \vqwarper{}, represents an \emph{explicit and
deterministic} baseline. Finally, we include the warped source image
itself as a baseline denoted by \midas{}.\\
\figacid
\noindent
Utilizing the probabilistic nature of our model, 
we analyze how
close we can get to a particular target image with a fixed amount of samples.
Tab.~\ref{tab:realsotacomp} and \ref{tab:acidsotacomp} report the
reconstruction metrics with 32 samples per target.
\noindent
The probabilistic variants consistently achieve the best values for the similarity metrics
PSIM, SSIM and PSNR on RealEstate, and are always among the best three
on ACID, where \vqwarper{} achieves the best PSIM values and the second
best PSNR values.
We show the reconstruction metrics on
RealEstate as a function of the number of samples in
Fig.~\ref{fig:reconstructionbothdepth}. With just four samples, the performance of \varv{} is better than all
other approaches except for the SSIM values of \threedp{},
which are overtaken
by \varv{}
with 16 samples, and do not saturate with 32 samples,
which demonstrates the advantages of a probabilistic formulation of
NVS. \\
\tabacidsotacomp
\enlargethispage{1\baselineskip}
\noindent
These results should be considered along with the competitive FID scores in Tab.~\ref{tab:realsotacomp} 
and \ref{tab:acidsotacomp} (where the implicit variants always constitute the best and second best value)
and the qualitative results in
Fig.~\ref{fig:realestate} and \ref{fig:acid}, 
underlining the high quality of our synthesized views.
It is striking that IS assigns the best scores to
\threedp{} and \midas{}, which contain large and plain regions of
gray color in regions where the source image does not provide information about
the content. Where the monocular depth estimation is accurate,
\threedp{} shows good results but it can only inpaint small areas. \synsin{} and
\infnat{} can fill larger areas but, for large camera motions, their results
become blurry and a similar observation holds for \vqwarper{}.
The probabilistic
variants \varv{} and \varvi{} consistently produce plausible results which are
largely consistent with the source image, although small details sometimes
differ. This shows that only the
probabilistic variants are able 
to synthesize high quality images for large camera changes.

\subsection{Probing for Geometry}
\vspace{-0.5em}
\label{subsec:extractthedepth}
\noindent 
Based on the experiments in Sec.~\ref{subsec:implicitvsexplicit} and
Sec.~\ref{subsec:comparesota}, which showed that the unbiased variant \varvi{}
is mostly on-par with the others, we investigate the question whether this
model is able to develop an implicit 3D ``understanding'' without explicit 3D
supervision.  To do so, we perform linear probing experiments
as described in Sec.~\ref{subsec:depthreadout}.
\enlargethispage{1\baselineskip}

\noindent
Fig.~\ref{fig:depthprobinglayers} plots the negative cross-entropy loss
and the negative PSIM reconstruction error of the recovered depth maps against
the layer depth of the transformer model.
Both metrics are consistent and quickly increase when probing
deeper representations of the transformer model.
Furthermore, both curves
exhibit a peak for $l=4$ (\ie after the third self-attention block) 
and then slowly decrease with increasing layer depth. The depth maps obtained from 
this linear map resemble the corresponding true depth maps qualitatively well
as shown in
Fig.~\ref{fig:depthprobinglayers}.
This figure demonstrates that a linear estimate of depth only becomes possible
through the representation learned by the transformer ($l=4$) but not by the
representation of the VQGAN encoder ($l=0$).
We hypothesize that, in order to map an input view onto a target view, the
transformer indeed develops an implicit 3D representation of the scene to solve
its training task. \vspace{-0.65em}

\label{subsec:probingdepth}
\figdepthprobinglayers
\figdepthprobesqualitative

\section{Discussion}
\vspace{-0.35em}
\noindent
We have introduced a probabilistic approach based on transformers for novel
view synthesis from a single source image with strong changes in viewpoint.
Comparing various explicit and implicit 3D inductive biases for the transformer
showed that explicitly using a 3D transformation in the architecture
does not help their performance significantly.
However, removing inductive biases also comes at a price.

\noindent
Without priors on camera movements or warping layers, the architecture must be
able to take relationships between arbitrary positions into account, which
requires a compressed representation. In our
experiments, compression artifacts dominate the error for small viewpoint
changes. Avoiding them increases computational costs (see Sec.~\ref{sec:compressionartifacts}).
\enlargethispage{1\baselineskip}
\blfootnote{\noindent This work has been supported by the German Research Foundation
(DFG) projects 371923335 and 421703927.}
\noindent
Synthesizing two views from the same image generally results in
two incompatible realizations. However, we can run our approach iteratively.
When synthesizing continuous trajectories, sampling still leads to flickering
but this can be alleviated with deterministic sampling (see Sec.~\ref{sec:addresults}).

\noindent
To conclude, our approach is not a final solution to novel view synthesis, but an
important step towards synthesizing large camera changes
and understanding the need for 3D priors. %
Our results demonstrate significant improvements over existing approaches, and
even with no depth information as input
our model learns to infer depth within its internal representations.
Future works should explore how to combine these capabilities and insights
with improved performance at synthesizing stable high-resolution
trajectories.

\newpage
\appendix
\onecolumn

\begin{center}
  \textbf{
    \Large Geometry-Free View Synthesis \\ \large \textsl{Transformers and no 3D Priors} \vspace{0.3em} } \\
  \Large 
-- \\
 \textbf{\large Supplementary Material} \\
\hspace{1cm}
\end{center}

\FloatBarrier
\noindent
In this supplementary, we provide additional results obtained with our models
in Sec.~\ref{sec:addresults}. Sec.~\ref{sec:architectures} summarizes models,
architectures and hyperparameters that were used in the main paper. After
describing details on the training and test data in
Sec.~\ref{sec:trainingdata} and on the uncertainty evaluations via the
entropy in Sec.~\ref{sec:entropydetails}, Sec.~\ref{sec:compressionartifacts} concludes the
supplementary material with a brief discussion of the compression artifacts introduced 
by the usage of the VQGAN as the compression model.

\section{Additional Results}
\label{sec:addresults}
\paragraph{Interactive Scene Exploration Interface}
Fig.~\ref{fig:videopreview} shows a preview of the videos available at
\url{https://git.io/JOnwn}, which demonstrate an interface for interactive 3D
exploration of images. Starting from a single image, a user can use keyboard
and mouse to move the camera freely in 3D. To provide orientation, we warp the
starting image to the current view using a monocular depth estimate
(corresponding to the \midas{} baseline in Sec.~\ref{subsec:comparesota}). This
enables a positioning of the camera with real-time preview of the novel view.
Once a desired camera position has been reached, the spacebar can be pressed to
autoregressively sample a novel view with our transformer model.

In the videos available at \url{https://git.io/JOnwn}, we use camera
trajectories from the test sets of RealEstate10K and ACID, respectively. The
samples are produced by our \varv{} model, and for an additional visual
comparison, we also include results obtained with the same methods that we
compared to in Sec.~\ref{subsec:comparesota}.

\figvideopreview

\paragraph{Small Viewpoint Changes \& Continuous Trajectories}
\noindent For very small viewpoint changes, distortions due to compression
dominate the error of our approach (left of Fig.~\ref{fig:compare}, where the x-axis
uses PSIM between source and target view as a proxy for the difficulty/magnitude 
of viewpoint change). Still, our approach outperforms
previous approaches when considering the average over small, medium and large
viewpoint changes (solid lines at the left) and the gap quickly increases when
considering more difficult examples (solid lines, right).
See also Sec.~\ref{sec:compressionartifacts} on the trade-off between distortion caused by
compression and computational efficiency.
Our approach can also be applied to small viewpoint changes and the generation of continuous, 
consistent trajectories; see Fig.~\ref{fig:sequence}.
Note that the ability of previous approaches to synthesize small
viewpoint changes well does not enable high-quality synthesis of even moderately
long trajectories.
\compare
\sequence

\paragraph{Transformer Variants Over the Course of Training}
\noindent
Fig.~\ref{fig:nllrealandacid} reports the negative log-likelihood (NLL) over
the course of training on RealEstate and ACID, respectively. The models overfit
to the training split of ACID early which makes training on ACID much quicker
and thus allows us to perform multiple training runs of each variant with
different initializations. This enables an estimate of the significance of the
results by computing the mean and standard deviation over three runs
(solid line and shaded area in
Fig.~\ref{fig:nllrealandacid}, respectively).

\paragraph{Additional Qualitative Results}
For convenience, we also include additional qualitative results directly in
this supplementary. Fig.~\ref{fig:supprealestate} and \ref{fig:suppacid} show
additional qualitative comparisons on RealEstate10K and ACID, as in
Fig.~\ref{fig:realestate} and \ref{fig:acid} of the main paper.
Fig.~\ref{fig:supprealestatesamples} and \ref{fig:suppacidsamples} demonstrate
the diversity and consistency of samples by showing them along with their
pixel-wise standard deviation. Fig.~\ref{fig:figdepthprobesqualitativesupp}
contains results from the depth-probing experiment of
Sec.~\ref{subsec:extractthedepth}, and Fig.~\ref{fig:entropysupp} from the
entropy visualization of Sec.~\ref{subsec:implicitvsexplicit}.

\newpage 
\fignllrealandacid

\section{Architectures \& Hyperparameters}
\label{sec:architectures}
\paragraph{Transformer}
The architecture of all transformer models discussed in this work follows the 
GPT-2 architecture \cite{radford2019language}. More specifically, 
the transformer consists of $L$ transformer blocks, 
where each block performs the following operation
on an input sequence $z \in \RR^{\vert z \vert \times d_e}$ (with $\vert z \vert$ the length of $z$): 
\begin{align}
z_1 &= \text{LayerNorm}(z) \\
z_2 &= \text{MultiHeadSelfAttention}(z_1) + z \\
z_3 &= \text{LayerNorm}(z_2)  \\
z_4 &= \text{MLP}(z_3) + z_2
\end{align}
In contrast to the global attention operation, the MLP is applied position-wise.

Given an input sequence $s$ and an embedding produced by the conditioning function $f(\xsrc, \cam) \in \RR^{n \times d_e}$ (see Sec.~\ref{subsec:encodingbiases}), the transformer maps $s$ to a learnable embedding 
$e(s) + e^{\text{pos}} =: \hat{e}^0 \in \RR^{\vert s \vert \times d_e}$, applies the $L$ transformer blocks on the concatenated 
sequence $[f(\xsrc, \cam), \hat{e}^0]$ and 
finally projects to $\vert \codebook \vert $ logits $\pi^L$
via a linear transformation $W_{\text{head}}$, which
correspond to a categorical distribution over sequence 
elements, \ie
\begin{align}
e^0 &= [f(\xsrc, \cam), e(s) + e^{\text{pos}}]  \\
e^l &= \text{TransformerBlock}(e^{l-1}), \quad l=1\dots L \\
\pi^L &= W_{\text{head}} \cdot \text{LayerNorm}(e^L).
\end{align}
Note that non-conditioning elements, \ie the last $\vert s \vert$ elements, are masked autoregressively \cite{DBLP:conf/nips/VaswaniSPUJGKP17}.
For all experiments, we use an embedding dimensionality $d_e = 1024$, $L=32$ transformer blocks, 16 attention heads, two-layer MLPs with 
hidden dimensionalities of $4\cdot d_e$ and a codebook of size $\vert \codebook \vert = 16384$. This setting results in
a transformer with $437 M$ parameters.
We train the model using the AdamW \cite{DBLP:conf/iclr/LoshchilovH19} optimizer (with $\beta_1=0.9$, $\beta_2=0.95$) and apply weight decay of 0.01 on 
non-embedding parameters. 
We train for $500k$ steps, where we first linearly increase the learning rate
from $2.5\cdot 10^{-6}$ to $1.5\cdot 10^{-4}$ during the first $5k$ steps, 
and then apply a cosine-decay learning rate schedule \cite{DBLP:conf/iclr/LoshchilovH17} towards zero.
\paragraph{VQGAN}
The architecture and training procedure of the VQGANs is adopted from \cite{DBLP:journals/corr/abs-2012-09841}, where we 
use a downsampling factor of $f=2^4$. For the codebook $\codebook$, we use $\vert \codebook \vert = 16384$ entries and 
a dimensionality of $d_z = 256$. This means that any input $x \in \RR^{H \times W \times C}$ will be mapped to a latent 
representation of size $\encoder(x) \in \RR^{H/2^4 \times W/2^4 \times 256}$. For our experiments on RealEstate and ACID, 
where $H=208$ and $W=368$, this corresponds to a latent code of size $13 \times 23$ (which is then unrolled to a sequence 
of length $\vert s \vert = 299$).
We use the authors' official implementation and pretrained models\footnote{see \url{https://github.com/CompVis/taming-transformers}} and perform finetuning on frames of RealEstate and ACID for 50'000 steps on either dataset, 
resulting in two dataset-specific VQGANs.

\paragraph{Other models} For monocular depth estimates, we use MiDaS
v2.1\footnote{see \url{https://github.com/intel-isl/MiDaS}}. We use the
official implementations and pretrained models for the comparison with
\threedp{}\footnote{see
\url{https://github.com/vt-vl-lab/3d-photo-inpainting}}, \synsin{}\footnote{see
\url{https://github.com/facebookresearch/synsin/}} and \infnat{}\footnote{see
\url{https://github.com/google-research/google-research/tree/master/infinite_nature}}.

\section{Training and Testing Data}
\label{sec:trainingdata}
\noindent 
Training our conditional generative model requires examples consisting
of $(\xdst, \xsrc, \cam)$. Such training pairs can be obtained via SfM
\cite{DBLP:journals/tog/ZhouTFFS18} applied
to image sequences, which provides poses $(R_i, t_i)$ for each frame $x^i$ with respect
to an arbitrary world coordinate system. For two frames $\xsrc, \xdst$ from the
sequence, the relative transformation
is then given by
$R=R_{\tdst}R_{\tsrc}^{-1}$ and $t=t_{\tdst}-R t_{\tsrc}$.
However, the scale of the
camera translations obtained by SfM is also arbitrary, and without access to
the full sequence, underspecified.

\noindent
To train the model and to meaningfully compute reconstruction errors for the
evaluation, we must resolve this ambiguity. To do this, we also triangulate a
sparse set of points for each sequence using COLMAP \cite{DBLP:conf/cvpr/SchonbergerF16}.
We then compute a
monocular depth estimate for each image using MiDaS \cite{Ranftl2020} and compute the
optimal affine scaling to align this depth estimate with the scale of the
camera pose. Finally, we normalize depth and camera translation by the minimum depth estimate.

\noindent All qualitative and quantitative results are obtained on a subset of the test
splits of RealEstate10K \cite{DBLP:journals/tog/ZhouTFFS18} and ACID
\cite{DBLP:journals/corr/abs-2012-09855}, consisting of 564 source-target
pairs, which have been selected to contain medium-forward, large-forward,
medium-backward and large-backward camera motions in equal parts. 
We will make this split publicly available along with our code.

\subsection*{A Note on Reported Metrics}
\noindent Since our stated goal is to model \emph{large} camera transformations,
our evaluation focuses on this ability and thus differs from the evaluation of
SynSin, which is biased to small changes; see Tab.~\ref{tab:thenumbers}:
With the SynSin evaluation (small cam-$\Delta$) we reproduce the officially reported numbers
and our choice to evaluate at the original aspect ratio (at $208\times 368$)
has minor effects.
The last three rows show the deterioration of metrics if we remove biases to small changes $\Delta$:
We (i) evaluate all test pairs,
not just the better of two views (w/o best of 2), (ii) remove 
$10.14\%$ of test pairs
that contain no camera change at all (w/o src $\equiv$ tgt) and (iii) add 
$50\%$ of larger viewpoint changes (w/ $50\%$ large). The
main paper reports results at $100\%$ medium and large viewpoint changes,
but we include an additional analysis with small changes in
Fig.~\ref{fig:compare}.
\thenumbers

\newpage
\section{Details on Entropy Evaluation}
\label{sec:entropydetails}
\figentropysupp
\noindent
As discussed in Sec.~\ref{subsec:implicitvsexplicit}, the relationship
between a source view $\xsrc$ and a target view $\xdst$ can
be quantified via the entropy of the probability distribution
that the transformer assigns to a target view $\xdst$, given 
a source frame $\xsrc$, camera transformation $\cam$ and conditioning
function $f$. More specifically, we first encode target, camera and source via
the encoder $\encoder$ and the conditioning function $f$ (see Sec.~\ref{subsec:transformertraining}), 
\ie $\sdst = \encoder(\xdst)$ and $f(\xsrc, \cam)$.
Next, for each element in the sequence $\sdst$, the (trained) transformer 
assigns a probability conditioned on the source and camera:
\begin{equation}
p\Big(\sdst_i \vert \sdst_{<i}, f(\xsrc, \cam)\Big), \quad 0 \leq i < \vert \sdst \vert
\end{equation}
where for our experiments the length of the target sequence is always
$\vert \sdst \vert = 13\cdot23= 299$, see also Sec.~\ref{sec:architectures}.
The entropy $\mathbb{H}(\sdst_i, \xsrc)$ for each position $i$ is then computed as 

\begin{equation}
\mathbb{H}(\sdst_i, \xsrc) ) = -\sum_k^{\vert \codebook \vert} p_k(\sdst_i \vert \sdst_{<i}, f(\xsrc, \cam)) \log p_k(\sdst_i \vert \sdst_{<i}, f(\xsrc, \cam))
\end{equation}
\noindent
Reshaping to the latent dimensionality $h \times w$ and bicubic upsampling to 
the input's size $H \times W$
then produces the visualizations of transformer entropy as in Fig.~\ref{fig:entropy} and Fig.~\ref{fig:entropysupp}.
Note that this approach quantifies the transformers uncertainty/surprise from a single example only and does
not need to be evaluated on multiple examples. 

\newpage
\section{Faithful Reconstructions/Compression Artifacts}
\label{sec:compressionartifacts}
\noindent Efficient training of the transformer models is enabled by the strong compression achieved with the VQGAN, which to some degree 
introduces artifacts but allows to trade compute
requirements for reconstruction quality. Larger discrete codes improve
fidelity (see Fig.~\ref{fig:compressions}, Tab.~\ref{tab:reconstructionmetrics}) but a 4$\times$ larger
code leads to approximately 16$\times$ larger costs when training the
transformer ($\mathcal{O}(n^2)$-complexity of attention).
Reducing such artifacts is thus a matter of scaling up hardware
or training time. Additionally, it would also increase the time required to sample a novel view, which is currently
$4.32\pm 0.04$ seconds.%

\reconstructionmetrics
\compressions

\figsupprealestate
\figsupprealestatesamples
\figsuppacid
\figsuppacidsamples
\figdepthprobesqualitativesupp

\FloatBarrier
\newpage
{\small
\bibliographystyle{ieee_fullname}
\bibliography{ms.bib}
}

\end{document}